\documentclass[10pt]{article}

\PassOptionsToPackage{sort,sort&compress}{natbib}

\usepackage[preprint]{tmlr}

\usepackage{emo}
\usepackage{styles/packages}
\usepackage{styles/fix-lmodern-sizes}
\usepackage{styles/orange}
\usepackage{styles/hyperref-setup}
\usepackage{styles/orangify-bad-ref}
\usepackage{styles/simpler-sections}
\usepackage{styles/appendix-patch}
\usepackage{styles/introduce-acronyms}
\usepackage{styles/symbol-img}
\usepackage{styles/improve-tables}
\usepackage{styles/stack-tri}
\usepackage{styles/minted-setup}
\usepackage{styles/appendix-toc}

\def\1{\bm{1}}

\DeclareMathAlphabet{\mathsfit}{\encodingdefault}{\sfdefault}{m}{sl}
\SetMathAlphabet{\mathsfit}{bold}{\encodingdefault}{\sfdefault}{bx}{n}

\def\gA{{\mathcal{A}}}
\def\gB{{\mathcal{B}}}

\def\gG{{\mathcal{G}}}

\def\gI{{\mathcal{I}}}

\def\gO{{\mathcal{O}}}
\def\gP{{\mathcal{P}}}

\def\gW{{\mathcal{W}}}
\def\gX{{\mathcal{X}}}

\title{GPTNT\emo{collision}: Benchmarking Real-Time Collaboration Between Multimodal Agents on \textit{Keep Talking And Nobody Explodes}}

\author{\centering\let\\\newline
\parbox[t]{\textwidth}{\centering
Amit Parekh$^{1*}$, Sabrina McCallum$^{2*}$, Kareem Al-Hasan$^{1*}$, \\
Malvina Nikandrou$^2$, Alessandro Suglia$^2$, Ioannis Konstas$^1$ \\
\normalfont
{\addr $^1$Heriot-Watt University} \qquad {\addr $^2$University of Edinburgh} \qquad {\addr $^{*}$Equal Contributions}}
}

\def\month{09}  
\def\year{2026} 
\def\openreview{\url{https://openreview.net/forum?id=BXJlDltbtq}} 

\newcommand{\gptnt}{\scalebox{1.08}[1.17]{\textls[15]{\textsc{gptnt}}}\xspace}
\newcommand{\ktane}{\scalebox{1.08}[1.17]{\textls[15]{\textsc{ktane}}}\xspace}

\begin{document}
\maketitle

\begin{abstract}

Multimodal models are increasingly deployed to solve tasks collaboratively with humans or other artificial agents. While existing benchmarks show that they possess the fundamental capabilities, the various conditions that coincide when collaborating---time pressure, information asymmetry, and imperfect communication---have traditionally been studied in isolation.\looseness=-1

To address this gap, we introduce \gptnt, a benchmark built on the cooperative video game \textit{Keep Talking and Nobody Explodes}, in which two agents must coordinate to defuse procedurally generated bomb puzzles against a live countdown.
One agent has access to the bomb but not the instructions for defusing it; the other holds the instructions but cannot see or manipulate the bomb.
Neither agent can succeed alone: the task requires contributions from both, and is solvable only through effective, efficient communication.
We remove turn-taking proxies or simplifications, instead requiring agents to act asynchronously and communicate in real time.\looseness=-1

\gptnt is designed to expose how models collaborate versus how they perform alone: the instruction manual, the partner, or both, can optionally be withheld to surface what a model has memorised versus what it derives in the moment. 
We demonstrate that \gptnt poses a considerable challenge to the state-of-the-art: not one of the closed- and open-source models we test defuses a single bomb in real time, a bar that human players clear. 
In a range of controlled experiments, we explore where capabilities break down, identifying critical weaknesses in state tracking, efficient acting within the time budget, handling ambiguity, and error recovery.\looseness=-1

We release \gptnt as a means for testing the collaborative performance that current benchmarks leave unmeasured. 
Since it runs on the real game, \gptnt benefits from procedural generation and inherits a living modding community: as models improve, the benchmark can be evolved to remain challenging, rather than being solved once and retired.\looseness=-1

\end{abstract}

\begin{figure}[tbh]
\vspace{-20px}
\centering
\includegraphics[width=1\linewidth]{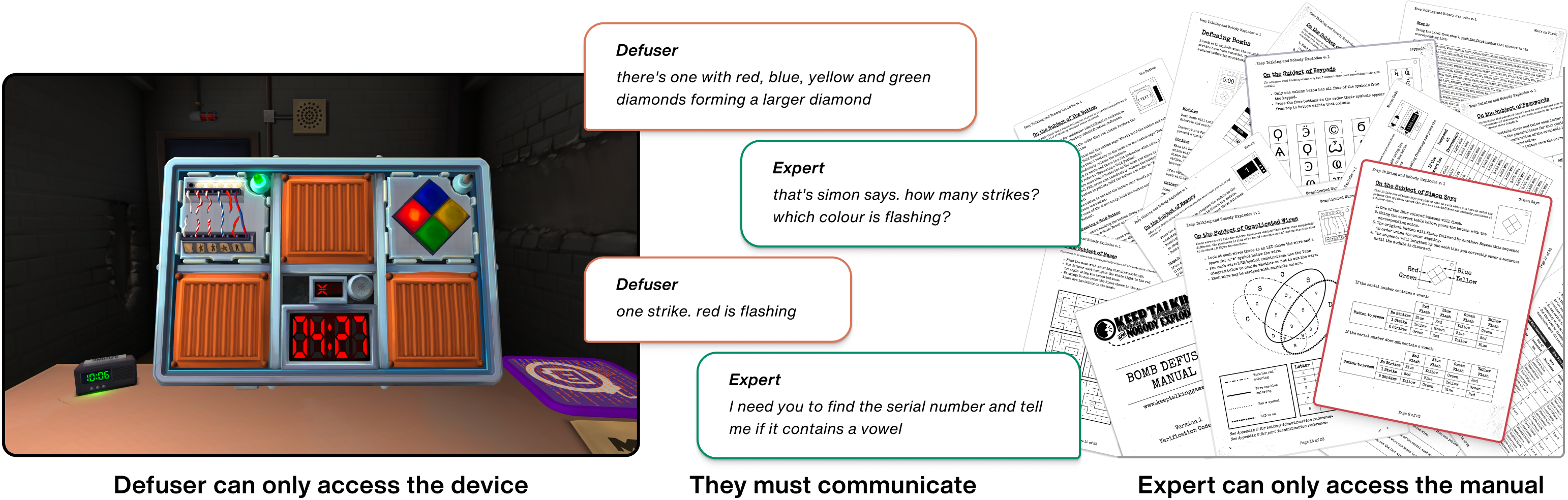}
\caption{High-level overview of the interaction in \textit{Keep Talking and Nobody Explodes}. Two agents must coordinate under asymmetric information and real-time pressure to solve a series of puzzles: the Defuser can see the bomb but has no manual, while the Expert has the manual but cannot see the bomb. }
\label{fig:high-level}
\end{figure}

\section{Introduction}

Multimodal large language models (MLLMs) can read documents \citep{Ma2024MMLONGBENCHDOCBenchmarkingLongcontext,Zhu2026MMDocBenchBenchmarkingLarge}, interpret complex visual scenes \citep{Wang2025MMLongBenchBenchmarkingLongContext,Li2025SurveyStateArt,Li2024MVBenchComprehensiveMultimodal}, and reason over long contexts \citep{Wang2025MMLongBenchBenchmarkingLongContext,Kuratov2024BABILongTestingLimits,wang2024needlemultimodalhaystack}---and they are being deployed to do so collaboratively, alongside humans and other agents \citep{Zou2025LLMBasedHumanAgentCollaboration}.
However, the conditions that define real-time collaboration---time pressure on decision-making and action execution, information asymmetry between partners, and the ensuing need to communicate effectively and efficiently across a visually complex and dynamic environment \citep{Gergle2004LanguageEfficiencyVisual,Kraut2003VisualInformationConversational}---are not typically tested together.

Generalist agentic benchmarks evaluate models across a broad range of interactive tasks in single-agent, static environments where communication is not required \citep{liu2024agentbench, ma2024agentboard, liu2024visualagentbench}.
Game-based environments offer richer visual complexity and clearer success criteria but remain fundamentally single-agent \citep{paglieri2024balrog, Zhang2025VideoGameBenchCanVisionLanguage};
cooperative settings exist but either remove the information asymmetry that makes communication necessary \citep{gong-etal-2024-mindagent},
bypass visual grounding \citep{Li2024EmbodiedAgentInterface,Shridhar2020ALFWorldAligningText}, or operate in environments that only change when agents act
\citep{Padmakumar2021TEAChTaskdrivenEmbodied, puig2020watchandhelp, Shridhar2020ALFWorldAligningText}.
No existing benchmark approximates real-world conditions by simultaneously demanding cooperative coordination under asymmetric information, real-time asynchronous action, visual grounding in a dynamic environment, and extended multi-turn communication.

The video game \textit{Keep Talking and Nobody Explodes} (\ktane; \citealp{KeepTalkingNobody})\footnote{\url{https://keeptalkinggame.com}} provides an ideal environment for testing these capabilities in an integrated fashion.
As illustrated by \cref{fig:high-level}, two players with strictly asymmetric information must collaborate in real time to defuse a procedurally generated bomb composed of a number of puzzle modules---one player can see the bomb but not the instruction manual, the other the manual but not the bomb. The rules in the instruction manual for defusing the bomb are arbitrary; therefore, possessing commonsense or prior knowledge about defusing real-world bombs or relevant domains such as engineering does not translate into an advantage for playing \ktane.
Fundamentally, neither player can succeed alone, and communication is the only mechanism through which the task can be solved.

We introduce \gptnt\footnote{Project page at \url{https://gptnt.github.io/}.}, a framework and corresponding benchmark that builds on top of \ktane to provide a rigorous evaluation environment for multimodal models.
Across the five state-of-the-art model families we test, \textbf{no model successfully defuses a single entry-level bomb in real time}---a bar that nine out of ten human player pairs can clear. Through systematic ablations that reduce time and task complexity, we reveal that models struggle to complete even the simplest setting: a bomb with a single puzzle module, and unlimited deliberation and generation time. We investigate how models perform zero-shot coordination in \textit{self-play} and \textit{other-play} \citep{Hu2020OtherPlayZeroShotCoordination,Stone2010AdHocAutonomous} and find that models dedicate a larger portion of the game to communication when collaborating with a different model.

Targeted offline evaluations and qualitative analyses confirm several well-documented low-level weaknesses of current models: models hallucinate visual and textual information \citep{Asadi2026MIRAGEIllusionVisual}, fail to identify when information provided by the other player is implausible \citep{Sharma2023UnderstandingSycophancyLanguage} or ambiguous \citep{Wang2025LearningAskWhen}, lose track of state across turns \citep{Laban2025LLMsGetLosta}, and rarely recover from mistakes \citep{Zhang2024HowLanguageModel}.
Beyond this, the benchmark aims to distinguish communicative competence from prior exposure to the game during training \citep{Magar2022DataContaminationMemorization}. To this end, we release evaluation components that serve both as diagnostic tools \textit{and} contamination checks, allowing us to systematically surface individual agents' reliance on memorisation shortcuts.



\gptnt is built to remain challenging as models improve.
\textbf{New missions can be procedurally generated and flexibly configured} by adjusting the time limit or varying the count, type, or placement of modules.
In addition, new module types released by \ktane's modding community can be introduced without rebuilding the framework---presenting a countermeasure to saturation and keeping the task space ahead of any fixed training distribution \citep{Foundation2026ARCAGI3NewChallenge}.

\needspace{1\baselineskip}
Our core contributions are as follows:
\begin{itemize}[leftmargin=*]
    \item \textbf{A benchmark that unifies the conditions for real-time collaboration.} \gptnt is, to our knowledge, the first benchmark to simultaneously require coordination under asymmetric information, real-time asynchronous action, visual grounding in a dynamic environment, and sustained multi-turn communication.



    \item \textbf{Comprehensive evaluation of frontier models.} We start from the most demanding setting---real-time collaboration on complex missions with multiple puzzle modules---and find that all tested frontier models break down (\cref{sec:actual-benchmark}). Following this, we systematically reduce time and task complexity to localise where model performance breaks down through simplified missions (\cref{sec:low-level}), an in-depth analysis of how models collaborate (\cref{sec:high-level}), and how they fail at both a higher level (\cref{sec:strategy:behaviour}) and a lower level (\cref{sec:statics}).


    \item \textbf{Diagnostic tools that surface memorisation.}
    We devise dedicated single-agent evaluations to measure the effect of collaboration, and remove access to the instruction manual to expose contamination (\cref{sec:ablations}).



\end{itemize}

\section{Related Work}

\begin{table}[tbh]
\vspace{-0.2\intextsep}
\setlength{\tabcolsep}{3pt}
\footnotesize
\centering
\renewcommand{\arraystretch}{1.2}
\begin{threeparttable}
\caption{Comparison with prior interactive MLLM benchmarks that require both vision and natural language. 
}
\label{tab:related_work}
\begin{tabularx}{\textwidth}{@{}X c c c  c c  c c c@{}}
\toprule
   & \multicolumn{3}{c}{Coordination}
   & \multicolumn{2}{c}{Environment}
   & \multicolumn{3}{c}{Context}
   \\
\cmidrule(lr){2-4} \cmidrule(lr){5-6} \cmidrule(l){7-9}
   & \makecell{Multi-Agent\tnote{*}} 
   & \makecell{Information\\Asymmetry}
   & \makecell{Asynchronous\\Action}
   & \makecell{Dynamic}
   & \makecell{Real-Time}
   & \makecell{Multi-Turn\\Communication}
   & \makecell{Image\\Sequence}
   & \makecell{Long\\Context}
   \\
\midrule
VisualWebArena \citeyearpar{koh2024visualwebarena} & \redcross & \redcross & \redcross & \redcross & \redcross & \redcross & \greencheck & \redcross \\ 
OSWorld \citeyearpar{xie2024osworld} & \redcross & \redcross & \redcross & \redcross & \redcross & \redcross & \greencheck & \greencheck \\ 
MineDojo \citeyearpar{fan2022minedojo} & \redcross & \redcross & \redcross & \greencheck & \redcross &  \redcross & \greencheck & \redcross \\
HAZARD \citeyearpar{zhou2024hazard} & \redcross & \redcross & \redcross & \greencheck & \redcross & \redcross & \redcross & \redcross \\
BALROG \citeyearpar{paglieri2024balrog} & \redcross & \redcross & \redcross & \greencheck & \redcross & \redcross & \greencheck & \greencheck \\
VideoGameBench \citeyearpar{Zhang2025VideoGameBenchCanVisionLanguage} & \redcross & \redcross & \redcross & \greencheck & \greencheck & \redcross & \greencheck & \greencheck \\
WatchAndHelp \citeyearpar{puig2018virtualhome, puig2020watchandhelp} & Co-op & \redcross & \greencheck & \redcross & \redcross & \redcross & \greencheck & \redcross \\ 
CoELA \citeyearpar{ICLR2024_54b8b4e0} & Co-op & \greencheck & \greencheck & \redcross & \redcross & \greencheck & \redcross & \redcross \\ 
TEACh \citeyearpar{Padmakumar2021TEAChTaskdrivenEmbodied} & Co-op & \greencheck & \redcross & \redcross & \redcross & \greencheck & \redcross & \redcross \\
Alexa Arena \citeyearpar{Gao2023AlexaArenaUserCentric} & Co-op & \greencheck & \redcross & \redcross & \redcross & \redcross & \greencheck & \redcross \\
WhodunitBench \citeyearpar{xie2024whodunitbench} & Comp & \greencheck & \redcross & \redcross & \redcross & \greencheck & \redcross & \greencheck \\
Multimodal CLEM \citeyearpar{hakimov-etal-2025-using} & Co-op & \greencheck & \redcross & \redcross & \redcross & \greencheck & \greencheck & \redcross \\
COMMA \citeyearpar{ossowski2024comma} & Co-op & \greencheck & \redcross & \redcross & \redcross & \greencheck & \redcross & \redcross \\
VS-Bench \citeyearpar{xu2025vs} & Mixed & \greencheck & \greencheck & \greencheck & \redcross & \redcross & \redcross & \redcross\\
\midrule
\textbf{\textls[50]{GPTNT}} & Co-op & \greencheck & \greencheck & \greencheck & \greencheck & \greencheck & \greencheck & \greencheck \\
\bottomrule
\end{tabularx}
\begin{tablenotes}
\item[*] Includes co-operative (Co-op), competitive (Comp), and mixed-motive (Mixed) games. We provide definitions in \cref{app:related-work}.
\end{tablenotes}
\end{threeparttable}
\end{table}

Benchmarks for evaluating MLLM agents range from static perception and reasoning tasks \citep{yue2025mmmu,chen2024mmstar,lu2024mathvista} to increasingly interactive settings that additionally demand planning, memory, and coordination.
\Cref{tab:related_work} compares \gptnt with existing interactive benchmarks that involve both vision and language, organised by the dimensions that distinguish them. We define each dimension and provide an extended literature review in \cref{app:related-work}.

Interactive GUI benchmarks such as OSWorld \citep{xie2024osworld} and VisualWebArena \citep{koh2024visualwebarena} evaluate visually-grounded decision-making for computer tasks, and progress in GUI agents has translated to video game interfaces \citep{OpenAI2025ComputerUsingAgent, wang2025uitars2technicalreportadvancing,wang2025game0tars0}.
In these non-dynamic environments, the state \textit{only} changes when the agent acts.
Certain game-based (e.g., BALROG, \citealp{paglieri2024balrog}; MineDojo, \citealp{fan2022minedojo}) and embodied benchmarks (e.g., HAZARD, \citealp{zhou2024hazard}) introduce dynamic environments that change independently of the agent.
Real-Time Reasoning Gym \citep{ICLR2026_ccbe1604} and VideoGameBench \citep{Zhang2025VideoGameBenchCanVisionLanguage} extend this to real-time evaluation, imposing latency constraints in text-based and multimodal games respectively.
All of these, however, test single agents, with no requirement to communicate.

Multi-agent settings with information asymmetry add a distinct challenge: when each agent holds knowledge the other lacks and neither can succeed alone, language becomes a necessary coordination channel \citep{xie2024whodunitbench, Padmakumar2021TEAChTaskdrivenEmbodied, Gao2023AlexaArenaUserCentric, ICLR2024_54b8b4e0}.
Cooperative environments often instantiate this asymmetry as an instructor--follower setup, but typically agents are not required to integrate information over a sequence of images, and traces stay short enough to avoid the long context that extended collaboration produces \citep{Laban2025LLMsGetLosta}.
Benchmarks, such as COMMA \citep{ossowski2024comma} and Multimodal CLEM \citep{hakimov-etal-2025-using}, focus on pragmatic communication in cooperative settings.
COMMA takes direct inspiration from \ktane puzzles but removes the multi-module structure, time pressure, and three-dimensional, dynamic environment to isolate communication as the main object of study.

As \cref{tab:related_work} shows, no existing benchmark instantiates all of these
properties at once within a single task, which limits how well they capture real collaborative interaction \citep{Paolo2024PositionCallEmbodied}.
VS-Bench \citep{xu2025vs} and VideoGameBench \citep{Zhang2025VideoGameBenchCanVisionLanguage} come closest in coverage, with VS-Bench being one of the few environments that support asynchronous coordination between agents---acting in parallel rather than in alternating turns.
However, both distribute their properties across heterogeneous game collections, instead of a single task to test these properties simultaneously.
\gptnt is the only benchmark where asynchronous multi-agent coordination under real-time pressure and extended multimodal context all hold at once---not because each is engineered in, but because \ktane itself demands them.





\begin{figure}[tbh]
\centering
\includegraphics[width=1\linewidth]{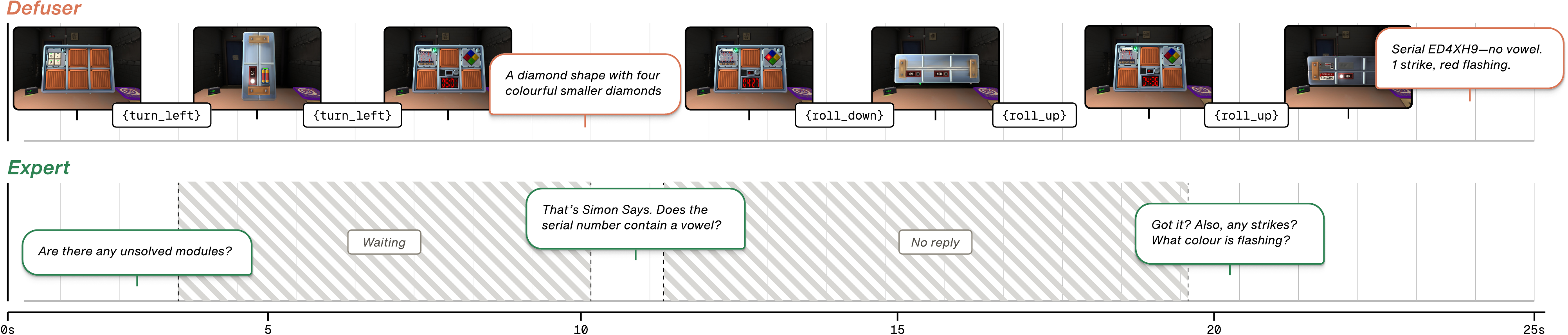}
\caption{Rollout of two agents interacting in real time. The Defuser gathers and relays information about modules and widgets incrementally by rotating the three-dimensional bomb. The Expert---who cannot see the bomb---identifies the module type from the Defuser's description and requests additional details on which the solution for the module depends (\cref{fig:simon-widget-ex}). Every action consumes game time, and agents do not share the same context: messages are the only channel between them.}
\label{fig:time-lane}
\vspace{-0.6\intextsep}
\end{figure}

\section{The Game: What is \textit{Keep Talking and Nobody Explodes?}}\label{sec:background}

\textit{Keep Talking and Nobody Explodes} (\ktane) is a popular cooperative bomb defusal video game where one player (the \textit{Defuser}) can see the bomb but lacks the knowledge needed to defuse it, while the other (the \textit{Expert}) has access to the instructions but cannot see the bomb.
The two players must bridge this difference through communication to defuse a bomb by solving a number of puzzle modules before the timer runs out and without exceeding the allowed number of strikes.
Crucially, neither player can directly apply any pre-existing knowledge about bombs, as the solutions to the puzzles in \ktane are specific to the game and are not applicable in any other setting. Additionally, there is a strict limit to how many random guesses can be attempted, as each failed attempt leads to a strike.
Together, these factors make \ktane a challenging test-bed for studying multimodal agentic collaboration under time pressure. 

\vspace{-1ex}
\paragraph{The Expert.}
Experts are limited to communicating with the other player or waiting to receive new information from them. Since the Expert cannot see the bomb, it must identify the correct instructions based exclusively on what the Defuser describes, making effective communication critical.

\vspace{-1ex}
\paragraph{The Defuser.}
Where the Expert can only communicate and wait, the Defuser can additionally explore and manipulate the bomb. Their actions ultimately depend on the instructions they receive from the Expert. 
Due to the three-dimensional nature of the bomb,\footnote{For an exploded view of an example bomb, see \cref{fig:exploded-layout}.} the Defuser's view is partial---only one side of the bomb is visible at a time. To inspect any other side, the Defuser must rotate the bomb vertically or horizontally (see \cref{fig:time-lane}).\looseness=-1

\begin{figure}[th]
\centering
\includegraphics[width=\linewidth]{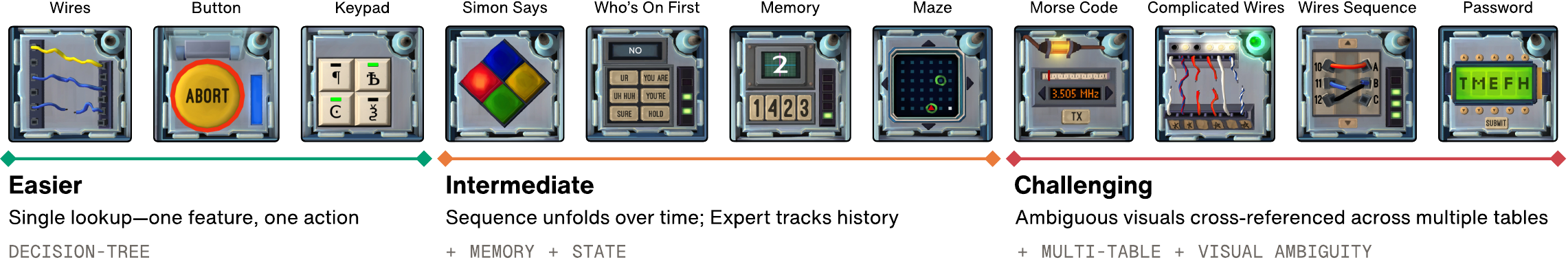}
\caption{Overview of the default puzzle module types in \ktane, grouped by their intended difficulty. Solving any module involves following conditional rules to a single correct 
action. Harder tiers increase difficulty by layering additional demands, introducing new perception, reasoning, and planning challenges.}
\label{fig:module-group}
\end{figure}

\paragraph{Modules.}
At a minimum, a bomb will contain the countdown timer and one puzzle module. At full capacity, a bomb can hold up to eleven modules across its front and back. All modules on a given bomb are independent of one another and can be solved in any order. There are eleven default\footnote{We exclude \textit{Needy Modules} from our experiments---a separate class of modules that require repeated attention at random intervals and cannot be evaluated in isolation. They rely on audio cues to signal when they require attention, even while the Defuser is focused elsewhere. We leave this to future work.} module \textit{types} in \ktane, ranging from easier to challenging (as illustrated in \cref{fig:module-group}), each of which poses a distinct challenge for the players. Individual instances of the same module type are sampled randomly and so, besides appearing visually distinct, they may require entirely different solutions.
We provide a description of each module type in \cref{app:ktane:modules}.\looseness=-1

\begin{wrapfigure}[16]{r}{0.25\textwidth}
\vspace{-1.2em}
\centering
\includegraphics[width=\linewidth]{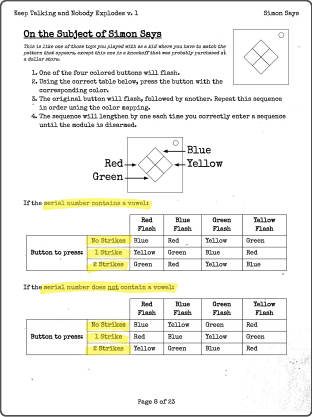}
\caption{Manual page for \simonsays; widget references highlighted.}
\label{fig:simon-widget-ex}
\end{wrapfigure}
\paragraph{Widgets.} The four sides of a bomb---left, right, top, and bottom---can carry a number of widgets: non-interactive components that are sampled randomly, and can include a serial number, various types of batteries, ports, and indicators. The correct solution to several module types depends on these values, for instance, whether an indicator is lit, the number of batteries, the presence of a specific type of port, or whether the last digit of the serial number is odd (e.g., \cref{fig:simon-widget-ex}).\looseness=-1

\paragraph{Missions.} In \ktane, a mission is a set of constraints used to procedurally generate similar bomb scenarios. These constraints include parameters such as the types and number of puzzle modules, the strike limit, and the time limit, and are used to achieve seven levels of increasing difficulty. Each of the seven difficulty levels is further broken down into sub-levels: for instance, level 2.1 introduces new modules, while level 2.4 tests this under a stricter time limit.
By default, each bomb scenario instantiated from the same mission definition will vary. For the purpose of the benchmark, our mission definitions include a pre-defined seed. This guarantees identical bomb scenarios for every player pair.\looseness=-1 

\paragraph{Outcomes.}
The bomb is defused, and a mission is considered successful when all puzzle modules on the bomb have been solved. Module-level progress can be tracked by observing an LED that lights up green when the module has been solved (see \complicatedwires; \cref{fig:module-group}). In some cases, modules provide additional partial progress signals (e.g., \memory; \cref{fig:module-group}). 
A failed mission ends in one of two terminal states:
a \textit{strikeout} occurs when the Defuser makes three mistakes, leading to three strikes, or a \textit{timeout} when the timer expires before all modules are solved.
The two states are diagnostically distinct: strikeout reflects acting on a wrong belief, timeout reflects failing to communicate and/or act efficiently enough.\looseness=-1

\section{The Benchmark: What is GPTNT?}\label{sec:exp-setup}

\gptnt turns \ktane into a zero-shot benchmark for evaluating real-time collaboration between MLLMs. 
Two model instances---one playing as the Defuser, and one as the Expert---coordinate to complete procedurally generated missions from the game. 
The model instances never share the same context: every piece of information can only be communicated in natural language (\cref{sec:agent-framework}).
As \ktane provides no native API, visual interface hook, or tooling for programmatic control, we develop this from scratch.
We release the full infrastructure required to run the benchmark: the game mod and microservice framework, the fixed suite of mission configurations, the processed manual, and the role-specific system prompts.\footnote{We do not provide access to \ktane itself, as that goes against the license. We discuss this further in \cref{sec:limitations}.}\looseness=-1






\subsection{Handling Time}

\begin{figure}[tbh]
\centering
\includegraphics[width=1\linewidth]{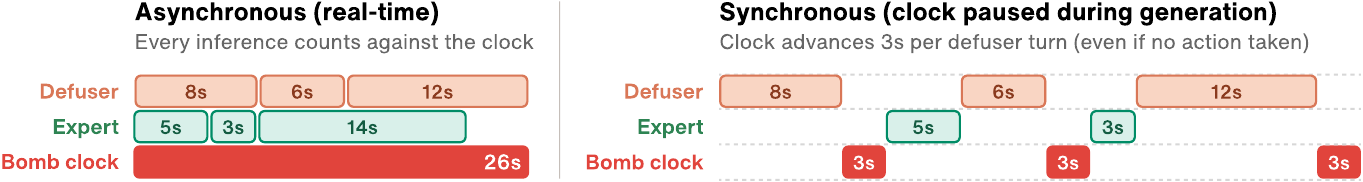}
\caption{Example showing how the generation time for each request relates to the in-game time for asynchronous and synchronous gameplay settings. 
In asynchronous mode (left), generation time per request counts against the bomb clock, simulating constraints of real-time decision-making. 
In synchronous mode (right), the game is paused during generation, decoupling task difficulty from generation latency.}
\label{fig:async-vis}
\end{figure}

\paragraph{\textit{Async}hronous mode.}
In the default setting of the benchmark, models play in \textit{asynchronous} real-time, mimicking conditions human players would encounter: the bomb clock advances in wall-clock time and does not pause while models are generating.
This departs from the turn-based protocol commonly used in agentic benchmarks (\cref{tab:related_work}), where the environment waits for each model to finish before the world state advances. 
We run the two perception-action loops in parallel; neither blocks the other, meaning the Expert can reason and speak while the Defuser acts (\cref{fig:async-vis}, left). 
Latency from generating tokens becomes part of the task, rather than an artefact to be controlled for---every generated token consumes real game time, penalising models that think exhaustively and communicate verbosely.\footnote{We cap the maximum number of new tokens per response as the reasoning models are systematically biased towards producing long reasoning chains that add little and consume disproportionately more generation time \citep{Sui2025StopOverthinkingSurvey,Chen2025NOTThinkThat}.}


\paragraph{\textit{Sync}hronous mode.} 
The benchmark additionally supports gameplay in a controlled \textit{synchronous} setting, in which the clock is paused during generation and instead advances a fixed increment per Defuser turn (\cref{fig:async-vis}, right). 
While this relaxes the real-time pressure of the asynchronous mode, models still need to complete missions within a fixed number of steps.  
This allows for separating the ability to reason accurately from the ability to reason quickly and succinctly, and lowers the barrier to entry by matching the turn-based interaction that current models typically assume.
In practice, we set the fixed increment to \SI{3}{\second}, allowing the transition animation after an action to complete before retrieving the updated visual state (\cref{app:time}). 

\subsection{Players}\label{sec:agent-framework}

We instantiate an independent agent process per role, each with its own perception-action loop, either from the \textit{same} model (self-play), or two \textit{different} models (other-play). 
As \gptnt is designed to probe certain capabilities of current models, we equip models with a minimal harness. 
A richer harness would conflate model capability with capabilities supplied by the surrounding system \citep{kapoor2025ai, zhang2026stop}, and any additional latency it incurs is directly counterproductive under the real-time constraints of the asynchronous setting.
We provide full implementation details, including the complete action schema, image resolutions per model, and the manual format, in \cref{app:framework}.

\paragraph{System prompt.} The Defuser and Expert receive separate system prompts built around their distinct role-specific responsibilities, action and observation spaces. Additionally, models are made aware of the specifics of time progression through their system prompts: in the synchronous setting, each action is specified to consume a fixed amount of game time, whereas in the asynchronous setting, models are informed that the clock advances in real time, proportional to the generation duration. The full prompts are provided in \cref{app:system_prompt}.\looseness=-1

\subsubsection{Observation space}
On each turn, the Defuser receives the current game frame (a screenshot from the game) alongside the immediately preceding frame.
Including the preceding frame is necessary as it contains the only reliable signal for inferring that an action produced a visible change in game state---such as receiving a strike, completing a sub-goal, or solving a module. By comparing the game time in the clocks between two frames, the model can also infer how much game time passes between actions.
For \morsecode and \simonsays, where information is encoded in flashing sequences that unfold over time, the Defuser instead receives a buffer of 16 frames captured at \SI{0.25}{\second} intervals---four seconds of gameplay, sized to contain at least one complete cycle of the flashing pattern (the longest letter for \morsecode; see \cref{app:mod:frames}).
The Expert receives each page of the bomb defusal manual as interleaved text and images, delivered in its first incoming message and protected from context truncation regardless of game length.
As the Expert cannot perceive the bomb, it entirely depends on the Defuser's descriptions to identify modules, request missing information, and reason over what to do.
Messages are free-form text and are delivered atomically to the recipient on its next forward pass; no information is dropped, summarised, or withheld in transit.

\begin{wrapfigure}[9]{r}{0.2\textwidth}%
\vspace{-1em}%
\includegraphics[width=\linewidth]{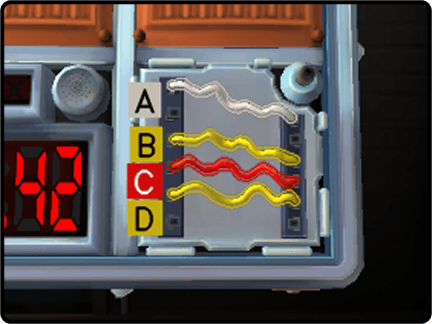}%
\captionsetup{font=small}%
\captionof{figure}{SoM on \textit{Wires}}%
\label{fig:som:wires-eg}%
\end{wrapfigure}%
\paragraph{Set-of-Marks.}
While \gptnt supports coordinate-based interaction, models using direct coordinate output perform substantially worse (see \cref{app:results:som-vs-coords} for preliminary results). We therefore adopt a \textit{set-of-marks} (SoM) approach \citep{Yang2023SetofMarkPromptingUnleashesa}, following its use in recent agentic settings \citep{Koh2024VisualWebArenaEvaluatingMultimodal,xie2024osworld,liu2024visualagentbench,Zhang2025UFOUIFocusedAgent}, to avoid obscuring agent capabilities with sub-par low-level execution. We overlay segmentation-mask labels on each frame, using letters rather than numbers to avoid ambiguity with numeric module content. Labels are positioned consistently relative to the interactive layout, and their colours match element properties where colour carries meaning (\cref{fig:som:wires-eg}). The full implementation details can be found in \cref{app:som}.

\subsubsection{Action space}
The Expert's action space is \textit{strictly} limited to sending a message or doing nothing---its sole contribution is providing instructions and guidance to the Defuser using natural language.
Beyond messaging and waiting, the Defuser can click, hold, or release interactive elements, zoom into or out of modules, as well as rotate or flip the bomb.
For both players, doing nothing is a first-class action. 
\ktane contains modules where, in fact, waiting is the correct action for the Defuser---for instance, \button requires releasing on a specific timer digit, and \morsecode and \simonsays require observing flashing sequences across multiple turns. The Expert may choose to wait while the Defuser is exploring and manipulating the bomb.\footnote{Waiting is also the result of the model causing any type of error, which we detail in \cref{app:nobf}.}

\paragraph{ReAct-style framework.}
Following recent work on agentic models \citep{Zhang2025VideoGameBenchCanVisionLanguage}, all models are prompted using a ReAct formulation \citep{Yao2022ReActSynergizingReasoning}. 
Beyond the established benefits of chain-of-thought reasoning \citep{Wei2022ChainofThoughtPromptingElicits}, the reasoning trace can act as an implicit working memory, allowing the Defuser to accumulate a representation of the game state from partial observations without having to provide every prior image as part of its context.
Specifically, models are prompted to output an explicit reasoning trace followed by a single action, delimited by XML tags, within a combined 1,000-token budget.\footnote{We do not take advantage of structured decoding or tool calling, and provide our rationale in \cref{app:output-format}.} 
While the benchmark supports extended reasoning---as seen in large reasoning models \citep{Xu2025LargeReasoningModels,Sui2025StopOverthinkingSurvey}---we effectively disable it for all evaluated models.
In preliminary experiments, models routinely consume over 4,000 reasoning tokens per turn before producing a response---equal to approximately one minute of wall-clock time.
In addition, thinking budgets are typically implemented at the API level, with no equivalent for locally served open-source models, and disabling thinking ensures configurations are comparable across models (full rationale in \cref{app:output-format}).

\subsection{Evaluation Protocol}

\gptnt is a zero-shot benchmark. 
The main evaluation comprises ten missions, configured at levels 2 and 3 out of the seven difficulty levels used in the game (full details in \cref{app:ktane-difficulty}).
The missions differ with respect to the number, types, and placement of modules, as well as the total time limit.
Unlike evaluating pass@$k$ on a single mission---where each additional attempt only provides evidence for how solvable that particular configuration is---sampling several missions and attempting each once (pass@1) allows us to estimate overall model competence across the wider task space at a lower compute cost \citep{Bouthillier2021AccountingVarianceMachine}.\looseness=-1

Besides mission success, we report partial success as the number of modules solved in those missions, and the cause of failure, distinguishing between accumulating too many mistakes or running out of time. 
While asynchronous gameplay is the default evaluation mode, we additionally report results in the synchronous setting (in \cref{sec:sync:multi}) to separate capability from response speed. 
Finally, for transparency, we report single-agent performance without manual access (in \cref{sec:e2-parametric-knowledge}) to disentangle collaborative capability from parametric knowledge of the game.

\section{Benchmarking Models on GPTNT}\label{sec:actual-benchmark}



We evaluate five recent MLLMs regarded as strong vision-language models, selected based on their prevalence and performance on existing benchmarks \citep{rein2024gpqa,osworld_verified,yue2025mmmu,chollet2025arc}.
These include \openai~GPT-5.2 \citep{OpenAI2025IntroducingGPT52}, \claude~Claude Sonnet 4.6 \citep{Anthropic2026IntroducingSonnet46}, \gemini~Gemini 3 Flash \citep{Google2025Gemini3Flash}, \qwen~Qwen3.5 \citep[27B;][]{QwenTeam2026Qwen35AcceleratingProductivity}, and \internvl~InternVL 3.5 \citep[38B;][]{Wang2025InternVL35AdvancingOpenSource}.
The open-source models are selected at the largest size that can be accommodated within our available compute. While considerably smaller than the evaluated closed-source ones, open-source models serve as reproducible and accessible baselines for the academic community. Models without sufficient context length and native multi-image inputs cannot support the task and are consequently not considered---see \cref{app:models:selection} for more details.
Further details on model selection and configuration are provided in \cref{app:models}.\looseness=-1

To contextualise model performance, we additionally collect a human reference point: ten pairs of human players (20 participants, nine of whom have never played \ktane) play the same ten multi-module missions under real-time conditions.
Since no pairing consists of two experienced players, we treat this as a soft upper bound on non-expert human performance.
Full recruitment details, protocol, and results are in \cref{app:humans}.
These results serve as a reference point rather than a controlled comparison with model performance, as participants communicate by voice instead of text.\footnote{Human players \textit{can} succeed using only text messages, confirming the task is not speech-dependent. However, models produce typed text significantly faster than most humans can type. In this iteration of the benchmark, we refrain from using implementations that rely on speech-to-text to avoid introducing transcription errors and other confounds. We discuss these limitations further in \cref{sec:limitations}.}
Each setup has structural advantages: humans benefit from the incrementality inherent to spoken dialogue, while models suffer no working-memory constraints, can access the full manual and exact dialogue history, and benefit from the precision of text when communicating symbols from various scripts, sidestepping the homophone ambiguities that arise in speech.



\begin{table}[tbh]
\caption{Game outcomes and number of modules solved for self-play across ten distinct missions in asynchronous mode, ordered by increasing difficulty (\cref{app:ktane-difficulty}). In the asynchronous setting, we do not stop the game clock while models are generating. Each mission is attempted once (pass@1). We also show the best and worst human pairings as reference points.}
\label{tab:e7-grid}
\centering
\footnotesize
\renewcommand{\arraystretch}{1.4}
\setlength{\tabcolsep}{3pt}
\setlength{\iconboxwidth}{0.45cm}
\sisetup{table-format=1.0}
\renewcommand{\iconboxheight}{0.4cm}
\begin{threeparttable}
\begin{tabular}{@{}
  l
  @{\hspace{\tabcolsep * 2}}
  c
  @{\hspace{\tabcolsep * 2}}
  c
  @{\hspace{\tabcolsep * 2}}
  c
  @{\hspace{\tabcolsep * 2 + 1pt}}
  c
  @{\hspace{\tabcolsep * 2 - 2pt}}
  c
  @{\hspace{\tabcolsep * 2 - 2pt}}
  c
  @{\hspace{\tabcolsep * 2 + 1pt}}
  c
  @{\hspace{\tabcolsep * 2}}
  c
  @{\hspace{\tabcolsep * 2}}
  c
  @{\hspace{\tabcolsep * 2}}
  c
  S
  @{}}
  \toprule
  & \multicolumn{10}{c}{Mission (ordered by difficulty)} \\
  \cmidrule(r){2-11}
  & 1 & 2 & 3 & 4 & 5 & 6 & 7 & 8 & 9 & 10 \\
  \cmidrule(r){2-11}
  \addlinespace[2pt]
  \multicolumn{1}{r}{\scriptsize\itshape\scshape front:}
    & \wiresicon[height=\iconboxheight,boxwidth=\iconboxwidth]\mazeicon[height=\iconboxheight,boxwidth=\iconboxwidth]
    & \buttonicon[height=\iconboxheight,boxwidth=\iconboxwidth]
    & \keypadicon[height=\iconboxheight,boxwidth=\iconboxwidth]\whosonfirsticon[height=\iconboxheight,boxwidth=\iconboxwidth]
    & \buttonicon[height=\iconboxheight,boxwidth=\iconboxwidth]\memoryicon[height=\iconboxheight,boxwidth=\iconboxwidth]
    & \wiresicon[height=\iconboxheight,boxwidth=\iconboxwidth]\memoryicon[height=\iconboxheight,boxwidth=\iconboxwidth]\mazeicon[height=\iconboxheight,boxwidth=\iconboxwidth]
    & \whosonfirsticon[height=\iconboxheight,boxwidth=\iconboxwidth]
    & \whosonfirsticon[height=\iconboxheight,boxwidth=\iconboxwidth]
    & \buttonicon[height=\iconboxheight,boxwidth=\iconboxwidth]
    & \keypadicon[height=\iconboxheight,boxwidth=\iconboxwidth]\wiresequenceicon[height=\iconboxheight,boxwidth=\iconboxwidth]\mazeicon[height=\iconboxheight,boxwidth=\iconboxwidth]
    & \wiresicon[height=\iconboxheight,boxwidth=\iconboxwidth]\morsecodeicon[height=\iconboxheight,boxwidth=\iconboxwidth]
  \\
  \multicolumn{1}{r}{\scriptsize\itshape\scshape back:}
    & \keypadicon[height=\iconboxheight,boxwidth=\iconboxwidth]
    & \wiresicon[height=\iconboxheight,boxwidth=\iconboxwidth]\keypadicon[height=\iconboxheight,boxwidth=\iconboxwidth]\memoryicon[height=\iconboxheight,boxwidth=\iconboxwidth]
    & \wiresicon[height=\iconboxheight,boxwidth=\iconboxwidth]\simonsaysicon[height=\iconboxheight,boxwidth=\iconboxwidth]
    & \whosonfirsticon[height=\iconboxheight,boxwidth=\iconboxwidth]
    & —
    & \keypadicon[height=\iconboxheight,boxwidth=\iconboxwidth]\simonsaysicon[height=\iconboxheight,boxwidth=\iconboxwidth]
    & \memoryicon[height=\iconboxheight,boxwidth=\iconboxwidth]\passwordsicon[height=\iconboxheight,boxwidth=\iconboxwidth]
    & \simonsaysicon[height=\iconboxheight,boxwidth=\iconboxwidth]\complicatedwiresicon[height=\iconboxheight,boxwidth=\iconboxwidth]
    & \passwordsicon[height=\iconboxheight,boxwidth=\iconboxwidth]
    & \whosonfirsticon[height=\iconboxheight,boxwidth=\iconboxwidth]\complicatedwiresicon[height=\iconboxheight,boxwidth=\iconboxwidth]
    & {\textit{Overall}}
  \\
  \midrule
  \claude~Sonnet 4.6
    & \textcolor{badbox}{\textbullet\kern-3pt\textopenbullet\kern-3pt\textopenbullet}\strikeout
    & \textcolor{badbox}{\textopenbullet\kern-3pt\textopenbullet\kern-3pt\textopenbullet\kern-3pt\textopenbullet}\strikeout
    & \textcolor{slowbox}{\textbullet\kern-3pt\textopenbullet\kern-3pt\textopenbullet\kern-3pt\textopenbullet}\timeout
    & \textcolor{slowbox}{\textbullet\kern-3pt\textopenbullet\kern-3pt\textopenbullet}\timeout
    & \textcolor{badbox}{\textopenbullet\kern-3pt\textopenbullet\kern-3pt\textopenbullet}\strikeout
    & \textcolor{badbox}{\textopenbullet\kern-3pt\textopenbullet\kern-3pt\textopenbullet}\strikeout
    & \textcolor{badbox}{\textbullet\kern-3pt\textopenbullet\kern-3pt\textopenbullet}\strikeout
    & \textcolor{slowbox}{\textbullet\kern-3pt\textopenbullet\kern-3pt\textopenbullet}\timeout
    & \textcolor{slowbox}{\textopenbullet\kern-3pt\textopenbullet\kern-3pt\textopenbullet\kern-3pt\textopenbullet}\timeout
    & \textcolor{badbox}{\textopenbullet\kern-3pt\textopenbullet\kern-3pt\textopenbullet\kern-3pt\textopenbullet}\strikeout
    & 0
  \\
  \gemini~Gemini 3 Flash
    & \textcolor{badbox}{\textopenbullet\kern-3pt\textopenbullet\kern-3pt\textopenbullet}\strikeout
    & \textcolor{slowbox}{\textopenbullet\kern-3pt\textopenbullet\kern-3pt\textopenbullet\kern-3pt\textopenbullet}\timeout
    & \textcolor{badbox}{\textopenbullet\kern-3pt\textopenbullet\kern-3pt\textopenbullet\kern-3pt\textopenbullet}\strikeout
    & \textcolor{badbox}{\textopenbullet\kern-3pt\textopenbullet\kern-3pt\textopenbullet}\strikeout
    & \textcolor{slowbox}{\textopenbullet\kern-3pt\textopenbullet\kern-3pt\textopenbullet}\timeout
    & \textcolor{slowbox}{\textbullet\kern-3pt\textopenbullet\kern-3pt\textopenbullet}\timeout
    & \textcolor{badbox}{\textbullet\kern-3pt\textopenbullet\kern-3pt\textopenbullet}\strikeout
    & \textcolor{slowbox}{\textbullet\kern-3pt\textopenbullet\kern-3pt\textopenbullet}\timeout
    & \textcolor{slowbox}{\textopenbullet\kern-3pt\textopenbullet\kern-3pt\textopenbullet\kern-3pt\textopenbullet}\timeout
    & \textcolor{badbox}{\textopenbullet\kern-3pt\textopenbullet\kern-3pt\textopenbullet\kern-3pt\textopenbullet}\strikeout 
    & 0
    \\
  \openai~GPT-5.2
    & \textcolor{slowbox}{\textbullet\kern-3pt\textopenbullet\kern-3pt\textopenbullet}\timeout
    & \textcolor{slowbox}{\textopenbullet\kern-3pt\textopenbullet\kern-3pt\textopenbullet\kern-3pt\textopenbullet}\timeout
    & \textcolor{slowbox}{\textbullet\kern-3pt\textbullet\kern-3pt\textopenbullet\kern-3pt\textopenbullet}\timeout
    & \textcolor{slowbox}{\textbullet\kern-3pt\textopenbullet\kern-3pt\textopenbullet}\timeout
    & \textcolor{slowbox}{\textopenbullet\kern-3pt\textopenbullet\kern-3pt\textopenbullet}\timeout
    & \textcolor{slowbox}{\textopenbullet\kern-3pt\textopenbullet\kern-3pt\textopenbullet}\timeout
    & \textcolor{badbox}{\textopenbullet\kern-3pt\textopenbullet\kern-3pt\textopenbullet}\strikeout
    & \textcolor{slowbox}{\textopenbullet\kern-3pt\textopenbullet\kern-3pt\textopenbullet}\timeout
    & \textcolor{slowbox}{\textopenbullet\kern-3pt\textopenbullet\kern-3pt\textopenbullet\kern-3pt\textopenbullet}\timeout
    & \textcolor{slowbox}{\textopenbullet\kern-3pt\textopenbullet\kern-3pt\textopenbullet\kern-3pt\textopenbullet}\timeout
    & 0
    \\
  \internvl~InternVL 3.5 (38B)
    & \textcolor{badbox}{\textopenbullet\kern-3pt\textopenbullet\kern-3pt\textopenbullet}\strikeout
    & \textcolor{slowbox}{\textopenbullet\kern-3pt\textopenbullet\kern-3pt\textopenbullet\kern-3pt\textopenbullet}\timeout
    & \textcolor{slowbox}{\textopenbullet\kern-3pt\textopenbullet\kern-3pt\textopenbullet\kern-3pt\textopenbullet}\timeout
    & \textcolor{badbox}{\textopenbullet\kern-3pt\textopenbullet\kern-3pt\textopenbullet}\strikeout
    & \textcolor{slowbox}{\textopenbullet\kern-3pt\textopenbullet\kern-3pt\textopenbullet}\timeout
    & \textcolor{slowbox}{\textopenbullet\kern-3pt\textopenbullet\kern-3pt\textopenbullet}\timeout
    & \textcolor{badbox}{\textopenbullet\kern-3pt\textopenbullet\kern-3pt\textopenbullet}\strikeout
    & \textcolor{slowbox}{\textbullet\kern-3pt\textopenbullet\kern-3pt\textopenbullet}\timeout
    & \textcolor{slowbox}{\textopenbullet\kern-3pt\textopenbullet\kern-3pt\textopenbullet\kern-3pt\textopenbullet}\timeout
    & \textcolor{slowbox}{\textopenbullet\kern-3pt\textopenbullet\kern-3pt\textopenbullet\kern-3pt\textopenbullet}\timeout 
    & 0
    \\
  \qwen~Qwen3.5 (27B)
    & \textcolor{slowbox}{\textopenbullet\kern-3pt\textopenbullet\kern-3pt\textopenbullet}\timeout
    & \textcolor{slowbox}{\textopenbullet\kern-3pt\textopenbullet\kern-3pt\textopenbullet\kern-3pt\textopenbullet}\timeout
    & \textcolor{slowbox}{\textopenbullet\kern-3pt\textopenbullet\kern-3pt\textopenbullet\kern-3pt\textopenbullet}\timeout
    & \textcolor{slowbox}{\textbullet\kern-3pt\textopenbullet\kern-3pt\textopenbullet}\timeout
    & \textcolor{slowbox}{\textbullet\kern-3pt\textopenbullet\kern-3pt\textopenbullet}\timeout
    & \textcolor{badbox}{\textopenbullet\kern-3pt\textopenbullet\kern-3pt\textopenbullet}\strikeout
    & \textcolor{badbox}{\textopenbullet\kern-3pt\textopenbullet\kern-3pt\textopenbullet}\strikeout
    & \textcolor{slowbox}{\textopenbullet\kern-3pt\textopenbullet\kern-3pt\textopenbullet}\timeout
    & \textcolor{slowbox}{\textopenbullet\kern-3pt\textopenbullet\kern-3pt\textopenbullet\kern-3pt\textopenbullet}\timeout
    & \textcolor{slowbox}{\textopenbullet\kern-3pt\textopenbullet\kern-3pt\textopenbullet\kern-3pt\textopenbullet}\timeout 
    & 0
    \\
  \midrule[0.1ex]
  \textit{Best Human Pairing}
    & \textcolor{goodbox}{\textbullet\kern-3pt\textbullet\kern-3pt\textbullet}\success
    & \textcolor{goodbox}{\textbullet\kern-3pt\textbullet\kern-3pt\textbullet\kern-3pt\textbullet}\success
    & \textcolor{goodbox}{\textbullet\kern-3pt\textbullet\kern-3pt\textbullet\kern-3pt\textbullet}\success
    & \textcolor{slowbox}{\textbullet\kern-3pt\textbullet\kern-3pt\textopenbullet}\timeout
    & \textcolor{goodbox}{\textbullet\kern-3pt\textbullet\kern-3pt\textbullet}\success
    & \textcolor{goodbox}{\textbullet\kern-3pt\textbullet\kern-3pt\textbullet}\success
    & \textcolor{slowbox}{\textbullet\kern-3pt\textbullet\kern-3pt\textopenbullet}\timeout
    & \textcolor{goodbox}{\textbullet\kern-3pt\textbullet\kern-3pt\textbullet}\success
    & \textcolor{slowbox}{\textbullet\kern-3pt\textbullet\kern-3pt\textbullet\kern-3pt\textopenbullet}\timeout
    & \textcolor{slowbox}{\textbullet\kern-3pt\textbullet\kern-3pt\textbullet\kern-3pt\textopenbullet}\timeout
    & \Uline{6}
    \\
  \textit{Worst Human Pairing}
    & \textcolor{badbox}{\textbullet\kern-3pt\textbullet\kern-3pt\textopenbullet}\strikeout
    & \textcolor{slowbox}{\textbullet\kern-3pt\textbullet\kern-3pt\textopenbullet\kern-3pt\textopenbullet}\timeout
    & \textcolor{badbox}{\textbullet\kern-3pt\textbullet\kern-3pt\textbullet\kern-3pt\textopenbullet}\strikeout
    & \textcolor{slowbox}{\textbullet\kern-3pt\textbullet\kern-3pt\textopenbullet}\timeout
    & \textcolor{slowbox}{\textbullet\kern-3pt\textbullet\kern-3pt\textopenbullet}\timeout
    & \textcolor{slowbox}{\textbullet\kern-3pt\textopenbullet\kern-3pt\textopenbullet}\timeout
    & \textcolor{slowbox}{\textbullet\kern-3pt\textbullet\kern-3pt\textopenbullet}\timeout
    & \textcolor{badbox}{\textbullet\kern-3pt\textopenbullet\kern-3pt\textopenbullet}\strikeout
    & \textcolor{badbox}{\textbullet\kern-3pt\textopenbullet\kern-3pt\textopenbullet\kern-3pt\textopenbullet}\strikeout
    & \textcolor{slowbox}{\textbullet\kern-3pt\textbullet\kern-3pt\textopenbullet\kern-3pt\textopenbullet}\timeout 
    & 0
    \\
  \bottomrule
\end{tabular}
\begin{tablenotes}
  \scriptsize
  \item Icons represent mission outcome: \success solved, \strikeout strikeout, \timeout timeout.
  \item Each bullet represents one module:\,\textbullet \,solved,\,\textopenbullet \,unsolved.
\end{tablenotes}
\end{threeparttable}
\end{table}
\subsection{Asynchronous Gameplay Results}\label{sec:async:multi_mod}

We first evaluate each model in asynchronous self-play---where time progresses in real time and without turn-taking constraints, and where both roles are filled by an instance of the same model---on ten missions with multiple puzzle modules. 
This setting combines the most demanding conditions we test.
Evaluated missions are configured according to the game's difficulty levels 2--3 out of seven difficulty levels in the game (full details in \cref{app:ktane-difficulty}) and  
\cref{tab:e7-grid} shows the game outcomes for these missions ordered by increasing difficulty.\looseness=-1

\paragraph{Models fail to solve a single mission.} No model fully completes a single mission in our full setting, though all models solve at least one module. The larger, closed-source models solve more individual modules than the smaller, open-source models, with Sonnet solving the most (5) and InternVL solving the fewest (1). 
Model performance falls far behind the human reference points: even the worst-performing human pair, while not completing any of the missions, solves more modules (18) than all models combined (15), whereas the best-performing human player pair solves twice as many modules (30), including six full missions. On average, humans manage to complete 2.8 out of 10 missions, and 20.8 out of 34 modules.

\paragraph{Both models and human players run out of time.} 
The dominant failure mode for both models and humans is timeout. Apart from Sonnet, which has more strikeouts than timeouts, models time out on anywhere between 50\% and 90\% of missions (detailed in \cref{app:results:multi-outcome-breakdown}). Humans fail by running out of time 74\% of the time across all games and player pairs (detailed in \cref{app:humans:results}).
However, interpreting timeouts in isolation is problematic: they conflate task difficulty with model-specific factors such as generation latency and verbosity, which leaves little room to understand the cause of failure.




\subsection{Synchronous Gameplay Results}\label{sec:sync:multi}


In the asynchronous setting---where the game runs in real time---total generation time becomes both a bottleneck and a confounder: a model pair that would otherwise succeed may simply time out.
For the purpose of this benchmark, we view total generation time as something shaped by model behaviour: a model that deliberates extensively or communicates verbosely will consume more of the time budget regardless of token throughput.
Some modules---such as \button---require that generated actions are \textit{aligned} with the time shown on the game clock; a response that arrives too late or too early will fail regardless of its correctness, making it impossible to attribute failure to capability, deliberation cost, timing, or speed alone.


We address this by running the same set of experiments in a \textit{synchronous mode}: between each Defuser turn, we freeze the game clock to give each model unlimited processing time.\footnote{We set Unity's internal time scale to zero to freeze the game in place.}
In this case, models take turns: the Defuser acts, the game advances, the Expert acts, and so on.
From this, we establish a capability ceiling on the examined missions: synchronous performance reflects an upper bound of what a model pair can achieve without putting time pressure on reasoning and communication. This helps to reveal collaborative abilities that may otherwise be hidden by end-to-end inference latency.\footnote{We calibrate each step to a real-time equivalent at three seconds per action; more details in \cref{app:time}.}

\begin{table}[htb]
\caption{Game outcomes and number of modules solved for self-play across ten distinct missions in synchronous mode, ordered by increasing difficulty (\cref{app:ktane-difficulty}). In the synchronous setting, we stop the game clock while models take turns generating. Each mission is attempted once (pass@1).}
\label{tab:e5-grid}
\centering
\footnotesize
\renewcommand{\arraystretch}{1.4}
\setlength{\tabcolsep}{3pt}
\setlength{\iconboxwidth}{0.45cm}
\sisetup{table-format=1.0}
\renewcommand{\iconboxheight}{0.4cm}
\begin{threeparttable}
\begin{tabular}{@{}
  l
  @{\hspace{\tabcolsep * 2}}
  c
  @{\hspace{\tabcolsep * 2}}
  c
  @{\hspace{\tabcolsep * 2}}
  c
  @{\hspace{\tabcolsep * 2 + 1pt}}
  c
  @{\hspace{\tabcolsep * 2 - 2pt}}
  c
  @{\hspace{\tabcolsep * 2 - 2pt}}
  c
  @{\hspace{\tabcolsep * 2 + 1pt}}
  c
  @{\hspace{\tabcolsep * 2}}
  c
  @{\hspace{\tabcolsep * 2}}
  c
  @{\hspace{\tabcolsep * 2}}
  c
  S
  @{}}
  \toprule
  & \multicolumn{10}{c}{Mission (ordered by difficulty)} \\
  \cmidrule(r){2-11}
  & 1 & 2 & 3 & 4 & 5 & 6 & 7 & 8 & 9 & 10 \\
  \cmidrule(r){2-11}
  \addlinespace[2pt]
  \multicolumn{1}{r}{\scriptsize\itshape\scshape front:}
    & \wiresicon[height=\iconboxheight,boxwidth=\iconboxwidth]\mazeicon[height=\iconboxheight,boxwidth=\iconboxwidth]
    & \buttonicon[height=\iconboxheight,boxwidth=\iconboxwidth]
    & \keypadicon[height=\iconboxheight,boxwidth=\iconboxwidth]\whosonfirsticon[height=\iconboxheight,boxwidth=\iconboxwidth]
    & \buttonicon[height=\iconboxheight,boxwidth=\iconboxwidth]\memoryicon[height=\iconboxheight,boxwidth=\iconboxwidth]
    & \wiresicon[height=\iconboxheight,boxwidth=\iconboxwidth]\memoryicon[height=\iconboxheight,boxwidth=\iconboxwidth]\mazeicon[height=\iconboxheight,boxwidth=\iconboxwidth]
    & \whosonfirsticon[height=\iconboxheight,boxwidth=\iconboxwidth]
    & \whosonfirsticon[height=\iconboxheight,boxwidth=\iconboxwidth]
    & \buttonicon[height=\iconboxheight,boxwidth=\iconboxwidth]
    & \keypadicon[height=\iconboxheight,boxwidth=\iconboxwidth]\wiresequenceicon[height=\iconboxheight,boxwidth=\iconboxwidth]\mazeicon[height=\iconboxheight,boxwidth=\iconboxwidth]
    & \wiresicon[height=\iconboxheight,boxwidth=\iconboxwidth]\morsecodeicon[height=\iconboxheight,boxwidth=\iconboxwidth]
  \\
  \multicolumn{1}{r}{\scriptsize\itshape\scshape back:}
    & \keypadicon[height=\iconboxheight,boxwidth=\iconboxwidth]
    & \wiresicon[height=\iconboxheight,boxwidth=\iconboxwidth]\keypadicon[height=\iconboxheight,boxwidth=\iconboxwidth]\memoryicon[height=\iconboxheight,boxwidth=\iconboxwidth]
    & \wiresicon[height=\iconboxheight,boxwidth=\iconboxwidth]\simonsaysicon[height=\iconboxheight,boxwidth=\iconboxwidth]
    & \whosonfirsticon[height=\iconboxheight,boxwidth=\iconboxwidth]
    & —
    & \keypadicon[height=\iconboxheight,boxwidth=\iconboxwidth]\simonsaysicon[height=\iconboxheight,boxwidth=\iconboxwidth]
    & \memoryicon[height=\iconboxheight,boxwidth=\iconboxwidth]\passwordsicon[height=\iconboxheight,boxwidth=\iconboxwidth]
    & \simonsaysicon[height=\iconboxheight,boxwidth=\iconboxwidth]\complicatedwiresicon[height=\iconboxheight,boxwidth=\iconboxwidth]
    & \passwordsicon[height=\iconboxheight,boxwidth=\iconboxwidth]
    & \whosonfirsticon[height=\iconboxheight,boxwidth=\iconboxwidth]\complicatedwiresicon[height=\iconboxheight,boxwidth=\iconboxwidth]
    & {\textit{Overall}}
  \\
  \midrule
  \claude~Sonnet 4.6
    & \textcolor{badbox}{\textopenbullet\kern-3pt\textopenbullet\kern-3pt\textopenbullet}\strikeout
    & \textcolor{goodbox}{\textbullet\kern-3pt\textbullet\kern-3pt\textbullet\kern-3pt\textbullet}\success
    & \textcolor{badbox}{\textopenbullet\kern-3pt\textopenbullet\kern-3pt\textopenbullet\kern-3pt\textopenbullet}\strikeout
    & \textcolor{slowbox}{\textbullet\kern-3pt\textbullet\kern-3pt\textopenbullet}\timeout
    & \textcolor{slowbox}{\textbullet\kern-3pt\textopenbullet\kern-3pt\textopenbullet}\timeout
    & \textcolor{slowbox}{\textbullet\kern-3pt\textbullet\kern-3pt\textopenbullet}\timeout
    & \textcolor{badbox}{\textopenbullet\kern-3pt\textopenbullet\kern-3pt\textopenbullet}\strikeout
    & \textcolor{badbox}{\textbullet\kern-3pt\textopenbullet\kern-3pt\textopenbullet}\strikeout
    & \textcolor{badbox}{\textopenbullet\kern-3pt\textopenbullet\kern-3pt\textopenbullet\kern-3pt\textopenbullet}\strikeout
    & \textcolor{badbox}{\textopenbullet\kern-3pt\textopenbullet\kern-3pt\textopenbullet\kern-3pt\textopenbullet}\strikeout
    & \Uline{1}
  \\
  \gemini~Gemini 3 Flash
    & \textcolor{badbox}{\textbullet\kern-3pt\textopenbullet\kern-3pt\textopenbullet}\strikeout
    & \textcolor{slowbox}{\textopenbullet\kern-3pt\textopenbullet\kern-3pt\textopenbullet\kern-3pt\textopenbullet}\timeout
    & \textcolor{slowbox}{\textbullet\kern-3pt\textopenbullet\kern-3pt\textopenbullet\kern-3pt\textopenbullet}\timeout
    & \textcolor{slowbox}{\textopenbullet\kern-3pt\textopenbullet\kern-3pt\textopenbullet}\timeout
    & \textcolor{slowbox}{\textopenbullet\kern-3pt\textopenbullet\kern-3pt\textopenbullet}\timeout
    & \textcolor{badbox}{\textopenbullet\kern-3pt\textopenbullet\kern-3pt\textopenbullet}\strikeout
    & \textcolor{slowbox}{\textbullet\kern-3pt\textbullet\kern-3pt\textopenbullet}\timeout
    & \textcolor{slowbox}{\textbullet\kern-3pt\textopenbullet\kern-3pt\textopenbullet}\timeout
    & \textcolor{badbox}{\textopenbullet\kern-3pt\textopenbullet\kern-3pt\textopenbullet\kern-3pt\textopenbullet}\strikeout
    & \textcolor{badbox}{\textopenbullet\kern-3pt\textopenbullet\kern-3pt\textopenbullet\kern-3pt\textopenbullet}\strikeout
    & 0
  \\
  \openai~GPT-5.2
    & \textcolor{slowbox}{\textopenbullet\kern-3pt\textopenbullet\kern-3pt\textopenbullet}\timeout
    & \textcolor{badbox}{\textbullet\kern-3pt\textopenbullet\kern-3pt\textopenbullet\kern-3pt\textopenbullet}\strikeout
    & \textcolor{badbox}{\textopenbullet\kern-3pt\textopenbullet\kern-3pt\textopenbullet\kern-3pt\textopenbullet}\strikeout
    & \textcolor{goodbox}{\textbullet\kern-3pt\textbullet\kern-3pt\textbullet}\success
    & \textcolor{slowbox}{\textbullet\kern-3pt\textopenbullet\kern-3pt\textopenbullet}\timeout
    & \textcolor{badbox}{\textbullet\kern-3pt\textopenbullet\kern-3pt\textopenbullet}\strikeout
    & \textcolor{badbox}{\textopenbullet\kern-3pt\textopenbullet\kern-3pt\textopenbullet}\strikeout
    & \textcolor{badbox}{\textbullet\kern-3pt\textopenbullet\kern-3pt\textopenbullet}\strikeout
    & \textcolor{badbox}{\textopenbullet\kern-3pt\textopenbullet\kern-3pt\textopenbullet\kern-3pt\textopenbullet}\strikeout
    & \textcolor{badbox}{\textopenbullet\kern-3pt\textopenbullet\kern-3pt\textopenbullet\kern-3pt\textopenbullet}\strikeout
    & \Uline{1}
  \\
  \internvl~InternVL 3.5 (38B)
    & \textcolor{slowbox}{\textbullet\kern-3pt\textopenbullet\kern-3pt\textopenbullet}\timeout
    & \textcolor{slowbox}{\textopenbullet\kern-3pt\textopenbullet\kern-3pt\textopenbullet\kern-3pt\textopenbullet}\timeout
    & \textcolor{slowbox}{\textbullet\kern-3pt\textopenbullet\kern-3pt\textopenbullet\kern-3pt\textopenbullet}\timeout
    & \textcolor{slowbox}{\textopenbullet\kern-3pt\textopenbullet\kern-3pt\textopenbullet}\timeout
    & \textcolor{slowbox}{\textopenbullet\kern-3pt\textopenbullet\kern-3pt\textopenbullet}\timeout
    & \textcolor{badbox}{\textopenbullet\kern-3pt\textopenbullet\kern-3pt\textopenbullet}\strikeout
    & \textcolor{badbox}{\textopenbullet\kern-3pt\textopenbullet\kern-3pt\textopenbullet}\strikeout
    & \textcolor{slowbox}{\textbullet\kern-3pt\textopenbullet\kern-3pt\textopenbullet}\timeout
    & \textcolor{slowbox}{\textopenbullet\kern-3pt\textopenbullet\kern-3pt\textopenbullet\kern-3pt\textopenbullet}\timeout
    & \textcolor{badbox}{\textbullet\kern-3pt\textopenbullet\kern-3pt\textopenbullet\kern-3pt\textopenbullet}\strikeout
    & 0
  \\
  \qwen~Qwen3.5 (27B)
    & \textcolor{badbox}{\textopenbullet\kern-3pt\textopenbullet\kern-3pt\textopenbullet}\strikeout
    & \textcolor{slowbox}{\textbullet\kern-3pt\textbullet\kern-3pt\textopenbullet\kern-3pt\textopenbullet}\timeout
    & \textcolor{badbox}{\textopenbullet\kern-3pt\textopenbullet\kern-3pt\textopenbullet\kern-3pt\textopenbullet}\strikeout
    & \textcolor{slowbox}{\textbullet\kern-3pt\textopenbullet\kern-3pt\textopenbullet}\timeout
    & \textcolor{slowbox}{\textopenbullet\kern-3pt\textopenbullet\kern-3pt\textopenbullet}\timeout
    & \textcolor{slowbox}{\textopenbullet\kern-3pt\textopenbullet\kern-3pt\textopenbullet}\timeout
    & \textcolor{slowbox}{\textopenbullet\kern-3pt\textopenbullet\kern-3pt\textopenbullet}\timeout
    & \textcolor{badbox}{\textbullet\kern-3pt\textopenbullet\kern-3pt\textopenbullet}\strikeout
    & \textcolor{slowbox}{\textopenbullet\kern-3pt\textopenbullet\kern-3pt\textopenbullet\kern-3pt\textopenbullet}\timeout
    & \textcolor{badbox}{\textopenbullet\kern-3pt\textopenbullet\kern-3pt\textopenbullet\kern-3pt\textopenbullet}\strikeout
    & 0
  \\
  \bottomrule
\end{tabular}
\begin{tablenotes}
\scriptsize
  \item Icons represent mission outcome: \success solved, \strikeout strikeout, \timeout timeout.
  \item Each bullet represents one module:\,\textbullet \,solved,\,\textopenbullet \,unsolved.
\end{tablenotes}
\end{threeparttable}\end{table}

\paragraph{The synchronous setting is only marginally less challenging for models.}
\cref{tab:e5-grid} shows that in this setting, which reduces \textit{time complexity}, models continue to struggle to solve our realistically complex missions. Only Sonnet and GPT, the models that solve the most and second-most individual modules in the asynchronous setting, complete a mission. While all models solve more individual modules than in the asynchronous setting, they still fail without solving a single module on at best half (Sonnet, GPT), and at worst seven (Qwen) of the ten missions.

\paragraph{Synchronous play shifts failures toward striking out.}
Across all models, strikeouts increase from 34\% in the asynchronous setting to half of total failures, matching the number of timeouts.
This is especially evident for the second-best performing model (GPT), for which 67\% of failures occur due to striking out, compared with only 10\% in the asynchronous setting (see \cref{app:results:multi-outcome-breakdown} for the full comparison).
However, models continue to exceed the time limit, with all models timing out on Mission \#5. This suggests that timeouts may also result from the specific combination of module difficulty and available game time.\footnote{Note that Missions \#4-6 have a reduced time limit, corresponding to level 2.4 in the game. See \cref{app:ktane-difficulty} for further details.}

\paragraph{Module difficulty shapes mission difficulty.}
Within the scope of the three- and four-module missions we test, module difficulty appears to be the most influential factor for overall success.
Models struggle most on the two hardest missions (\#9 and \#10), which each contain three of the more challenging module types.
The effect of module difficulty is inextricably intertwined with other characteristics of multi-module missions. Every additional module is a potential visual distractor and source of ambiguity, and models must understand and utilise incremental feedback as each module is solved.
\section{Evaluating Models on Simplified Missions}
\label{sec:low-level}

Single-module missions reduce the planning and perceptual demands of multi-module settings and allow us to evaluate model performance on each module type in isolation.
Critically, however, this isolation does not reduce the task to triviality: a single-module bomb still constitutes a sequential decision-making problem, requiring multiple perception-action loops between agents to resolve.
For each module type (illustrated in \cref{fig:module-group}), we sample ten sufficiently distinct missions.\footnote{We describe the protocol for determining the seeds used to instantiate the missions in \cref{app:seed-selection}.} In line with \cref{sec:actual-benchmark}, we consider the pass@1 across all ten missions of a given module type.


\subsection{Asynchronous Single-Module Gameplay Results}\label{sec:async:single_mod}

\begin{table}[ht]\centering\footnotesize\renewcommand{\arraystretch}{1.4}
\sisetup{table-format=2.1}
\begin{threeparttable}
\caption{Success rates (\%) for self-play on single-module missions in asynchronous mode, where the game clock continues to run down while models are generating. We sample ten distinct missions per module type. Each mission is attempted once (pass@1).}
\label{tab:e6-mod}
\begin{tabular}{@{}l *{12}{S} @{}}\toprule
 & {\wires*} & {\button*} & {\keypad*} & {\simonsays*} & {\whosonfirst*} & {\memory*} & {\maze*} & {\morsecode*} & {\complicatedwires*} & {\wiresequence*} & {\passwords*} & {Avg.}\\ \midrule
\claude*~Sonnet 4.6 & \Uline{60.0} & 30.0 & 30.0 & \color{black!40}0.0 & \color{black!40}0.0 & \color{black!40}0.0 & \color{black!40}0.0 & \color{black!40}0.0 & \color{black!40}0.0 & \color{black!40}0.0 & \Uline{20.0} & 12.7\\
\gemini*~Gemini 3 Flash & 20.0 & 10.0 & \Uline{40.0} & \color{black!40}0.0 & \Uline{20.0} & \Uline{50.0} & \color{black!40}0.0 & \Uline{20.0} & \color{black!40}0.0 & \color{black!40}0.0 & \color{black!40}0.0 & \Uline{14.5}\\
\openai*~GPT-5.2 & 30.0 & \Uline{60.0} & \color{black!40}0.0 & \color{black!40}0.0 & \color{black!40}0.0 & \color{black!40}0.0 & \color{black!40}0.0 & \color{black!40}0.0 & \color{black!40}0.0 & \color{black!40}0.0 & \color{black!40}0.0 & 8.2\\
\internvl*~InternVL 3.5 (38B) & 30.0 & 10.0 & 10.0 & \color{black!40}0.0 & \color{black!40}0.0 & \color{black!40}0.0 & \color{black!40}0.0 & \color{black!40}0.0 & \color{black!40}0.0 & \color{black!40}0.0 & \color{black!40}0.0 & 4.5\\
\qwen*~Qwen3.5 (27B) & 30.0 & 40.0 & 10.0 & \color{black!40}0.0 & \color{black!40}0.0 & \color{black!40}0.0 & \color{black!40}0.0 & \color{black!40}0.0 & \color{black!40}0.0 & \color{black!40}0.0 & \color{black!40}0.0 & 7.3\\
\midrule[0.1ex]
\textit{Average} & 34.0 & 30.0 & 18.0 & \color{black!40}0.0 & 4.0 & 10.0 & \color{black!40}0.0 & 4.0 & \color{black!40}0.0 & \color{black!40}0.0 & 4.0 & 9.5\\
\bottomrule
\end{tabular}
\end{threeparttable}\end{table}

\paragraph{Models achieve limited success on single-module missions.} 
As shown in \cref{tab:e6-mod}, all models are able to solve several single-module missions with the game running in real time.
This is in contrast with models' performance on the original missions with multiple modules (\cref{tab:e7-grid}), where models are unable to solve \textit{any} mission.
Gemini is the best-performing model in this setting, with an average success rate of 14.5\%, solving at least one mission on around half of the module types, while in the original multi-module missions, the model solves fewer individual modules than the other closed-source models. This suggests a disconnect between how reliably different models can solve individual modules \textit{in isolation} and in the presence of distractors.





\paragraph{In simplified missions, module difficulty begins to emerge.}
The single-module results broadly reflect the intended difficulty level for the different module types (see \cref{fig:module-group}), but reveal additional nuances within the difficulty tiers. Models are most successful on \wires, with all models solving at least two missions, followed by \button and \keypad, for which almost all models are able to solve at least a single mission.
Only Gemini succeeds on \whosonfirst and \memory missions, which correspond to the intermediate module types.
Models struggle with some of the module types in the most challenging tier---no model is able to solve a single mission for \complicatedwires and \wiresequence.
Observed difficulty deviates from the intended tiers for four modules: \simonsays and \maze prove harder than other intermediate modules---no model solves either---while \morsecode and \passwords prove easier than other challenging ones.\looseness=-1
\subsection{Synchronous Single-Module Gameplay Results}\label{sec:sync:single_mod}





While the results in \cref{sec:async:single_mod} show that models can solve \textit{some} missions with a single module, the real-time setting conflates problem-solving capability with time pressure: a failure may be the result of a missing skill or simply insufficient time to apply it.
We therefore repeat the single-module evaluation with the synchronous setting, freezing the game clock while models take turns generating.
\cref{tab:e1-mod-self-play-vs-async} reports success rates per module type, and illustrates the change in performance compared to the asynchronous setting in \cref{tab:e6-mod}.

\begin{table}[ht]
\centering
\footnotesize
\renewcommand{\arraystretch}{1.4}
\setlength{\tabcolsep}{4.5pt}
\sisetup{table-format=2.1}
\begin{threeparttable}
\caption{Success rates (\%) for self-play on single-module missions in synchronous mode, where we stop the game clock while models take turns generating. We sample ten distinct missions per module type. Each mission is attempted once (pass@1).}
\label{tab:e1-mod-self-play-vs-async}
\begin{tabular}{@{}
    l
    S
    @{\hspace{0.0pt}}
    l
    S
    @{\hspace{0.0pt}}
    l
    S
    @{\hspace{0.0pt}}
    l
    S
    @{\hspace{0.0pt}}
    l
    S
    @{\hspace{0.0pt}}
    l
    S
    @{\hspace{0.0pt}}
    l
    S
    @{\hspace{0.0pt}}
    l
    S
    @{\hspace{0.0pt}}
    l
    S
    @{\hspace{0.0pt}}
    l
    S
    @{\hspace{0.0pt}}
    l
    S[table-format=3.1]
    @{\hspace{0.0pt}}
    l
    S
    @{\hspace{0.0pt}}
    l
    @{}}
    \toprule
     & {\wires*} &
     & {\button*} &
     & {\keypad*} &
     & {\simonsays*} &
     & {\whosonfirst*} &
     & {\memory*} &
     & {\maze*} &
     & {\morsecode*} &
     & {\complicatedwires*} &
     & {\wiresequence*} &
     & {\passwords*} &
     & {\textit{Avg.}}
     \\
     \midrule
    \claude*~Sonnet 4.6
        & \Uline{80.0}
        & \inctrii
        & 50.0
        & \inctrii
        & \Uline{60.0}
        & \inctrii
        & \color{black!40}0.0
        &
        & 20.0
        & \inctrii
        & \Uline{80.0}
        & \inctrii
        & \Uline{10.0}
        & \inctrii
        & \Uline{10.0}
        & \inctrii
        & 20.0
        & \inctrii
        & \color{black!40}0.0
        &
        & \Uline{100.0}
        & \inctrii
        & \Uline{39.1} & {\scriptsize \inctrii} \\
    \gemini*~Gemini 3 Flash
        & 20.0
        &
        & 30.0
        & \inctrii
        & 40.0
        &
        & \color{black!40}0.0
        &
        & \Uline{50.0}
        & \inctrii
        & 60.0
        & \inctrii
        & \Uline{10.0}
        & \inctrii
        & \color{black!40}0.0
        & \dectrii
        & \color{black!40}0.0
        &
        & \Uline{10.0}
        & \inctrii
        & 10.0
        & \inctrii
        & 20.9 & {\scriptsize \inctrii} \\
    \openai*~GPT-5.2
        & 60.0
        & \inctrii
        & \Uline{80.0}
        & \inctrii
        & 30.0
        & \inctrii
        & \Uline{30.0}
        & \inctrii
        & 40.0
        & \inctrii
        & \Uline{80.0}
        & \inctrii
        & \color{black!40}0.0
        &
        & \Uline{10.0}
        & \inctrii
        & 10.0
        & \inctrii
        & \Uline{10.0}
        & \inctrii
        & 70.0
        & \inctrii
        & 38.2 & {\scriptsize \inctrii} \\
    \internvl*~InternVL 3.5 (38B)
        & 40.0
        & \inctrii
        & 40.0
        & \inctrii
        & 20.0
        & \inctrii
        & \color{black!40}0.0
        &
        & 10.0
        & \inctrii
        & \color{black!40}0.0
        &
        & \color{black!40}0.0
        &
        & \color{black!40}0.0
        &
        & \Uline{30.0}
        & \inctrii
        & \color{black!40}0.0
        &
        & \color{black!40}0.0
        &
        & 12.7 & {\scriptsize \inctrii} \\
    \qwen*~Qwen3.5 (27B)
        & 40.0
        & \inctrii
        & 40.0
        &
        & 30.0
        & \inctrii
        & \color{black!40}0.0
        &
        & \color{black!40}0.0
        &
        & \color{black!40}0.0
        &
        & \color{black!40}0.0
        &
        & \color{black!40}0.0
        &
        & \color{black!40}0.0
        &
        & \color{black!40}0.0
        &
        & 10.0
        & \inctrii
        & 10.9 & {\scriptsize \inctrii} \\
\midrule[0.1ex]
    \textit{Average}
        & 48.0
        & \inctrii
        & 48.0
        & \inctrii
        & 36.0
        & \inctrii
        & 6.0
        & \inctrii
        & 24.0
        & \inctrii
        & 44.0
        & \inctrii
        & 4.0
        & \inctrii
        & 4.0
        &
        & 12.0
        & \inctrii
        & 4.0
        & \inctrii
        & 38.0
        & \inctrii
        & 24.4
        & {\scriptsize \inctrii} \\
\bottomrule
\end{tabular}
\begin{tablenotes}\scriptsize
\item Indicators compare success against the asynchronous setting (\cref{tab:e6-mod}).
\end{tablenotes}
\end{threeparttable}
\end{table}

\paragraph{Collaborative gameplay itself is a bottleneck.} 
While stopping the game clock during generation and introducing turn-taking provides a notable boost across all models and module types, we reveal that the real-time constraints are not the only bottleneck. Even in the simplest setting, where we simplify missions to single modules \textit{and} grant models unlimited time to generate, only one-in-four missions are solved (\cref{tab:e1-mod-self-play-vs-async}). Both open- and closed-source models fail entirely on several module types, and even the strongest closed-source model cannot surpass an overall success rate of 39.1\%. This shows that models struggle with the core demands of our collaborative gameplay setting, and not solely the complexities introduced by real-time pressure or missions with multiple modules.

\paragraph{The gap between larger and smaller models becomes more apparent.}
Sonnet and GPT achieve the highest average success rates (39.1\% and 38.2\% respectively), with Gemini performing around average at 20.9\%. 
The closed-source models perform substantially better than the open-source models, solving around 3$\times$ as many missions as InternVL (12.7\%) and Qwen (10.9\%). This gap between open- and closed-source models likely reflects model scale, which is consistent with broader findings in the MLLM literature that more complex tasks benefit from scale \citep{Wei2022EmergentAbilitiesLarge,Li2025InferenceOptimalVLMs,Duan2024BotChatEvaluatingLLMs,Lee2025MultiVerseMultiturnConversation}. 

\paragraph{Some module types become easier than the intended difficulty.} Performance in the synchronous setting (\cref{tab:e1-mod-self-play-vs-async}) generally reflects the intended difficulty (\cref{fig:module-group})---with two notable exceptions. For models, the difficulty of the intermediate \memory aligns with the easier tier, which is plausible given every observation the Defuser makes is relayed to the Expert in text, converting what is a demanding working-memory task for humans into a context-retrieval task that models handle well. A similar shift becomes evident for \passwords, which is categorised as a hard module\footnote{The asynchronous results in \cref{tab:e6-mod} corroborate this assessment for real-time gameplay.} but presents as much less challenging. This module requires cycling through letter slots to find a valid word, a process that is procedurally straightforward once the time pressure is removed, with no implicit state to construct and no perceptual ambiguity.\footnote{
To solve \passwords missions, Sonnet performs on average 44.3 click actions in synchronous mode, while managing to complete an average of 34.1 clicks within the time limit in the asynchronous mode.}

\paragraph{Poor performance on certain module types reveals weaknesses in state tracking.} The modules that models struggle most with share a common structure: success requires constructing an \textit{implicit} representation of the game state across turns, whether a growing colour sequence (\simonsays), a cumulative wire-colour count (\wiresequence), a decoded letter sequence (\morsecode\footnote{We observe models getting stuck in runaway repetition of the visual signal (e.g., ``dot dash dash dot...'').}), or a spatial map of a 6×6 grid (\maze). These point to two distinct aspects of state tracking: losing detail when observations are not preserved in the dialogue (as in \maze and \wiresequence), and struggling to integrate multiple observations within a single turn (as in \simonsays and \morsecode).

\FloatBarrier

\section{Investigating Collaboration Patterns}\label{sec:high-level}


The preceding sections evaluate models under self-play conditions, where both roles are filled by the same model. This establishes per-model performance on the benchmark, but real-world collaboration often requires adapting to different partners with varying capability levels and communication approaches \citep{Stone2010AdHocAutonomous,Hu2020OtherPlayZeroShotCoordination}.
In this section, we expand the evaluation to all pairwise model combinations to determine how performance patterns vary with each model's role and partner. This also allows us to explore the relationship between communication rate---the proportion of the game that models spend communicating---and model pairings.
For our pairwise evaluations, we use the simplified single-module missions in synchronous mode to allow us to surface nuances in behaviour.










\subsection{Influence of Role and Partner}\label{sec:roles}

\begin{table}[ht]
\centering\footnotesize
\renewcommand{\arraystretch}{1.4}
\setlength{\tabcolsep}{5pt}
\sisetup{table-format=2.1}
\caption{Success rate (\%) of all model pairings for single-module missions in synchronous mode.}
\begin{threeparttable}
\label{tab:e1-matrix}
\begin{tabular}{@{}p{1.0mm} l *{6}{S} @{}}
\toprule
& & \multicolumn{6}{c}{\textbf{Expert}} \\
\cmidrule(l){3-8}
& & {\claude*~Sonnet 4.6\textsuperscript{$\bigstar$}}
  & {\gemini*~Gemini 3 Flash}
  & {\openai*~GPT-5.2}
  & {\internvl*~InternVL 3.5}
  & {\qwen*~Qwen3.5}
  & {\textit{Average}} \\
\midrule
\parbox[t]{1.0mm}{\multirow{6}{*}{\rotatebox[origin=c]{90}{\textbf{\textls[25]{Defuser}}}}}
& \claude*~Sonnet 4.6\textsuperscript{$\bigstar$} & 39.1 & \Uline{40.9} & 32.7 & 20.9 & 23.6 & 31.5 \\
& \gemini*~Gemini 3 Flash                        & 37.3 & 20.9          & 13.6 & 19.1 & 11.8 & 20.5 \\
& \openai*~GPT-5.2                               & 36.4 & 33.6          & 38.2 & 22.7 & 22.7 & 30.7 \\
& \internvl*~InternVL 3.5                        & 10.9 & 6.4           & 6.4  & 12.7 & 8.2  & 8.9  \\
& \qwen*~Qwen3.5                                 & 16.4 & 11.8          & 12.7 & 11.8 & 10.9 & 12.7 \\
\cmidrule[0.1ex](l){2-8}
& \textit{Average} & 28.0 & 22.7 & 20.7 & 17.5 & 15.5 & 20.9 \\
\bottomrule
\end{tabular}
\begin{tablenotes}
\scriptsize
\item[$\bigstar$\!\!] Highest average success rate for the role, taken over all pairings (\textit{Average} row and column). Highest success rate of any single pairing \Uline{underlined}.
\end{tablenotes}
\end{threeparttable}
\end{table}

\Cref{tab:e1-matrix} reports the success rate of each of the 25 pairings on single-module missions in synchronous mode. 
Success rates across pairings vary widely, ranging from 6.4\% to 40.9\%, indicating that performance is highly dependent on the exact model combination.

\paragraph{Performance varies more across Defusers than across Experts.} The spread of average performance across Defusers (22.6\%) is wider than that across Experts (13.5\%), confirming the intuition that the Defuser has a greater influence on overall success. 
A strong Expert cannot compensate for a weak Defuser: Qwen's poor performance as a Defuser improves somewhat with stronger Expert partners, but never approaches the stronger Defusers'. No Expert can make up for InternVL's shortcomings, pointing to a fundamental limitation in its Defuser capabilities. We return to the nature of this bottleneck in \cref{sec:statics:defuser}, where targeted offline evaluations isolate perception and reasoning failures. 

\paragraph{Models differ in their suitability for each role.} Comparing each model's average performance as Defuser and as Expert reveals an asymmetric role affinity. 
While Sonnet is the strongest player in both roles, the remaining model rankings reorder across roles. 
For instance, Gemini is stronger as an Expert (22.7\%) than as a Defuser (20.5\%), while GPT shows the reverse pattern, averaging 30.7\% as Defuser against 20.7\% as Expert. 
The direction of this asymmetry differs across models, pointing to role-specific suitability rather than a uniform difference in role difficulty.

\paragraph{Partner compatibility may shape performance beyond role capability.}
Gemini most evidently illustrates a sensitivity to compatibility. 
As Defuser, Gemini scores 19.1\% with InternVL as Expert, but only 13.6\% with GPT as Expert despite GPT being the stronger Expert on average. Gemini is also the only Expert that does not negatively affect Sonnet's performance as Defuser compared with self-play (39.1\%): when a Sonnet--Defuser is paired with another model was the Expert, performance drops to 20.9--32.7\%, but remains comparable when paired with Gemini (40.9\%). One hypothesis for this is that particular Defuser--Expert pairings coordinate better than their individual role skill would indicate, with one plausible cause being that divergent communication styles are shaped during post-training. Given the limited sample size---discussed further in \cref{sec:limitations}---we leave testing this hypothesis to future work.

\subsection{Communication Rates Across Model Pairings}\label{sec:communication:partners}


\begin{wraptable}[12]{r}{0.4\textwidth}
\vspace{-\intextsep}\vspace{-0.24em}
\footnotesize
\renewcommand{\arraystretch}{1.3}
\setlength{\tabcolsep}{6pt}
\sisetup{table-format=2.1}
\begin{threeparttable}
\caption{Communication rates (\% game steps with messages) for self- vs. other-play.\looseness=-1}
\label{tab:e1-msg-pct-self-vs-nonself}
\begin{tabularx}{\linewidth}{@{}X SSS @{}}
\toprule
Communicate with & {Self} & \multicolumn{2}{c}{Other} \\
\cmidrule(l){3-4}
 & & {as Defuser} & {as Expert} \\
\midrule
\claude*~Sonnet 4.6      & 22.5 & 34.0  & 26.3  \\
\gemini*~Gemini 3 Flash      & 25.5 & 36.6  & 24.4  \\
\openai*~GPT-5.2      & 35.5 & 27.1 &  38.7  \\
\internvl*~InternVL 3.5    & 34.3 & 30.6  & 41.0 \\
\qwen*~Qwen3.5        & 30.0 & 32.6  & 30.4   \\
\midrule[0.1ex]
\textit{Average} & 29.6 & 32.2  & 32.2  \\
\bottomrule
\end{tabularx}
\end{threeparttable}
\end{wraptable}

At any given point in a mission, models choose whether to send a message, wait, or in the case of the Defuser, interact with the bomb. 
We analyse the proportion of game steps involving a message from either of the players to reveal how models approach collaboration.\footnote{While we focus on the self- vs. other-play differences, we compare communication rates across all model pairings in \cref{app:extra:communication}.}

\paragraph{Communication rates differ in self-play and other-play.} 
As \cref{tab:e1-msg-pct-self-vs-nonself} summarises, on average, models dedicate more game time to communication in other-play, that is, when paired with a partner from a different model class. 
For some models, this holds regardless of the role---whether acting as the Defuser or the Expert, they consistently communicate more frequently when paired with another model. We specifically observe this for Sonnet.
Other models only exhibit this behaviour in other-play when acting as the Defuser (GPT and InternVL) or as the Expert (Gemini and Qwen).
Models collaborate in an ad hoc teamwork fashion \citep{Stone2010AdHocAutonomous,Hu2020OtherPlayZeroShotCoordination}: they are unaware who they are partnered with, and any adaptation is derived purely from the content and style of the messages they receive.\looseness=-1

\begin{figure}[tbh]
\centering
\includegraphics[width=\linewidth]{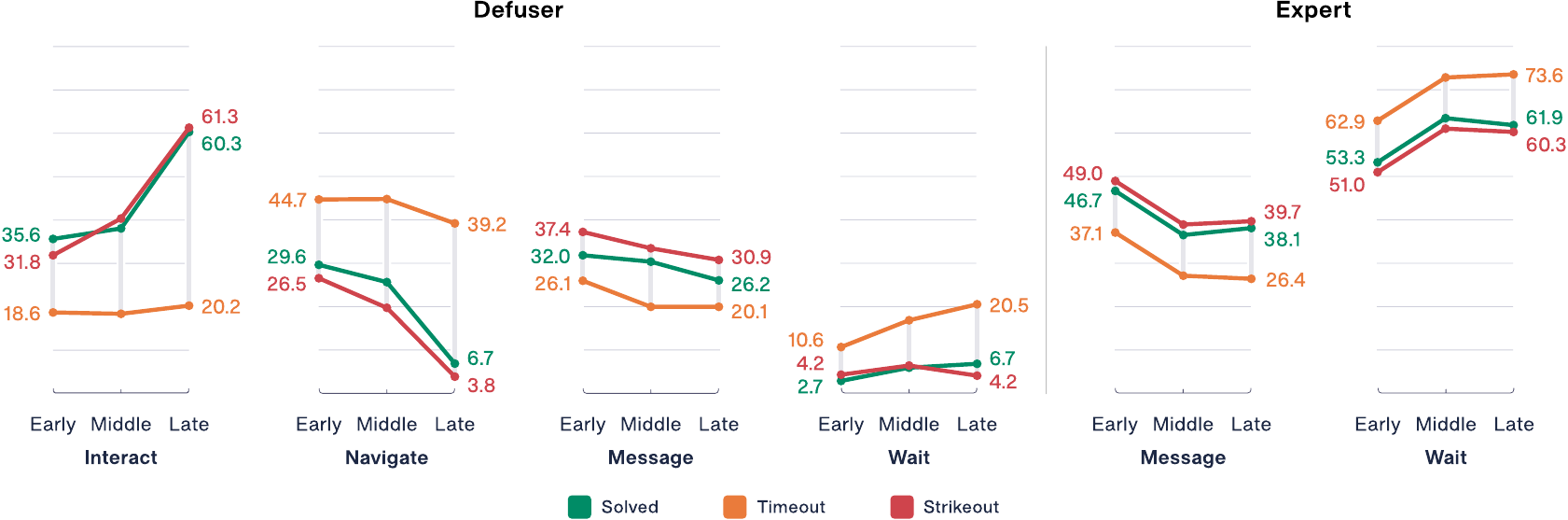}
\caption{
Distribution of action types (\% of game steps) by role and mission outcome across the early, middle, and late phases of games played by all possible model pairings on single-module missions in synchronous mode.\looseness=-1}
\label{fig:behaviour}
\end{figure}

\begin{figure}[tbh]
    \centering
    \includegraphics[width=0.9\linewidth]{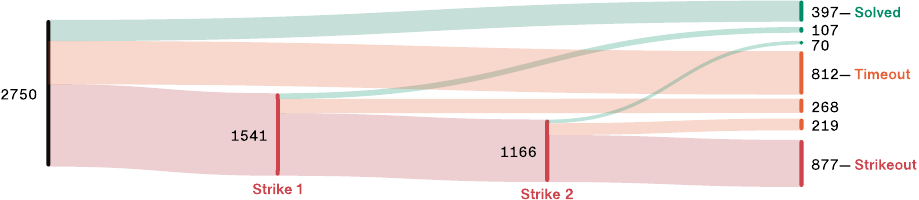}
    \caption{Frequency of game outcomes depending on strikes across games played by all possible model pairings on single-module missions in synchronous mode.}
    \label{fig:outcome-sankey}
\end{figure}

\section{Analysing How Models Fail}\label{sec:strategy:behaviour}

A mission succeeds only if every module is solved before the bomb reaches three strikes or runs out of time. 
We track strikeouts and timeouts separately, and examine what behavioural patterns distinguish them.
 
\paragraph{Timeouts and strikeouts have distinct signatures.}
Failed games are characterised by structurally different behavioural patterns depending on the
outcome (\cref{fig:behaviour}). Games ending in a timeout have a distinctive shape: Defusers spend roughly twice as much of the time played navigating around the bomb as those that succeed or strike out, and interactions with modules are almost half as frequent.
Experts mirror this behaviour by choosing to wait increasingly across game phases. Neither role fills the gap with communication, indicating that models might be stuck in decision loops that do not move the game forward. 
Games ending in a strikeout look mostly similar to solved games. The one consistent difference is messaging: both roles communicate more on games with a
strikeout than on either solved or timed-out ones, suggesting that the players are unable to reach sufficient common ground.\looseness=-1
\paragraph{Recovery is possible, but rare.}
In 56\% of synchronous, single-module missions, the players accumulate at least one strike.
\cref{fig:outcome-sankey} traces how those missions unfold. The single most common event following a strike is another strike, suggesting that models continue on the same course rather than correcting. While a small share of missions with one or even two strikes are still completed, demonstrating that recovery is possible, receiving a strike is a stronger predictor of further failure than any other outcome. 
Whether these patterns reflect compounding errors from an incorrect reasoning trace, an inability to recognise the strike as a meaningful signal and identify an alternative strategy, or a combination, cannot be disentangled without further systematic tests---we leave this to future work. 

\section{Isolating Underlying Capabilities Offline}
\label{sec:statics}

Interactive performance requires, and consequently conflates, many individual capabilities. To isolate these from the multi-turn demands of gameplay, we conduct targeted single-step offline
evaluations for each role.

\subsection{Manual VQA}\label{sec:statics:manual-vqa}

Given the manual and descriptions of what the Defuser sees, the Expert must identify the correct module type and corresponding page(s), determine what additional information they need from the Defuser, and use all available information to arrive at the correct instructions using deductive reasoning. While solutions are deterministic, the path to them can be derailed by ambiguity or inaccurate information supplied by the Defuser.\looseness=-1

\begin{figure}[t]
    \centering
    \includegraphics[width=1\linewidth]{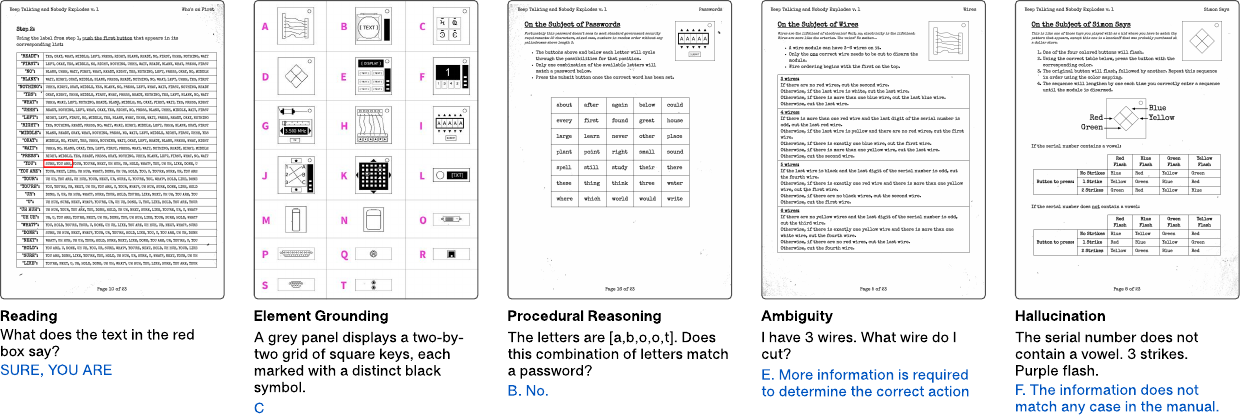}
\caption{Examples for each task in the offline Manual VQA.}
\label{fig:statics:manual}
\end{figure}

We isolate five distinct capabilities that a strong Expert should possess---reading comprehension, element grounding, procedural reasoning, ambiguity detection, and hallucination detection---and generate 884 targeted multiple-choice questions to assess each one independently.
These tasks are illustrated in \cref{fig:statics:manual}, and full construction details are provided in \cref{app:experiment-design:manual}.



\begin{table}[h]\centering\footnotesize\renewcommand{\arraystretch}{1.4}
\begin{threeparttable}
\sisetup{table-format=2.1, uncertainty-mode = separate}
\caption{Accuracy (\%) on offline Manual VQA tasks by model and capability measured.}
\label{tab:expert-statics-capabilities}
\begin{tabular}{@{}l *{5}{S} S @{}}\toprule
 & {Reading} & {Element Grounding} & {Procedural Reasoning} & {Ambiguity} & {Hallucination} & {\textit{Average}} \\ \midrule
\claude*~Sonnet 4.6 & 75.9 & 73.3 & \Uline{67.1} & 51.6 & \Uline{62.9} & 65.3 \\
\gemini*~Gemini 3 Flash & \Uline{81.2} & 88.3 & 64.1 & 51.6 & 50.0 & \Uline{69.3} \\
\openai*~GPT-5.2 & 65.4 & \Uline{90.0} & 61.5 & \Uline{59.4} & 61.4 & 64.4 \\
\internvl*~InternVL 3.5 (38B) & 29.3 & 38.3 & 44.5 & 23.4 & 25.7 & 36.1 \\
\qwen*~Qwen3.5 (27B) & 68.6 & 55.0 & 53.7 & 46.9 & 48.6 & 56.0 \\
\midrule[0.1ex]
\textit{Average} & 64.1 & 69.0 & 58.2 & 46.6 & 49.7 & 58.2 \\
\bottomrule
\end{tabular}
\end{threeparttable}\end{table}

\paragraph{Models exhibit distinct deficiencies in individual capabilities.}
Results in \cref{tab:expert-statics-capabilities} show that the Manual VQA tasks are non-trivial, with models achieving between 36.1 and 69.3\% overall accuracy. 
Despite comparable overall performance, the larger closed-source models show divergent capability profiles. Gemini's overall performance is driven by strong reading comprehension and element grounding, while GPT peaks on element grounding.
Sonnet, by contrast, is similarly capable on perceptual and reasoning tasks.
However, no model achieves a perfect accuracy on any task.  
Therefore, if any deficiency manifests at a single step, it can compound across multiple turns, leading to further deviations and errors.

\paragraph{Assessing information reliably is the crux.}
Both ambiguity and hallucination detection lag well behind perceptual and reasoning performance. 
GPT leads in detecting incomplete information, while Sonnet is the most capable at identifying hallucinated inputs that do not match the manual descriptions.
This is consistent with qualitative observations, where Sonnet more readily withholds instructions and requests clarification rather than reasoning over unreliable inputs.\footnote{See examples in \cref{fig:example:intern-claude-big_button-849,fig:example:intern-claude-password-798}.}  
Whether these failures reflect a reasoning limitation or a tendency toward sycophancy---where models are more likely to leave incorrect beliefs or information unchallenged \citep{Ibrahim2026TrainingLanguageModels}---remains an open question.








\paragraph{Manual VQA trends do not predict gameplay trends.} 
For some of the models, we observe a notable disconnect between their offline, single-step performance and the interactive setting. The superior performance of Gemini across these isolated core capabilities does not consistently translate to competitive gameplay (\cref{tab:e1-matrix}). InternVL is the weakest model on Manual VQA by a large margin, but performs comparably to Qwen in the interactive setting. This suggests that strong perceptual and reasoning capabilities can only ever be the basis for effective collaboration.

\subsection{Simulator VQA}\label{sec:statics:defuser}



\begin{figure}[tbh]
\vspace{0.7\intextsep}
    \centering
\includegraphics[width=1\linewidth]{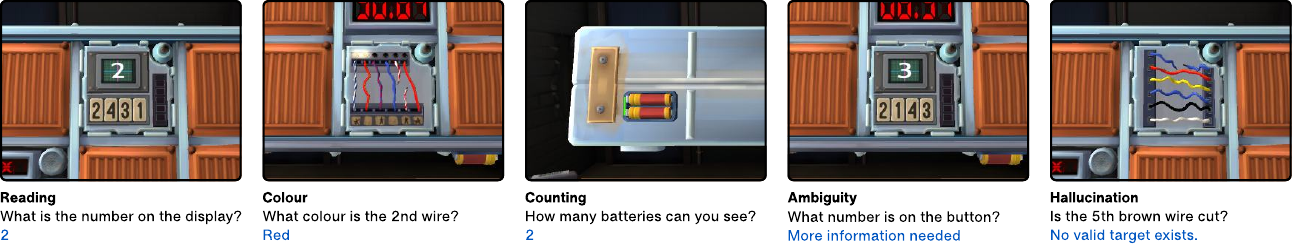}
\caption{Examples from the offline Simulator VQA evaluation tasks.}
\label{fig:statics:simulator-vqa}
\end{figure}

The Expert's contribution is only as useful as the information the Defuser supplies---if the Defuser supplies the Expert with inaccurate or insufficient information, the instructions the Expert provides will in turn be incorrect.
As the Defuser, a player must operate directly from visuals---reading characters, counting, and identifying elements based on their attributes---to describe the game state with enough precision for the Expert to reason over. 
We probe relevant abilities by generating 640 multiple-choice questions spanning basic perception (reading, colour, counting) and robustness probes for ambiguity and hallucination detection, examples for which are illustrated in \cref{fig:statics:simulator-vqa}. Full dataset construction details are provided in \cref{app:experiment-design:simulator}.



\paragraph{Reading is easy, counting is hard.} With an average performance of 88.9\%, reading alphanumeric characters on bomb elements is a seemingly straightforward task for models, confirming that Defusers are given a considerably simpler visual parsing task compared with the text-heavy manual pages the Expert must process (see \cref{tab:expert-statics-capabilities}). Colour recognition is relatively more challenging (77.7\%), and all models perform worst on counting (46.0\%). This is not surprising: while cardinality is trivial for humans, it remains a persistent weakness for large models \citep{Vo2025VisionLanguageModels,Paiss2023TeachingCLIPCount}. The harder the task, the more apparent the gap between the strongest and weakest model becomes; on counting tasks, the strongest model's accuracy is almost double that of the weakest (61.5\% vs. 35.1\%). The capabilities we measure are likely not simply scale-dependent: on the easier tasks, Qwen's performance is on par with that of the closed-source models.

\begin{table}[tbh]\centering\footnotesize\renewcommand{\arraystretch}{1.4}
\sisetup{table-format=2.1, uncertainty-mode = separate}
\caption{Accuracy (\%) on offline Simulator VQA tasks by model and capability measured.}
\begin{threeparttable}
\label{tab:defuser-statics-vqa-capabilities}
\begin{tabular}{@{}l *{5}{S} S @{}}\toprule
 & {Reading} & {Colour} & {Counting} & {Ambiguity} & {Hallucination} & {\textit{Average}} \\ \midrule
\claude*~Sonnet 4.6 & 96.7 & 82.0 & 47.8 & 6.6 & 77.9 & 63.0 \\
\gemini*~Gemini 3 Flash & \Uline{98.7} & 83.6 & \Uline{61.5} & \Uline{12.1} & \Uline{84.0} & \Uline{70.0} \\
\openai*~GPT-5.2 & 82.9 & 83.6 & 46.8 & 7.7 & 74.0 & 58.9 \\
\internvl*~InternVL 3.5 (38B) & 78.3 & 52.5 & 35.1 & 1.1 & 38.2 & 42.8 \\
\qwen*~Qwen3.5 (27B) & 88.2 & \Uline{86.9} & 39.0 & 8.8 & 67.9 & 56.9 \\
\midrule[0.1ex]
\textit{Average} & 88.9 & 77.7 & 46.0 & 7.3 & 68.4 & 58.3 \\
\bottomrule
\end{tabular}
\end{threeparttable}\end{table}



\paragraph{Detecting ambiguity is a bottleneck.}
Models can detect when they are asked or instructed about elements that do not exist in the image, averaging 68.4\%. 
However, all models struggle to detect ambiguous, underspecified questions. Even Gemini, the strongest model on most tasks, only achieves 12.1\%, and InternVL fails almost entirely, scoring only 1.1\%. 
This suggests that models can gauge when a question or instruction refers to an element that is clearly absent, but struggle when a description fits several elements. 
This hints at an important downstream bottleneck in communication between two players: even if the Expert and Defuser were maximally capable, the information asymmetry and time pressure make it almost impossible to avoid ambiguous exchanges entirely \citep{Piantadosi2012CommunicativeFunctionAmbiguity,Lemon2022ConversationalGroundingEmergent}, and knowing when to ask for clarification rather than proceeding under uncertainty becomes a critical skill \citep{Chiyah-Garcia2024RepairsBlockWorld}.\footnote{We include an example where the Expert fails to detect implausible information from the Defuser in \cref{fig:example:intern-claude-wire_sequence-813}.}

\paragraph{Low-level skills only partly explain in-game performance.} Echoing the findings in \cref{sec:statics:manual-vqa}, Gemini is the best-performing model across the Simulator VQA tasks, outperforming the other models by a large margin on the more difficult counting task, despite middling in-game performance as Defuser. On the other hand, InternVL's consistently poor performance on Simulator VQA tasks suggests that its in-game performance as Defuser is constrained by inadequate low-level capabilities.

\subsection{Simulator Localisation Results}\label{sec:statics:localisation}
\vspace{0.3em}

\begin{wrapfigure}[21]{l}{0.28\textwidth}
\vspace{-1.3em}
\centering
\begin{subfigure}[t]{\linewidth}
\centering
\includegraphics[width=\linewidth]{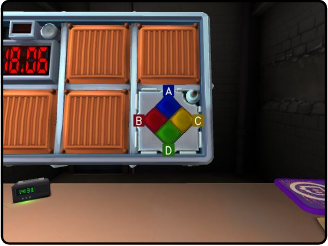}
\caption{\textit{Click on the green button.}}
\end{subfigure}%
\\[\medskipamount]%
\begin{subfigure}[t]{\linewidth}
\centering
\includegraphics[width=\linewidth]{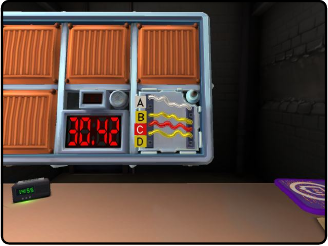}
\caption{\textit{Click on the first wire.}}
\end{subfigure}%
\caption{SoM localisation.}
\label{fig:statics:simulator-localisation}
\end{wrapfigure}

In \ktane, the Defuser must not only perceive, but also act on what it perceives. To interact with UI elements, the Defuser must localise the correct element according to the Expert's instructions. 
As discussed in \cref{sec:exp-setup}, we facilitate localisation via \textit{set-of-marks} (SoM; \citealp{Yang2023SetofMarkPromptingUnleashesa}). Labelled segmentation masks are overlaid on the image, reducing the action space to selecting the correct label. 
We distil this into an offline task consisting of 665 concise instructions, as illustrated by the examples in \cref{fig:statics:simulator-localisation}, and report the results in \cref{tab:localization-som}.
We also provide a comparison between SoM and direct coordinate prediction in \cref{app:results:som-vs-coords}.

\paragraph{Fine-grained localisation remains challenging.} A clear pattern is the prevalence of perfect accuracy across multiple modules, indicating that models are broadly capable of correctly identifying and targeting relevant elements given direct localisation instructions.
This corresponds to modules with sparse and visually distinct interaction targets, such as \button, \simonsays and \maze, suggesting that any downstream failures on these modules are unlikely to be grounding failures.
By contrast, modules that require interaction with smaller or more densely-packed elements---such as \wires, \whosonfirst, \wiresequence and \passwords---prove more challenging across models. This aligns with the intuition that models struggle when targets are smaller or when the screen contains many closely-clustered elements, making precise selection more difficult. 

\begin{table}[htb]\centering\footnotesize\renewcommand{\arraystretch}{1.4}
\caption{Accuracy (\%) on offline Simulator Localisation tasks, and in-game localisation errors.}
\setlength{\tabcolsep}{5pt}
\begin{threeparttable}
\label{tab:localization-som}
\begin{tabular}{@{}l
    S[table-format=2.1]
    S[table-format=3.1]
    S[table-format=2.1]
    S[table-format=3.1]
    S[table-format=2.1]
    S[table-format=3.1]
    S[table-format=3.1]
    S[table-format=3.1]
    S[table-format=2.1]
    S[table-format=2.1]
    S[table-format=2.1]
    S[table-format=2.1]
    @{}p{5mm}@{}
    S[table-format=2.1]
@{}}
\toprule
 & {\wires*} & {\button*} & {\keypad*} & {\simonsays*} & {\whosonfirst*} & {\memory*} & {\maze*} & {\morsecode*} & {\complicatedwires*} & {\wiresequence*} & {\passwords*} & {\textit{Avg.}} && {Errors\tnote{*}} \\
\midrule
\claude*~Sonnet 4.6 & 64.3 & \Uline{100.0} & 71.7 & \Uline{100.0} & \Uline{77.8} & 95.5 & \Uline{100.0} & \Uline{100.0} & \Uline{74.0} & 60.9 & 76.5 & 80.6 && 0.4 \\
\gemini*~Gemini 3 Flash & 64.3 & \Uline{100.0} & \Uline{98.1} & \Uline{100.0} & 53.7 & \Uline{100.0} & \Uline{100.0} & \Uline{100.0} & \Uline{74.0} & \Uline{69.6} & \Uline{97.1} & \Uline{83.9} && 0.0 \\
\openai*~GPT-5.2 & \Uline{71.4} & \Uline{100.0} & 81.1 & \Uline{100.0} & 68.5 & 95.5 & \Uline{100.0} & 96.9 & 72.0 & 67.4 & 64.7 & 80.4 && 0.2 \\
\internvl*~InternVL 3.5 (38B) & 50.0 & 95.5 & 71.7 & 66.7 & 22.2 & 18.2 & \Uline{100.0} & 65.6 & 58.0 & 37.0 & 23.5 & 53.7 && 28.4 \\
\qwen*~Qwen3.5 (27B) & 64.3 & \Uline{100.0} & 73.6 & \Uline{100.0} & 70.4 & \Uline{100.0} & \Uline{100.0} & 96.9 & 64.0 & 58.7 & 58.8 & 76.8 && 11.4 \\
\midrule[0.1ex]
\textit{Average} & 62.9 & 99.1 & 79.2 & 93.3 & 58.5 & 81.8 & 100.0 & 91.9 & 68.4 & 58.7 & 64.1 & 75.1 && 8.1 \\
\bottomrule
\end{tabular}
\begin{tablenotes}\scriptsize
\item[$*$\!\!] Proportion of attempted module interactions where the model hallucinates the location marker during \textit{interactive} gameplay.
\end{tablenotes}
\end{threeparttable}\end{table}

\paragraph{Localisation failure acts as a hard ceiling on downstream success.} This is especially prominent for InternVL, which fails to demonstrate sufficient localisation capabilities on the majority of modules. This fundamental gap is further evident in InternVL's in-game behaviour: 28.4\% of its attempted interactions with modules fail because the model hallucinates a location marker. Conversely, we observe a number of cases where the open-source models mistake the location marker for a part of the UI, despite extensive guidance for identifying location markers in the system prompt.\footnote{We provide examples of this behaviour in \cref{fig:example:intern-claude-big_button-849,fig:example:qwen-qwen-whos_on_first-337-async,fig:example:intern-claude-password-798} and the system prompt in \cref{fig:prompt:mechanics:defuser:som}.}
For the remaining models, however, localisation accuracy is high.
This suggests that solving modules like \simonsays, \morsecode, and \maze is hard not because models are unable to ground instructions to the correct element, but because correctly interpreting and acting on the module's state requires complementary capabilities.
\section{Ablating Collaboration and Surfacing Contamination}

\label{sec:ablations}

Information asymmetry forces every exchange through an imperfect channel that introduces noise and ambiguity, especially in other-play, where models are partnered up with different models (see \cref{sec:roles}).
We sever this channel and devise two single-agent scenarios to single out the effect of communication on the one hand, and prior exposure to \ktane on the other. 
While both violate the rules of the game, these settings serve as ablations to probe the importance of collaboration within the benchmark. We run both single-agent ablations on the single-module missions in synchronous mode, and compare them to the equivalent collaborative setting.

\begin{table}[tbh]
\centering
\footnotesize
\renewcommand{\arraystretch}{1.5}
\setlength{\tabcolsep}{4.5pt}
\sisetup{table-format=2.1}
\begin{threeparttable}
\caption{Success rate (\%) on simplified single-module missions in synchronous mode for agents playing in pairs (with any partner) vs single agents with and without manual access.}
\label{tab:ablation-mod}
\begin{tabular}{@{}
    p{1.0mm} 
    l
    S
    @{\hspace{0.0pt}}l
    S
    @{\hspace{0.0pt}}l
    S
    @{\hspace{0.0pt}}l
    S
    @{\hspace{0.0pt}}l
    S
    @{\hspace{0.0pt}}l
    S
    @{\hspace{0.0pt}}l
    S
    @{\hspace{0.0pt}}l
    S
    @{\hspace{0.0pt}}l
    S
    @{\hspace{0.0pt}}l
    S
    @{\hspace{0.0pt}}l
    S
    @{\hspace{0.0pt}}l
    S
    @{\hspace{0.0pt}}l 
    @{}}
\toprule
    &
    & {\wires*} &
    & {\button*} &
    & {\keypad*} &
    & {\simonsays*} &
    & {\whosonfirst*} &
    & {\memory*} &
    & {\maze*} &
    & {\morsecode*} & 
    & {\complicatedwires*} &
    & {\wiresequence*} & 
    & {\passwords*} &
    & {\textit{Avg.}}
    \\
    \midrule
    & \textit{Random Baseline}
     & 13.8 &  & 7.0  &  & \color{black!40}0.0 &  & \color{black!40}0.0 &
     & 0.1  &  & \color{black!40}0.0 &  & 2.1 &  & 0.9 &
     & 5.5 &  & \color{black!40}0.0  &  & \color{black!40}0.0 &  & 2.7 & \\
    \midrule
    \multirow{6}{*}{\rotatebox[origin=c]{90}{\textbf{\textls[25]{Pairwise}}\tnote{a}}}
    & \claude*~Sonnet 4.6
     & \Uline{70.0} &  & 48.0 &  & 20.0 &  & \color{black!40}0.0 &
     & \Uline{28.0} &  & \Uline{58.0} &  & 8.0 &  & \Uline{18.0} &
     & \Uline{28.0} &  & 4.0 &  & \Uline{64.0} &  & \Uline{31.5} & \\
    & \gemini*~Gemini 3 Flash
     & 36.0 &  & 56.0 &  & 24.0 &  & 2.0  &
     & 20.0 &  & 40.0 &  & \Uline{12.0} &  & 4.0  &
     & 6.0  &  & \Uline{18.0} &  & 8.0 &  & 20.5 & \\
    & \openai*~GPT-5.2
     & \Uline{70.0} &  & \Uline{66.0} &  & \Uline{32.0} &  & \Uline{12.0} &
     & 24.0 &  & 54.0 &  & 6.0 &  & 4.0 &
     & 14.0 &  & 6.0  &  & 50.0 &  & 30.7 & \\
    & \internvl*~InternVL 3.5 (38B)
     & 30.0 &  & 30.0 &  & 10.0 &  & \color{black!40}0.0 &
     & 4.0  &  & 4.0  &  & 2.0 &  & \color{black!40}0.0 &
     & 18.0 &  & \color{black!40}0.0  &  & \color{black!40}0.0 &  & 8.9 & \\
    & \qwen*~Qwen3.5 (27B)
     & 32.0 &  & 40.0 &  & 16.0 &  & \color{black!40}0.0 &
     & 6.0  &  & 20.0 &  & 6.0 &  & \color{black!40}0.0 &
     & 2.0  &  & \color{black!40}0.0  &  & 18.0 &  & 12.7 & \\
    \arrayrulecolor{black!70}\cmidrule[0.05ex](l){2-26}\arrayrulecolor{black}
    & \textit{Average}
     & 47.6 &  & 48.0 &  & 20.4 &  & 2.8  &
     & 16.4 &  & 35.2 &  & 6.8  &  & 5.2  &
     & 13.6 &  & 5.6  &  & 28.0 &  & 20.9 & \\
    \midrule
    \multirow{6}{*}{\rotatebox[origin=c]{90}{\textbf{\textls[25]{No partner}}\tnote{b}}}
    & \claude*~Sonnet 4.6
     & 90.0  & \inctrii & 90.0 & \inctrii & 70.0 & \inctrii & \color{black!40}0.0 &
     & 20.0  &          & 20.0 & \dectrii & \color{black!40}0.0 &          & 10.0 &
     & 40.0  & \inctrii & \color{black!40}0.0 &  & 100.0 & \inctrii
     & 40.0 & {\scriptsize \inctrii} \\
    & \gemini*~Gemini 3 Flash
     & 10.0 & \dectrii & 60.0 &          & 20.0 &          & \color{black!40}0.0 &
     & \color{black!40}0.0 & \dectrii & 20.0 & \dectrii & \color{black!40}0.0 & \dectrii & 30.0 & \inctrii
     & \color{black!40}0.0 & & \color{black!40}0.0 & \dectrii & \color{black!40}0.0 &
     & 12.7 & {\scriptsize \dectrii} \\
    & \openai*~GPT-5.2
     & 100.0 & \inctrii & 80.0 & \inctrii & 60.0 & \inctrii & 10.0 & \notri
     & 30.0  &          & 50.0 &          & \color{black!40}0.0 & & \color{black!40}0.0 &
     & 40.0  & \inctrii & \color{black!40}0.0 &  & 10.0 & \dectrii
     & 34.5 & {\scriptsize \inctrii} \\
    & \internvl*~InternVL 3.5 (38B)
     & 70.0 & \inctrii & 30.0 &  & 30.0 & \inctrii & \color{black!40}0.0 &
     & \color{black!40}0.0 &  & \color{black!40}0.0 &  & 10.0 &  & \color{black!40}0.0 &
     & 10.0 &  & \color{black!40}0.0 &  & \color{black!40}0.0 &
     & 13.6 & {\scriptsize \inctrii} \\
    & \qwen*~Qwen3.5 (27B)
     & 80.0 & \inctrii & 40.0 &  & 30.0 & \inctrii & \color{black!40}0.0 &
     & 10.0 &  & 30.0 & \inctrii & \color{black!40}0.0 & & 20.0 & \inctrii
     & 40.0 & \inctrii & \color{black!40}0.0 &  & 20.0 &
     & 24.5 & {\scriptsize \inctrii} \\
    \arrayrulecolor{black!70}\cmidrule[0.05ex](l){2-26}\arrayrulecolor{black}
    & \textit{Average}
     & 70.0 & \inctrii & 60.0 & \inctrii & 42.0 & \inctrii & 2.0  &
     & 12.0 & \dectrii & 24.0 & \dectrii & 2.0 & \dectrii & 12.0 & \inctrii
     & 26.0 & \inctrii & \color{black!40}0.0 & \dectrii & 26.0 & \dectrii
     & 25.1 & {\scriptsize \inctrii} \\
    \midrule
    \multirow{6}{*}{\rotatebox[origin=c]{90}{\textbf{\textls[25]{No manual}}\tnote{c}}}
    & \claude*~Sonnet 4.6
     & 50.0 & \dectrii & 20.0 & \dectrii & 40.0 & \inctrii & \color{black!40}0.0 &
     & 10.0 & \dectrii & 10.0 & \dectrii & 10.0 &          & 10.0 &
     & 70.0 & \inctrii & \color{black!40}0.0 &  & 70.0 &
     & 26.4 & {\scriptsize \dectrii} \\
    & \gemini*~Gemini 3 Flash
     & 10.0 & \dectrii & 60.0 &          & 40.0 & \inctrii & \color{black!40}0.0 &
     & 10.0 & \dectrii & \color{black!40}0.0 & \dectrii & \color{black!40}0.0 & \dectrii & \color{black!40}0.0 &
     & 10.0 & & \color{black!40}0.0 & \dectrii & \color{black!40}0.0 &
     & 11.8 & {\scriptsize \dectrii} \\
    & \openai*~GPT-5.2
     & 100.0 & \inctrii & 90.0 & \inctrii & \color{black!40}0.0 & \dectrii & 20.0 & \notri
     & \color{black!40}0.0 & \dectrii & 10.0 & \dectrii & \color{black!40}0.0 & & \color{black!40}0.0 &
     & 20.0 & & \color{black!40}0.0 &  & 20.0 & \dectrii
     & 23.6 & {\scriptsize \dectrii} \\
    & \internvl*~InternVL 3.5 (38B)
     & 10.0 & \dectrii & 30.0 &  & 10.0 &  & \color{black!40}0.0 &
     & \color{black!40}0.0 &  & \color{black!40}0.0 &  & \color{black!40}0.0 &  & 10.0 & \inctrii
     & 60.0 & \inctrii & \color{black!40}0.0 &  & \color{black!40}0.0 &
     & 10.9 & {\scriptsize \inctrii} \\
    & \qwen*~Qwen3.5 (27B)
     & 70.0 & \inctrii & \color{black!40}0.0 & \dectrii & \color{black!40}0.0 & \dectrii & \color{black!40}0.0 &
     & 10.0 &  & \color{black!40}0.0 & \dectrii & 10.0 & & 10.0 & \inctrii
     & 20.0 & \inctrii & 10.0 & \inctrii & \color{black!40}0.0 & \dectrii
     & 11.8 & \\
    \arrayrulecolor{black!70}\cmidrule[0.05ex](l){2-26}\arrayrulecolor{black}
    & \textit{Average}
     & 48.0 &          & 40.0 & \dectrii & 18.0 & \dectrii & 4.0  &
     & 6.0  & \dectrii & 4.0  & \dectrii & 4.0  & \dectrii & 6.0  &
     & 36.0 & \inctrii & 2.0  & \dectrii & 18.0 & \dectrii
     & 16.9 & {\scriptsize \dectrii} \\
    \bottomrule
\end{tabular}
\begin{tablenotes}\scriptsize
    \item[a] Pairwise collaboration aggregated over all Expert partners.
    \item[b] Single-agent play with access to the manual.
    \item[c] Single-agent play without access to the manual, testing parametric game knowledge.
    \item Symbols show changes from Pairwise, with a minimum change of one mission at that aggregation level.
\end{tablenotes}
\end{threeparttable}
\end{table}

\subsection{Single-Agent Results}
\label{sec:solo-player}

To test the effect of information asymmetry, we introduce a single agent ablation giving the model direct access to both the game and the manual. As shown in
\cref{tab:ablation-mod}, most models improve when given direct access to both the bomb and the manual, confirming that the communication channel itself is a source of error in the collaborative setting. 
Only Gemini is impacted negatively overall, suggesting that the structure of the role specialisation outweighs the overhead imposed by the information asymmetry.
At the module level, effects are mixed, creating two broad groups: modules where getting information directly from the manual helps, and modules where it does not, which we analyse in turn below.

\paragraph{Decisions no longer rely on second-hand information.} 
Performance increases across several modules, with most consistent improvements on the easier ones, \wires, \button, and \keypad.
In the collaborative setting, the Expert must incrementally guide the Defuser to collect specific details relevant to the module, such as batteries, indicators, or the serial number. 
In turn, the Defuser must communicate them accurately before the correct instruction can be identified.
A single agent, on the other hand, has complete situational awareness from the outset and can apply the rules directly without first establishing shared understanding. Additionally, as shown in \cref{sec:statics:manual-vqa}, models struggle to identify ambiguous or potentially hallucinated information from a partner. Removing the communication channel eliminates this error source entirely.

\paragraph{Shared references simplify gameplay.} 
Improvements on \keypad indicate that communication is especially costly when there is a need to negotiate a shared vocabulary with a partner. 
When we isolate the symbol-description step as a VQA-style task (more details in \cref{app:statics:keypad}), different models generate different descriptions for the same symbol on more than half of the questions.
For instance, \pashtote is described as `smiley face' by Gemini and GPT, but `magnet' by Qwen; the only symbol all models (implicitly) agree on is \hollowstar, which they all call a `star'.

\paragraph{A partner helps where state-tracking demands are high.}
For modules with multiple stages (\whosonfirst, \memory, \wiresequence), removing the partner appears to harm performance. 
\wiresequence provides the clearest illustration: tracking which wire colours have already appeared across up to twelve sequential steps is a working memory demand that direct manual access does not address---and one that is compounded by the need to integrate information across an increasingly long context. 
Performance on \simonsays and \maze remains near zero across all conditions, pointing to state-tracking as a capability gap that persists regardless of how much information access is simplified.

\subsection{Contamination Results}\label{sec:e2-parametric-knowledge}



While commonsense or prior knowledge from other domains does not directly apply to the problem-solving rules of the game, it is possible that, given the game's substantial online presence since its 2015 release, models have encountered \ktane{}-related content during pre-training.\footnote{We confirm this in \cref{app:how-do-you-do}, where models produce plausible game-specific responses without any provided context.}
Models are, in principle, able to memorise the defusal manual. Solutions are presented as branching decision trees---which is precisely the kind of structured, rule-based content that models memorise without difficulty \citep{Hartmann2023SoKMemorizationGeneralPurpose, Wang2024GeneralizationVsMemorization, Kiyomaru2024ComprehensiveAnalysisMemorization}. 
Although we explicitly prompt models to rely solely on the information provided (\cref{app:system_prompt}), this cannot---and does not---prevent them from drawing on what they already know.\footnote{We show one example of models explicitly using parametric knowledge in \cref{app:example:model-referring-to-param}.}
For transparency, we probe whether models take such shortcuts by treating the task as a recall problem rather than a collaborative one \citep{Hosseini2025SeeingWhatsNot, Jacovi2023StopUploadingTest, Cheng2025PositionBenchmarkingBroken, Hermann2024FoundationsShortcutLearning}.
We test whether they can solve any missions solely by relying on their parametric knowledge, that is, \textit{without access to a partner or the manual}.

\paragraph{All models use their knowledge of KTANE.}
The results in \cref{tab:ablation-mod} show that models attempting to solve a puzzle module without the manual outperform a random baseline.\footnote{The random baseline is computed by running an agent that selects uniformly at random from plausible actions at each step, averaged over 100 attempts per mission ($n=1{,}000$ per module).} This confirms that while it is possible to chance upon the correct solution for some module types, models generally possess and are using knowledge of the task from their training alone. 
However, the extent to which different models can leverage parametric knowledge varies. All models possess some parametric knowledge of both the \wires and \complicatedwires modules; GPT seems to possess sufficient parametric knowledge to solve \textit{all ten} \wires missions. 
In several cases, models perform better than in the collaborative setting. 
This suggests that models not only have sufficient parametric knowledge for these module types, but that the full collaborative setting introduces additional complexity that obscures parametric knowledge. 

\begin{wraptable}[16]{r}{0.34\textwidth}%
\vspace{-.2em}
\vspace{-\intextsep}
\footnotesize
\renewcommand{\arraystretch}{1.3}
\begin{threeparttable}
\caption{Expert procedural reasoning accuracy, with and without manual access.}
\label{tab:expert-vqa-manual-comparison}
\begin{tabularx}{\linewidth}{@{}X S[table-format=2.1] c @{}}
\toprule
 & {Manual} & {No Manual} \\
\midrule
\claude*~Sonnet 4.6 & 67.1 & 21.6\dectrii \\
\gemini*~Gemini 3 Flash & 64.1 & 36.5\dectrii \\
\openai*~GPT 5.2 & 61.5 & 19.6\dectrii \\
\internvl*~InternVL 3.5 & 44.5 & 22.0\dectrii \\
\qwen*~Qwen3.5 & 53.7 & 10.4\dectrii \\
\midrule[0.1ex]
\textit{Average} & 58.2 & 22.0\dectrii \\
\midrule[0.1ex]
\textit{Random}\tnote{\dag} & 21.3 & 21.3\notri \\
\bottomrule
\end{tabularx}
\begin{tablenotes}\scriptsize
\item[\dag] Based on the possible answer options for the multiple choice questions.
\end{tablenotes}
\end{threeparttable}
\end{wraptable}

\paragraph{Visual context scaffolds manual recall.} To disentangle the ability to recall knowledge of the manual acquired during training from the ability to accurately perceive and interact with the bomb, we return to the procedural reasoning questions from the offline tests introduced in \cref{sec:statics:manual-vqa} and evaluate models on the same questions without providing the manual. 
We focus on this question type as an indicator of the ability to deduce accurate instructions based solely on prior knowledge. 
As shown in \cref{tab:expert-vqa-manual-comparison}, almost all models perform at or below random chance level.
Gemini is the only exception showing meaningful performance. 
This stands in contrast to Defuser results, where models demonstrably apply parametric knowledge of module interactions. One plausible explanation for the asymmetry is that Defuser recall is scaffolded by visual context: seeing the module may act as a retrieval cue, making relevant actions easier to recover than the precise conditional logic of the manual, which must be retrieved without any perceptual anchor.\looseness=-1

\FloatBarrier
\section{Conclusion}

We introduce \gptnt, a framework that wraps the interactive 3D bomb defusal video game \ktane to provide a challenging testbed for multimodal, multi-agent collaboration in real time. \gptnt fills a gap in the existing MLLM evaluation landscape by being the first benchmarking environment to simultaneously require coordination under asymmetric information, asynchronous action under time pressure, visual grounding in a dynamic environment, and sustained multi-turn communication. This reflects an ecologically valid set of conditions that models could encounter in real-world deployment scenarios, making \gptnt an intentional step towards more holistic, integrated evaluation of MLLM skills.

We demonstrate that, unlike human players, current state-of-the-art models are unable to play a single successful game in the full setting, acting asynchronously on missions with multiple puzzle modules. Even the simplest setting we test, where models are given unlimited deliberation and generation time for less complex single-module missions, poses significant challenges for the models we test. These challenges are exacerbated in other-play, especially when models have to adapt to potentially weaker partners. 
Single-agent diagnostic evaluations reveal that, while models have been exposed to \ktane{}-specific data during training, the benchmark surfaces persistent limitations in both collaborative and low-level capabilities.
Across online and offline evaluations, we find that models struggle to act efficiently under a time budget, maintain an accurate state representation across turns, recognise when information is underspecified or hallucinated, and recover from mistakes.

The \gptnt environment is uniquely suited to stay ahead of growing model capabilities and mitigate the risk of the benchmark becoming saturated. The set of core missions we introduce in this iteration of the benchmark only covers relatively low difficulty levels (2--3 of 7, see \cref{app:ktane-difficulty}), and more challenging missions can easily be configured and procedurally generated at scale. Furthermore, new puzzle modules contributed by an active modding community can be leveraged to increase mission difficulty while simultaneously reducing the effect of contamination as training corpora expand. 



\subsection*{Future Work}

\begin{figure}[tb]
\centering
\includegraphics[width=1\linewidth]{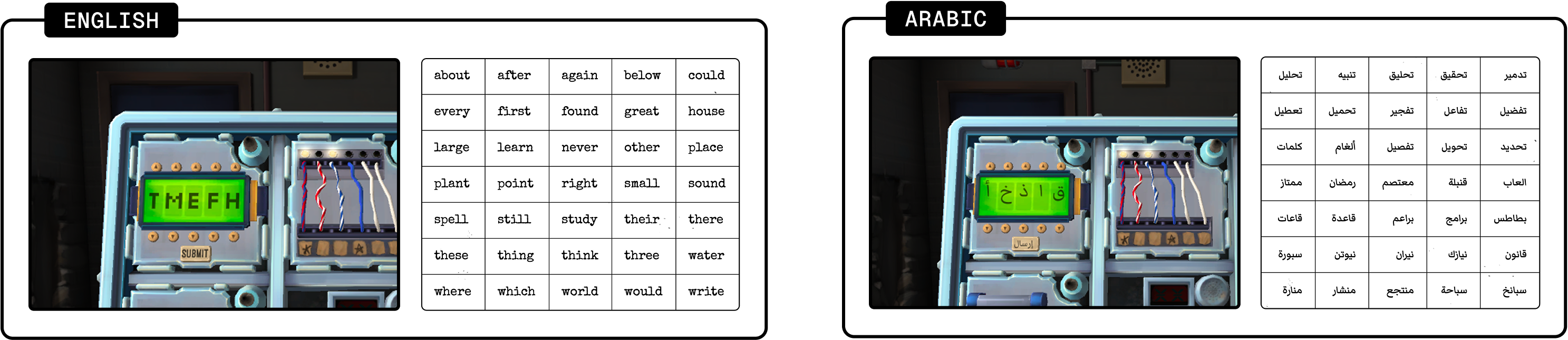}
\caption{Example comparing \passwords across English and Arabic in KTANE.}
\label{fig:en-vs-ar}
\end{figure}


\ktane provides the foundation for direct extensions of our benchmark across languages and modalities.
The game's availability in 27 different languages can be leveraged for cross-lingual evaluation of communicative grounding without framework changes (\cref{fig:en-vs-ar}).
The game supports so-called \textit{needy} modules, modules that are activated randomly during the game. These modules signal that they require immediate attention not only visually, but also by emitting audio cues, which is crucial when the Defuser is viewing a different side of the bomb at the time of activation. As they require audio understanding and the ability to (re)prioritise, bombs including needy modules represent a qualitatively harder difficulty tier.
In a similar vein, native speech support would remove text-play artefacts such as the Unicode \keypad workaround (see \cref{app:statics:keypad}), surface real-time collaboration phenomena that text suppresses, and
enable human--agent experiments that do not place the burden on the human players \citep{Cui2025RecentAdvancesSpeech,Chen2025TurnTakingSynchronousDialogue}.

Several behavioural phenomena we observe warrant further investigation. 
Models adapt their communication \textit{rate} to different models in other-play (\cref{sec:communication:partners}), but how \textit{content} shifts---vocabulary, specificity, repair strategies---remains uncharacterised. 
The cascade from the first strike to a second, and potentially a third (\cref{fig:outcome-sankey}) could reflect compounding errors, an inability to leverage a strike as a feedback signal, or both. 
A principled investigation into models' understanding and ability to leverage visual feedback more broadly is therefore warranted. 
Given that time pressure is a core dimension of the benchmark, models’ ability to accurately gauge remaining time and adapt their behaviour to perceived urgency merits attention in future work (see \cref{sec:app:time-effect} for qualitative examples).


As models improve, more complex missions containing a higher number of puzzle modules can be introduced. Such missions may benefit from scaling the number of Experts, allowing multiple players to divide and conquer the manual; with a single Defuser as the bottleneck, this setting would create additional coordination and communication challenges worth studying.

Beyond zero-shot evaluation of existing capabilities, \gptnt can serve as a training environment, as long as the end goal is to improve performance on different held-out collaborative tasks. 
However, whether training on \gptnt can yield such generalisable performance improvements is orthogonal to this work.

Allowing for the rules for solving the different module types to be randomised, rather than defaulting to those underlying the standard manual available online, could negate the effects of pre-training knowledge and counteract attempts at overfitting on the proposed suite of missions.




\section{Limitations}\label{sec:limitations}


\paragraph{Game Cost.}
Running \gptnt requires purchasing a copy of \textit{Keep Talking and Nobody Explodes}, available for \$14.99 from the Humble store\footnote{\url{https://humblebundle.com/store/keep-talking-and-nobody-explodes}} (covering Windows, macOS, and Linux). We do not distribute the game or any source code from it, as it would violate its licence. We do, however, release all infrastructure to run the benchmark with our mod.

\paragraph{API Cost.}
Reproducing the full experimental suite costs approximately \$3,000 in API spend, with each open-source model served locally on 2$\times$\;H100 80GB GPUs. 
Cost also shapes model coverage: hard filters on context length and available compute bias the evaluation toward large closed-source models or smaller open-source ones, with the middle ground excluded. 
The resulting open--closed comparison conflates family with scale; the performance gap most likely reflects model size rather than anything intrinsic to the open-source paradigm.
A breakdown of token usage and cost per model and experiment is provided in \cref{app:model-cost}.


\paragraph{Configuration.}
Resolution is fixed at 640$\times$480---the smallest the game supports before UI elements clip---to minimise context length and the generation latency that erodes real game time in asynchronous play. 
Although results on offline evaluations (\cref{sec:statics:manual-vqa,sec:statics:defuser}) show that models achieve strong performance on standalone reading probes, higher resolutions are worth exploring, as serving becomes faster and cheaper. Additionally, as coordinate grounding is currently unreliable (\cref{app:results:som-vs-coords}), we treat set-of-marks interaction as standard, though both are supported and the choice can be revisited as models improve (\cref{app:action_space:interactions}). 


\paragraph{Outputs.}
Communication is text-only, placing models in their strongest modality while diverging from how \ktane is naturally played. For instance, the Unicode workaround (\cref{app:statics:keypad}) observed for Keypad symbols is a text-play artefact that would not exist in spoken dialogue.
Output is capped at 1,000 tokens per turn (details in \cref{app:output-format}) and temperature is fixed at 0.6 across all models rather than tuned per model, to avoid conflating sampling configuration with capability (details in \cref{app:model:sampling-config}).

\paragraph{Evaluation.}
The evaluation is pass@1: models are given one attempt at each of the ten multi-module missions, and for single-module missions, we report the average over ten missions per module type. 
This provides a strict but comprehensive evaluation of the capabilities of current models across diverse missions, rather than of specific task instantiations.
We leave evaluating reliability across repeated attempts to future work \footnote{As a preliminary investigation, we ran Qwen3.5 for 50 independent attempts on the easiest mission (\#1) in asynchronous mode. None of the attempts was successful, with 35/50 ending in timeouts, preserving the overall trend we see with one attempt across models.}.
Results are point-in-time snapshots whose enduring value lies in the behavioural patterns they reveal. 


\paragraph{Human reference point.}
Our motivation was to establish whether agents---humans or AI---can play the game in real time, and not whether they can play the game in the same way. 
We allow models and humans to engage with the task in the most natural way for them, with access to the conditions they are most suited to. 
Our aim is not to make a direct comparison, which we understand is a common precedent in many tasks. 
However, even enforcing the same input and output modality for both humans and AI would not guarantee experimental parity for our task. For example, even when both humans and AI play the game using text, their action spaces are fundamentally different: humans act using keyboard and mouse, whereas AI generates JSON (and requiring humans to do the same would make the task harder for them). 
To use an analogy, if comparing the speed of a fish with the speed of a human: a human will win if they’re both on land, a fish will win if they’re in the water. 


\newpage

\subsubsection*{Acknowledgments}


We would like to thank Joe Gilligan, Godugu Anil Himam, Fraser Holman, Phoenix May, Suhaib Salim, Ethan Smyth, Rohan Veit, Szymon Wlodarczyk, and Tristan Xuan for their early contributions to this project, and to Sophie Higham, Riccardo Izzo, Joe Kelliher, Muhy Eddin Za'ter, and Georgios Pantazopoulos for their valuable feedback and insightful discussions. 

This work was supported by the Edinburgh International Data Facility (EIDF) and the Data-Driven Innovation Programme at the University of Edinburgh and the Heriot-Watt University high-performance computing facility (DMOG). Additional compute resources were provided through the Anthropic Student Builder Programme and the Google for Researchers Programme.
We are grateful to Jurriaan Hage and the academic support team within the School of Mathematical and Computer Sciences at Heriot-Watt University, and the research computing support team around Jose Manual Menendez Montes and Graeme Pollock for their continued support at Heriot-Watt University.\looseness=-1

This work was supported by EPSRC as part of PhD studentships in the Centre for Doctoral Training in Robotics and Autonomous Systems awarded to Sabrina McCallum and Malvina Nikandrou, and the School of Mathematical and Computer Sciences at Heriot-Watt University as part of a PhD studentship awarded to Amit Parekh.\looseness=-1

Lastly, we thank Steel Crate Games for creating \textit{Keep Talking and Nobody Explodes}.
The \gptnt mod builds on \texttt{KtaneTwitchPlays}, a community-maintained mod that exposes the game to external text-command control. We are grateful to its maintainers, and to the wider \ktane modding community, whose work made \gptnt possible.\looseness=-1


\FloatBarrier

\newpage
\bibliography{main}
\bibliographystyle{tmlr}

\appendix

\FloatBarrier
\clearpage

\section{Extended Related Work}\label{app:related-work}

\begin{table}[t]
\centering
\footnotesize
\renewcommand{\arraystretch}{1.4}
\caption{Definitions of key properties used to characterise multi-agent environments.}
\label{tab:related-work-defs}
\begin{tabularx}{\linewidth}{@{} l X @{}}
    \toprule
    Property & Definition \\
    \midrule
    Multi-Agent Environment  & Environment with two or more agents, where they have to work either towards a shared goal (Cooperative), or against one another (Competitive). \\
    Information Asymmetry    & When different agents have access to different knowledge or observations about the environment or task, creating information imbalances that affect decision-making and interactions. \\
    Asynchronous Action      & The agents act asynchronously, taking independent actions in parallel. This contrasts with synchronous systems, where agents must wait for their turn in a predetermined order. \\
    Dynamic Environment      & The environment changes independently of agent actions. Here, we do not consider state changes resulting from the actions of another agent. \\
    Real-Time Evaluation     & The environment changes while the agent is thinking. For example, a countdown timer that continues while the agent deliberates. \\
    Multi-Turn Communication & Use language to either describe the task or to coordinate the agents over multiple turns. \\
    Image Sequence           & Agent has to, at some point, deal with a stream of images. This might require integrating information across multiple images or understanding temporal differences. \\
    Long Context       & Agents need to retrieve information from long text and/or image inputs. \\
    \bottomrule
\end{tabularx}
\end{table}

\paragraph{General agentic benchmarks.}

The most foundational cluster of agentic benchmarks---AgentBench \citep{liu2024agentbench}, AgentBoard \citep{ma2024agentboard}, and VisualAgentBench \citep{liu2024visualagentbench}---establishes a baseline by evaluating agents across a range of interactive tasks, using breadth to offer a comprehensive picture of how agents generalise across qualitatively different interactive settings. 
But breadth is not the same as depth of interaction complexity: these benchmarks are uniformly single-agent, operating in largely static environments, and place no demand on agents to coordinate, communicate, or reason under information asymmetry. The interaction model they assume---one agent, one task, one episode---is a reasonable starting point, but it systematically excludes properties that only emerge when agents must work alongside others in a world that does not wait for them.



\paragraph{Video games as evaluation settings.}

Video games have long served as proving grounds for AI research.
From the Atari Learning Environment catalysing breakthroughs in deep reinforcement learning \citep{Bellemare2013ArcadeLearningEnvironment,Mnih2015HumanlevelControlDeep}, to StarCraft II \citep{vinyals2019grandmaster} and Dota 2 \citep{berner2019dota} exposing new frontiers in long-horizon strategic reasoning, games have consistently forced AI systems to confront temporal pressure, uncertainty, and emergent complexity. 
This tradition has naturally extended to MLLM evaluation through benchmarks spanning open-world exploration (MineDojo, \citealp{fan2022minedojo}), roguelikes and puzzle games---such as BALROG \citep{paglieri2024balrog}, Baba Is AI \citep{cloos2024baba}, COMMA \citep{ossowski2024comma}, and Overcooked-AI \citep{carroll2019utility, Gessler2024OvercookedV2RethinkingOvercooked}---as well as commercial titles \citep{Zhang2025VideoGameBenchCanVisionLanguage},
all providing the rich visual inputs, stochastic dynamics, and clear success criteria that general agentic benchmarks lack. 
A subtler concern is that several of these benchmarks use simplified or modified versions of their source games, which trade the complexity that made those environments compelling for more tractable problem spaces. 
A deeper structural issue also exists: all of these remain fundamentally single-agent settings. Even MindAgent \citep{gong-etal-2024-mindagent}, which introduces multi-agent coordination within a game environment, does so through a centralised planning architecture, thereby removing the information asymmetry and communication pressure that makes coordination difficult. The dynamism and visual richness that game environments bring to MLLM evaluation have not yet been placed in service of decentralised, cooperative, multi-agent interaction.

\paragraph{Embodied environments.}

Embodied simulation environments represent the most direct attempt to address what game-based benchmarks leave unresolved: placing agents together in visually rich, interactive worlds and asking them to cooperate on extended tasks. TEACh \citep{Padmakumar2021TEAChTaskdrivenEmbodied}, Watch-and-Help \citep{puig2020watchandhelp}, and the Alexa Arena \citep{Gao2023AlexaArenaUserCentric} come closest to this---in all three, one agent holds knowledge of the task goal that the other must act upon, creating a structural information asymmetry that makes natural language a necessary coordination channel. However, they all reveal the same limitation: the environment remains static between agent actions and does not stress agents across the long interaction traces that genuine cooperative tasks tend to produce---a dimension that matters given that MLLM performance is known to degrade towards the tail of long context windows \citep{wang2024needlemultimodalhaystack}. 
HAZARD \citep{zhou2024hazard} moves in a complementary direction by introducing natural environment dynamics---where the world evolves independently of what agents do---but does so in a single-agent setting, forfeiting the cooperative structure entirely. 
\citet{li2024embodied} takes yet another path, trading visual grounding for richer interaction modelling and operating on text-based state representation that bypasses the visual reasoning demands that real environments present. 
A pattern emerges across this cluster: each benchmark makes progress on one dimension by accepting a regression in another, and no single environment refuses these tradeoffs simultaneously.


\paragraph{Cooperative multi-agent settings with asymmetric information.}

The benchmarks that take information asymmetry most seriously as a design principle represent the most deliberate attempt to force genuine communication — directly addressing what might be called the \textit{knowing-doing gap} \citep{paglieri2024balrog}: in many multi-agent settings, agents can bypass language entirely and act autonomously, rendering communication ornamental rather than functional.
Information asymmetry structurally closes this gap---when each agent holds knowledge that is necessary but insufficient for task completion, natural language becomes the \textit{only} viable coordination channel. 
Competitive social deduction games such as Avalon \citep{light2023avalonbench} and Werewolf \citep{wu2024werewolf} operationalise this best, with hidden roles creating a communicative pressure under which agents must reason strategically over what others know and believe.
WhodunitBench \citep{xie2024whodunitbench} extends this with visual grounding over evidential images, adding an absent multimodal layer to purely text-based deduction.
But all three are fundamentally competitive---the communicative pressure they generate is adversarial, not cooperative, and the dynamics of strategic deception are quite different from those of collaborative problem-solving.
COMMA \citep{ossowski2024comma}, Multimodal CLEM 
\citep{hakimov-etal-2025-using}, and Concordia 
\citep{smith2024the} are all cooperative and preserve information asymmetry. 
COMMA and Multimodal CLEM both require visual understanding, though COMMA is the only one that requires visual grounding as part of the interaction. 
But even COMMA stops short of the full combination---agents receive static observations rather than image sequences across time, actions remain synchronised rather than asynchronous, and interaction traces stay well within standard context lengths.
The pattern from the previous cluster holds: each benchmark isolates and advances a subset of the required properties, and the specific combination---cooperative, visually grounded, dynamic, asymmetric, asynchronous, and long-horizon---remains unoccupied.

\paragraph{GPTNT positioning.}

\gptnt occupies this unaddressed intersection by grounding evaluation in Keep Talking and Nobody Explodes, a cooperative game whose structure instantiates all of these properties simultaneously. 
Information asymmetry is inherent to the game's design; neither agent can succeed without the other, which closes the knowing-doing gap structurally as language is the \textit{only} coordination mechanism. 
The bomb's countdown timer introduces natural environmental dynamics that persist regardless of agent deliberation, and the multi-module structure of each mission generates the extended interaction traces that stress long-context reasoning in ways that shorter episodes cannot. 

\section{About \textit{Keep Talking and Nobody Explodes}}

Keep Talking and Nobody Explodes (\ktane) is a cooperative puzzle video game built around a fundamental information asymmetry: one player sees the bomb but has no instructions, while the other holds the manual but cannot see the bomb. Players must bridge this gap through communication, working against a countdown timer and a limited strike budget. Its combination of procedural puzzle generation, role separation, and real-time pressure makes it a rich environment for studying collaborative behaviour under constraint.
This appendix describes the game's core elements as they are relevant to understanding the benchmark.

\begin{figure}[tbh]
\centering
\includegraphics[width=1\linewidth]{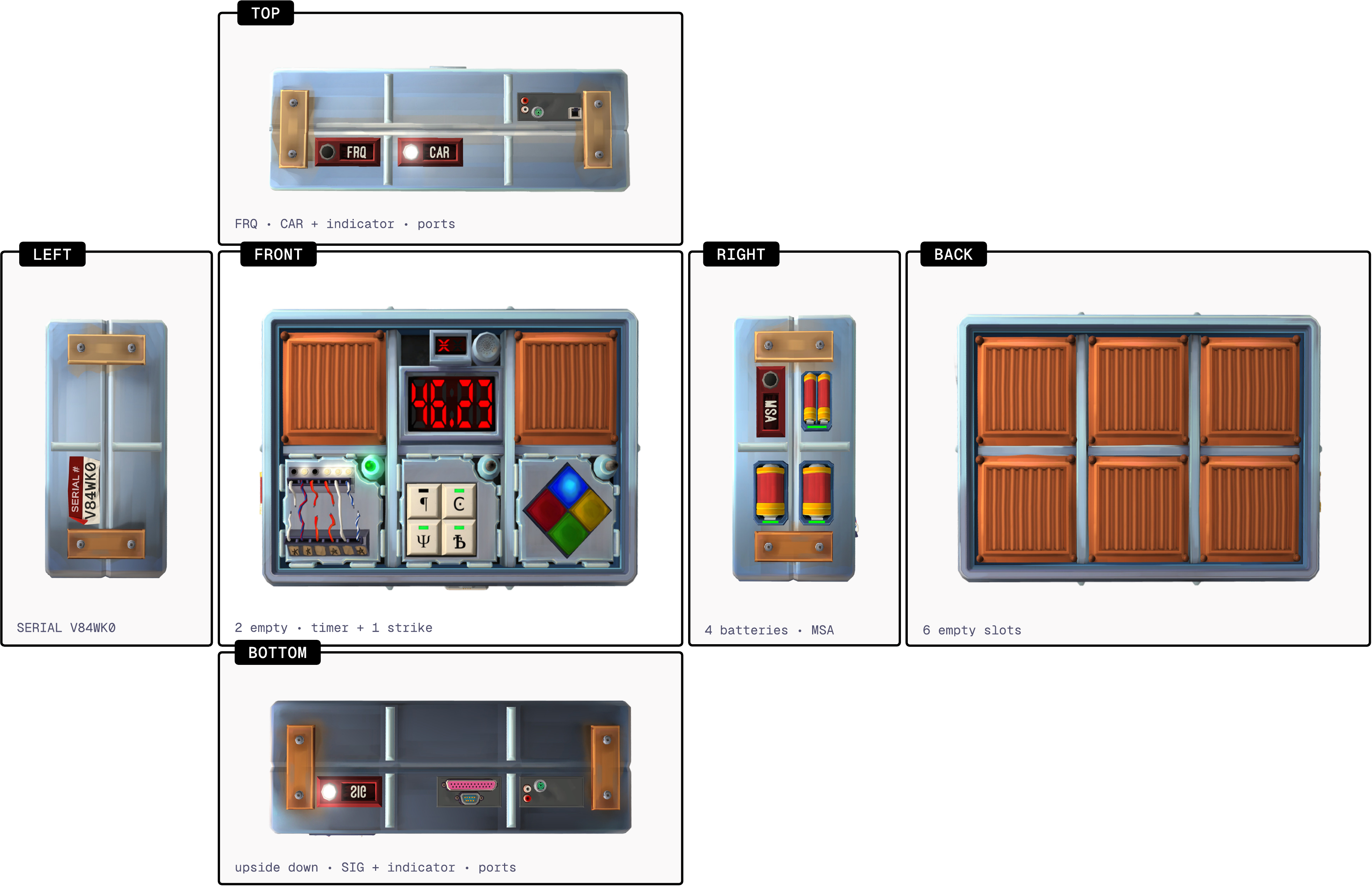}
\caption{An unfolded view of all six faces of the bomb, showing the timer, strikes, modules, and widgets. The Defuser observes one side at a time, navigating across sides to collect game-relevant information.}
\label{fig:exploded-layout}
\end{figure}

\subsection{Game Elements}\label{app:game-modules}
\paragraph{Manual.}
The official manual---available at \url{https://bombmanual.com}---contains detailed instructions on how to solve each module, including illustrations to aid the Expert in their task as they cannot see the game.

\paragraph{Bomb Layout.}
As shown in \cref{fig:exploded-layout}, each bomb consists of several module slots on the larger front and back faces, one of which is designated for the \introduce{Timer}, while the remaining slots can be filled with one or more modules of different types.
The sides contain widgets, such as the serial number, batteries, ports and light indicators, which provide contextual information needed to solve several modules.

\begin{figure}[htb]
    \centering

    \begin{subfigure}[t]{0.30\textwidth}
        \centering
        \begin{tikzpicture}
            \clip [rounded corners=5pt] (0,0) rectangle (0.8\textwidth, 0.8\textwidth);
            \node[anchor=south west, inner sep=0] at (0,0)
                {\includegraphics[width=0.8\textwidth]{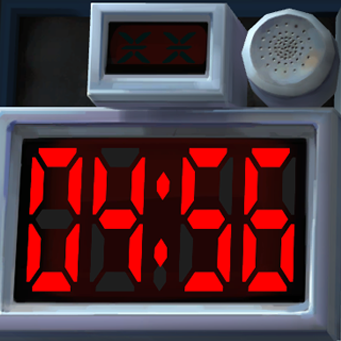}};
        \end{tikzpicture}
        \caption{No Strikes}
        \label{fig:timer-no-strikes}
    \end{subfigure}%
    \hspace{\fill}
    \begin{subfigure}[t]{0.30\textwidth}
        \centering
        \begin{tikzpicture}
            \clip [rounded corners=5pt] (0,0) rectangle (0.8\textwidth, 0.8\textwidth);
            \node[anchor=south west, inner sep=0] at (0,0)
                {\includegraphics[width=0.8\textwidth]{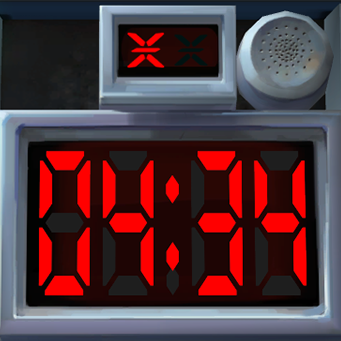}};
        \end{tikzpicture}
        \caption{One Strike}
        \label{fig:timer-one-strike}
    \end{subfigure}%
    \hspace{\fill}
    \begin{subfigure}[t]{0.30\textwidth}
        \centering
        \begin{tikzpicture}
            \clip [rounded corners=5pt] (0,0) rectangle (0.8\textwidth, 0.8\textwidth);
            \node[anchor=south west, inner sep=0] at (0,0)
                {\includegraphics[width=0.8\textwidth]{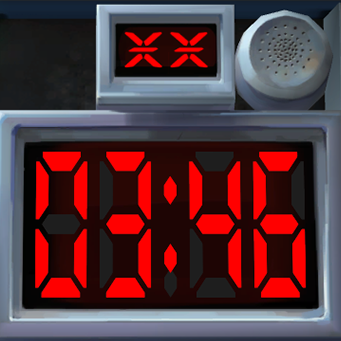}};
        \end{tikzpicture}
        \caption{Two Strikes}
        \label{fig:timer-two-strike}
    \end{subfigure}%

    \caption{Strikes on the timer unambiguously show the Defuser the number of remaining attempts during the game.}
    \label{fig:timer-strikes}
\end{figure}

\paragraph{Time and Strikes.}
The \textit{Timer} shows the remaining time to defuse the bomb in \texttt{MM:SS} format, switching to \texttt{SS.ms} during the final minute.
As shown in \cref{fig:timer-strikes}, there is an LED display above the timer that shows the number of strikes that have occurred, indicating how many attempts the Defuser has left. Importantly, for each strike received, the timer speeds up.

\begin{figure}[htb]
    \centering

    \begin{subfigure}[t]{0.5\textwidth}
        \centering
        \begin{tikzpicture}
            \clip[rounded corners=5pt] (0,0) rectangle (0.6\textwidth, 0.6\textwidth);
            \node[anchor=south west, inner sep=0] at (0,0) {
                \includegraphics[width=0.6\textwidth]{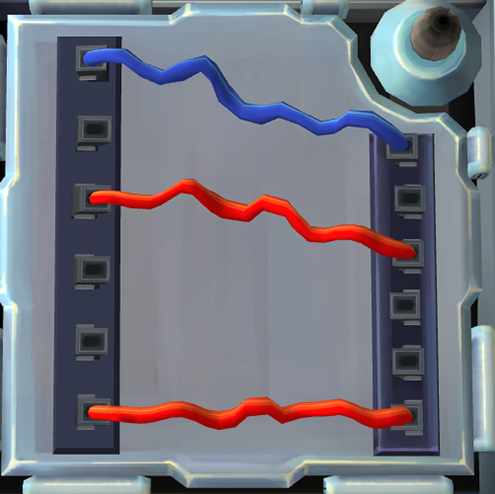}
            };
        \end{tikzpicture}
        \caption{Unsolved module}
        \label{fig:module-not-solved}
    \end{subfigure}%
    \hfill
    \begin{subfigure}[t]{0.5\textwidth}
        \centering
        \begin{tikzpicture}
            \clip[rounded corners=5pt] (0,0) rectangle (0.6\textwidth, 0.6\textwidth);
            \node[anchor=south west, inner sep=0] at (0,0) {
                \includegraphics[width=0.6\textwidth]{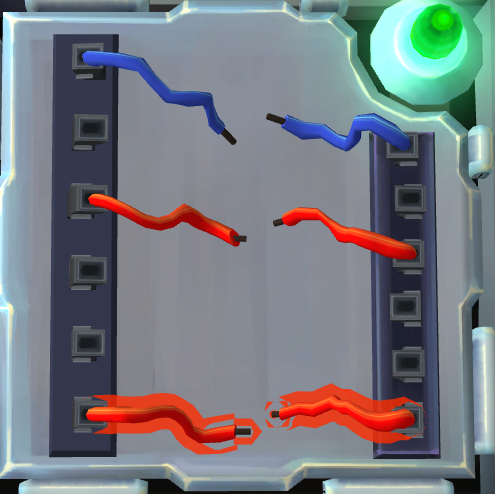}
            };
        \end{tikzpicture}
        \caption{Solved module}
        \label{fig:module-solved}
    \end{subfigure}
    \caption{The LED light in the top right of each module glows green when the module is defused.}
    \label{fig:module-solved-led}
\end{figure}

\paragraph{When is a module solved?}
As shown in \cref{fig:module-solved-led}, there is an LED light located at the top-right corner of each module. When solved, it will glow green.
If the Defuser makes a mistake while attempting to defuse the bomb, the LED on the corresponding module will \textit{briefly} flash red to indicate that a strike is recorded, and a strike will permanently appear above the clock.

\FloatBarrier
\subsection{Individual Modules}\label{app:ktane:modules}








\noindent
\begin{minipage}[t]{0.2\textwidth}
    \vspace{0pt}
    \centering
    \begin{tikzpicture}
      \clip [rounded corners=5pt] (0,0) rectangle (\textwidth, \textwidth);
      \node[anchor=south west, inner sep=0] at (0,0)
        {\includegraphics[width=\textwidth,height=\textwidth]{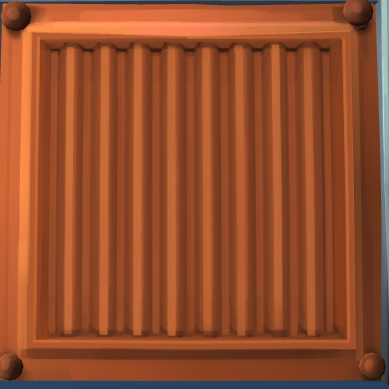}};
    \end{tikzpicture}
\end{minipage}
\hfill
\begin{minipage}[t]{0.77\textwidth}
    \vspace{0pt}
    \textbf{\textit{Empty}} \\
    \textbf{Stages:} 0 \\
    \textbf{Difficulty:} N/A---Requires zero player input; difficulty is purely visual.
    \vspace{2mm} \\
    The Empty module is not a puzzle for the bomb to solve. It is used as a placeholder because a given module slot might not contain any modules. For all intents and purposes, this option is considered during the sampling; however, it requires no input from the player and can be considered as ``automatically solved''.
\end{minipage}

\vspace{8mm}

\noindent
\begin{minipage}[t]{0.2\textwidth}
    \vspace{0pt}
    \centering
    \begin{tikzpicture}
      \clip [rounded corners=5pt] (0,0) rectangle (\textwidth, \textwidth);
      \node[anchor=south west, inner sep=0] at (0,0)
        {\includegraphics[width=\textwidth,height=\textwidth]{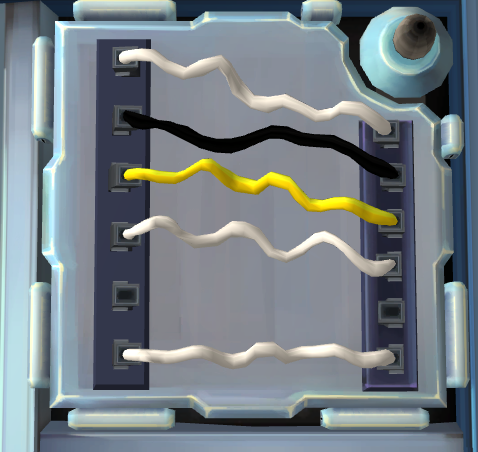}};
    \end{tikzpicture}
\end{minipage}
\hfill
\begin{minipage}[t]{0.77\textwidth}
    \vspace{0pt}
    \textbf{\wires} \\
    \textbf{Widgets Needed:} Serial Number \\
    \textbf{Stages:} 1 \\
    \textbf{Difficulty:} Relies on identifying easily communicated details (colour, count, order) with straightforward instructions.
    \vspace{2mm} \\
    For the Wires module, the Defuser needs to cut the correct wire. The correct wire depends on several key variables, including the total number of wires, each colour and specific position, the frequency of each colour, and the serial number of the bomb.\looseness=-1
\end{minipage}

\vspace{8mm}

\noindent
\begin{minipage}[t]{0.2\textwidth}
    \vspace{0pt}
    \centering
    \begin{tikzpicture}
      \clip [rounded corners=5pt] (0,0) rectangle (\textwidth, \textwidth);
      \node[anchor=south west, inner sep=0] at (0,0)
        {\includegraphics[width=\textwidth,height=\textwidth]{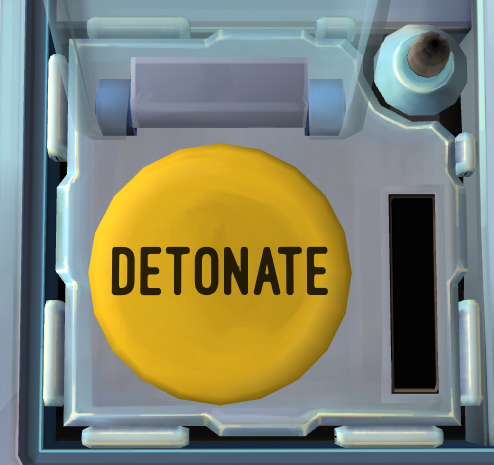}};
    \end{tikzpicture}
\end{minipage}
\hfill
\begin{minipage}[t]{0.77\textwidth}
    \vspace{0pt}
    \textbf{\button} \\
    \textbf{Widgets Needed:} Indicators, Batteries \\
    \textbf{Stages:} 1 \\
    \textbf{Difficulty:} Adds a layer of temporal difficulty; the Defuser might need to wait for the right moment to solve the module.
    \vspace{2mm} \\
    The Button module is a large button with a label and an LED strip. The Defuser is expected to press and release according to various factors including the colour, word, widget information, and the colour of the LED strip when the button is held down.
\end{minipage}

\vspace{8mm}

\noindent
\begin{minipage}[t]{0.2\textwidth}
    \vspace{0pt}
    \centering
    \begin{tikzpicture}
      \clip [rounded corners=5pt] (0,0) rectangle (\textwidth, \textwidth);
      \node[anchor=south west, inner sep=0] at (0,0)
        {\includegraphics[width=\textwidth,height=\textwidth]{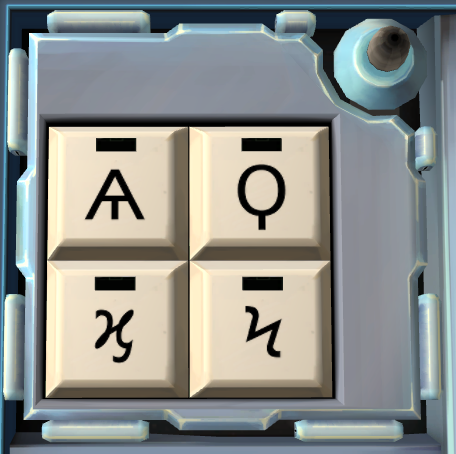}};
    \end{tikzpicture}
\end{minipage}
\hfill
\begin{minipage}[t]{0.77\textwidth}
    \vspace{0pt}
    \textbf{\keypad} \\
    \textbf{Widgets Needed:} None \\
    \textbf{Stages:} 1 \\
    \textbf{Difficulty:} Requires building a ``shared understanding'' of abstract symbols that are not common knowledge.
    \vspace{2mm} \\
    Contains four buttons, each with a symbol. To solve, the Defuser must press them in a specific order based on manual lists. The primary challenge is describing symbols accurately between players.
\end{minipage}

\vspace{8mm}

\noindent
\begin{minipage}[t]{0.2\textwidth}
    \vspace{0pt}
    \centering
    \begin{tikzpicture}
      \clip [rounded corners=5pt] (0,0) rectangle (\textwidth, \textwidth);
      \node[anchor=south west, inner sep=0] at (0,0)
        {\includegraphics[width=\textwidth,height=\textwidth]{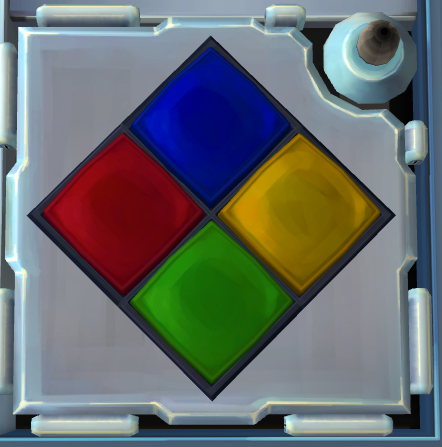}};
    \end{tikzpicture}
\end{minipage}
\hfill
\begin{minipage}[t]{0.77\textwidth}
    \vspace{0pt}
    \textbf{\simonsays} \\
    \textbf{Widgets Needed:} Serial Number, Strike Count \\
    \textbf{Stages:} 5 \\
    \textbf{Difficulty:} Complexity is derived from state tracking. 
    \vspace{2mm} \\
    Consists of four coloured buttons, one of which flashes. Players must gather details about the flashing colour and current strike count to follow the correct button sequence.\looseness=-1
\end{minipage}

\vspace{8mm}

\noindent
\begin{minipage}[t]{0.2\textwidth}
    \vspace{0pt}
    \centering
    \begin{tikzpicture}
      \clip [rounded corners=5pt] (0,0) rectangle (\textwidth, \textwidth);
      \node[anchor=south west, inner sep=0] at (0,0)
        {\includegraphics[width=\textwidth,height=\textwidth]{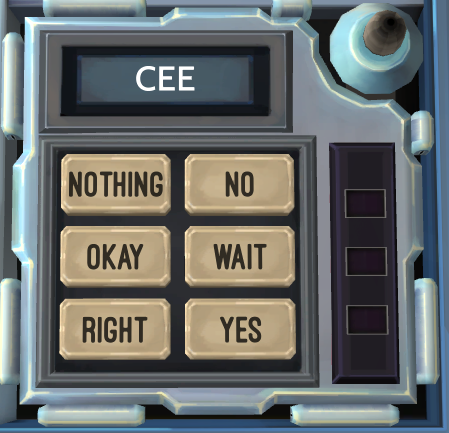}};
    \end{tikzpicture}
\end{minipage}
\hfill
\begin{minipage}[t]{0.77\textwidth}
    \vspace{0pt}
    \textbf{\whosonfirst} \\
    \textbf{Needs Side Info:} False \\
    \textbf{Stages:} 3 \\
    \textbf{Difficulty:} Relies on phonetic confusion (homophones like ``there/their/they're''). This is harder for human vocalisation than text.
    \vspace{2mm} \\
    The Expert must identify labels based on displayed words and then find the first word in a list that appears on the buttons.
\end{minipage}

\vspace{8mm}

\noindent
\begin{minipage}[t]{0.2\textwidth}
    \vspace{0pt}
    \centering
    \begin{tikzpicture}
      \clip [rounded corners=5pt] (0,0) rectangle (\textwidth, \textwidth);
      \node[anchor=south west, inner sep=0] at (0,0)
        {\includegraphics[width=\textwidth,height=\textwidth]{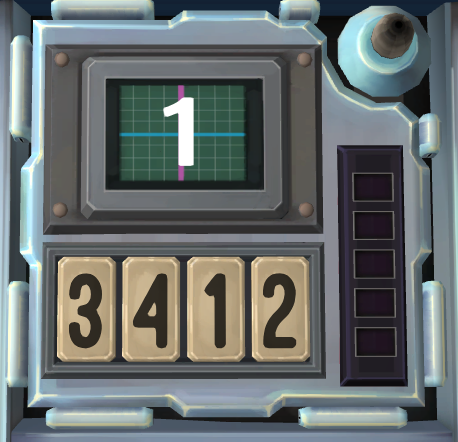}};
    \end{tikzpicture}
\end{minipage}
\hfill
\begin{minipage}[t]{0.77\textwidth}
    \vspace{0pt}
    \textbf{\memory} \\
    \textbf{Needs Side Info:} False \\
    \textbf{Stages:} 5 \\
    \textbf{Difficulty:} Requires high cognitive load to maintain an ``intrinsic state'' across five panels with shuffled positions.
    \vspace{2mm} \\
    After each correct press, buttons shuffle. The Expert must remember both the numbers and their corresponding positions from every previous stage to proceed.
\end{minipage}

\vspace{8mm}

\noindent
\begin{minipage}[t]{0.2\textwidth}
    \vspace{0pt}
    \centering
    \begin{tikzpicture}
      \clip [rounded corners=5pt] (0,0) rectangle (\textwidth, \textwidth);
      \node[anchor=south west, inner sep=0] at (0,0)
        {\includegraphics[width=\textwidth,height=\textwidth]{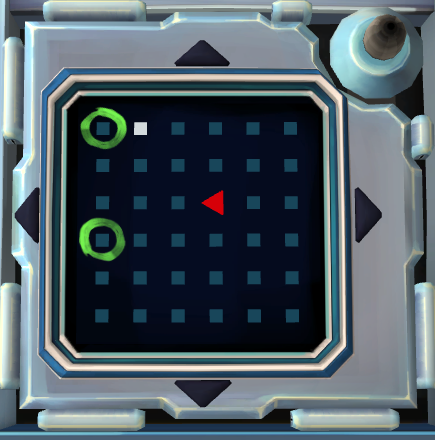}};
    \end{tikzpicture}
\end{minipage}
\hfill
\begin{minipage}[t]{0.77\textwidth}
    \vspace{0pt}
    \textbf{\maze} \\
    \textbf{Needs Side Info:} False \\
    \textbf{Stages:} 1 \\
    \textbf{Difficulty:} Requires spatial reasoning and the ability for players to maintain an intrinsic state of a $6\times6$ coordinate map.
    \vspace{2mm} \\
    The Expert must identify the correct map based on two green points and guide the Defuser to a red triangle without crossing hidden walls.
\end{minipage}

\vspace{8mm}

\noindent
\begin{minipage}[t]{0.2\textwidth}
    \vspace{0pt}
    \centering
    \begin{tikzpicture}
      \clip [rounded corners=5pt] (0,0) rectangle (\textwidth, \textwidth);
      \node[anchor=south west, inner sep=0] at (0,0)
        {\includegraphics[width=\textwidth,height=\textwidth]{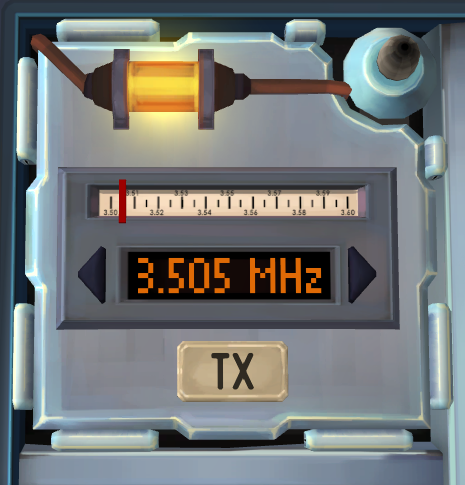}};
    \end{tikzpicture}
\end{minipage}
\hfill
\begin{minipage}[t]{0.77\textwidth}
    \vspace{0pt}
    \textbf{\morsecode} \\
    \textbf{Needs Side Info:} False \\
    \textbf{Stages:} 1 \\
    \textbf{Difficulty:} Requires the Defuser to accurately interpret and communicate a sequence of timing-based blinks.
    \vspace{2mm} \\
    The Expert must decode the Morse code to identify a word and transmit the frequency. It is difficult due to the high risk of communication error regarding the blinks.
\end{minipage}

\vspace{8mm}

\noindent
\begin{minipage}[t]{0.2\textwidth}
    \vspace{0pt}
    \centering
    \begin{tikzpicture}
      \clip [rounded corners=5pt] (0,0) rectangle (\textwidth, \textwidth);
      \node[anchor=south west, inner sep=0] at (0,0)
        {\includegraphics[width=\textwidth,height=\textwidth]{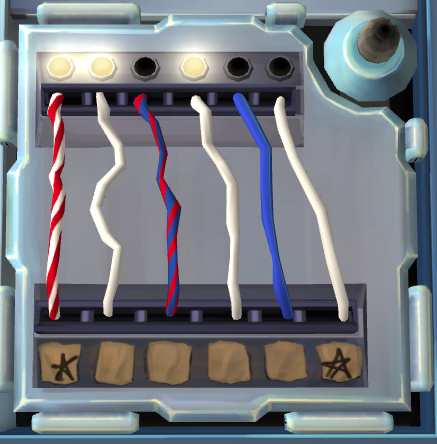}};
    \end{tikzpicture}
\end{minipage}
\hfill
\begin{minipage}[t]{0.77\textwidth}
    \vspace{0pt}
    \textbf{\complicatedwires} \\
    \textbf{Needs Side Info:} Serial, Batteries, Parallel Port \\
    \textbf{Stages:} 1 \\
    \textbf{Difficulty:} Visually complex Venn diagram instructions increase the Expert's cognitive burden compared to other tasks.
    \vspace{2mm} \\
    Requires determining whether to cut wires based on LEDs, star symbols, and bomb-wide state. The Expert's task is much more difficult than the Defuser's.
\end{minipage}

\vspace{8mm}

\noindent
\begin{minipage}[t]{0.2\textwidth}
    \vspace{0pt}
    \centering
    \begin{tikzpicture}
      \clip [rounded corners=5pt] (0,0) rectangle (\textwidth, \textwidth);
      \node[anchor=south west, inner sep=0] at (0,0)
        {\includegraphics[width=\textwidth,height=\textwidth]{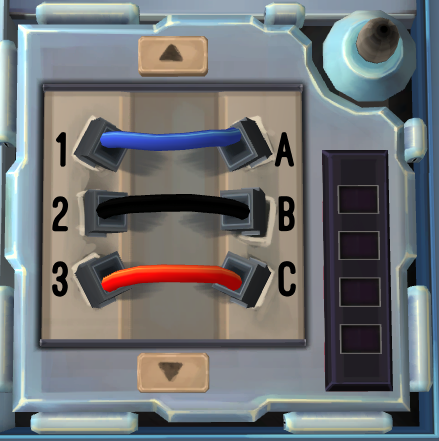}};
    \end{tikzpicture}
\end{minipage}
\hfill
\begin{minipage}[t]{0.77\textwidth}
    \vspace{0pt}
    \textbf{\wiresequence} \\
    \textbf{Needs Side Info:} False \\
    \textbf{Stages:} 4 \\
    \textbf{Difficulty:} Players must track a cumulative count of wire colours across multiple sequential panels without losing track.
    \vspace{2mm} \\
    The Expert determines whether to cut based on colour and its cumulative appearance count. Tracking the count is the primary source of difficulty.
\end{minipage}

\vspace{8mm}

\noindent
\begin{minipage}[t]{0.2\textwidth}
    \vspace{0pt}
    \centering
    \begin{tikzpicture}
      \clip [rounded corners=5pt] (0,0) rectangle (\textwidth, \textwidth);
      \node[anchor=south west, inner sep=0] at (0,0)
        {\includegraphics[width=\textwidth,height=\textwidth]{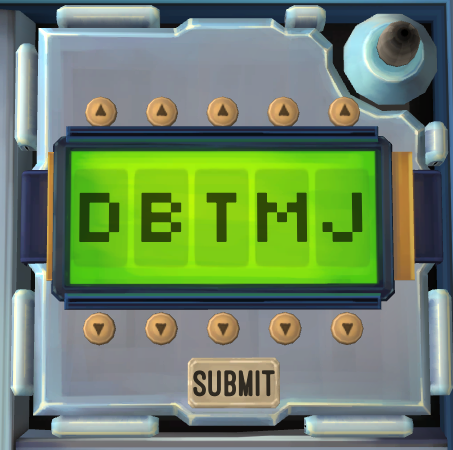}};
    \end{tikzpicture}
\end{minipage}
\hfill
\begin{minipage}[t]{0.77\textwidth}
    \vspace{0pt}
    \textbf{\passwords} \\
    \textbf{Needs Side Info:} False \\
    \textbf{Stages:} 1 \\
    \textbf{Difficulty:} High potential for ``search space'' error; players must cycle through multiple letters to find a matching word.
    \vspace{2mm} \\
    Each slot contains six possible letters. The Defuser must cycle through each slot to form a correct word found in the manual.
\end{minipage}

\FloatBarrier
\subsection{Multi-Module Missions}\label{app:ktane-difficulty}

\ktane missions get progressively harder as players play the game. For our multi-module missions, we sample modules and set time limits according to the rules for levels 2.1--3.3 in the wiki.\footnote{\url{https://ktane.fandom.com/wiki/Missions}} \cref{tab:mission-ktane-levels} maps our missions to the corresponding level, and highlights any deliberate deviations from the rules.
Across missions, module types are chosen from pools of like-difficulty, grouped as follows: 

\begin{itemize}[leftmargin=*]
    \item \textbf{Easier:} \wires, \button, \keypad (introduced in Level 1)
    \item \textbf{Intermediate:} \simonsays, \whosonfirst, \memory, \maze (introduced in Level 2)
    \item \textbf{Challenging:} \morsecode, \complicatedwires, \wiresequence, \passwords (introduced in Level 3)\looseness=-1
\end{itemize}

\begin{table*}[tbh]\footnotesize
\renewcommand{\arraystretch}{1}
\caption{Mapping between multi-module missions and corresponding \ktane mission levels, with deliberate deviations indicated in \textbf{bold}.}
\label{tab:mission-ktane-levels}
\begin{tabularx}{\linewidth}{@{}r l c X c @{}}
\toprule
\#
& Modules
& Level
& Description (\textbf{bold}=deliberate deviations)
& {$t(s)$} \\
\midrule

\multirow{3}{*}{1}
& {\wires} & \multirow{3}{*}{2.1}
& \multirow{3}{=}{3 modules\newline 2\(\times\) \keypad/\,\button/\,\wires\newline 1\(\times\) \maze/\,\simonsays/\,\memory} & \multirow{3}{*}{300} \\
& {\keypad} & & & \\
& {\maze} & & & \\
\midrule[0.1pt]

\multirow{4}{*}{2}
& {\wires} & \multirow{4}{*}{2.1}
& \multirow{4}{=}{\textbf{4} modules (default: 3)\newline \textbf{3}\(\times\) \keypad/\,\button/\,\wires (default: 2\(\times\))\newline 1\(\times\) \maze/\,\simonsays/\,\memory} & \multirow{4}{*}{300} \\
& {\button} & & & \\
& {\keypad} & & & \\
& {\memory} & & & \\
\midrule[0.1pt]

\multirow{4}{*}{3}
& \wires & \multirow{4}{*}{2.3}
& \multirow{4}{=}{4 modules\newline 2\(\times\) \keypad/\,\button/\,\wires\newline 1\(\times\) \maze/\,\simonsays\newline 1\(\times\) \memory/\,\whosonfirst} & \multirow{4}{*}{300} \\
& \keypad & & & \\
& \simonsays & & & \\
& \whosonfirst & & & \\
\midrule[0.1pt]

\multirow{3}{*}{4}
& \keypad & \multirow{3}{*}{2.4}
& \multirow{9}{=}{3 modules\newline 1\(\times\) \keypad/\,\button/\,\wires\newline 2\(\times\) \maze/\,\simonsays/\,\memory/\,\whosonfirst} & \multirow{3}{*}{180} \\
& \simonsays & & & \\
& \whosonfirst & & & \\
\cmidrule[0.1pt](r){1-3}\cmidrule[0.1pt](l){5-5}

\multirow{3}{*}{5}
& \wires & \multirow{3}{*}{2.4} & & \multirow{3}{*}{180} \\
& \memory & & & \\
& \maze & & & \\
\cmidrule[0.1pt](r){1-3}\cmidrule[0.1pt](l){5-5}

\multirow{3}{*}{6}
& \button & \multirow{3}{*}{2.4} & & \multirow{3}{*}{180} \\
& \whosonfirst & & & \\
& \memory & & & \\
\midrule[0.1pt]

\multirow{4}{*}{7}
& \multirow{4}{*}{\shortstack[l]{\whosonfirst\\ \memory\\ \passwords}}
& \multirow{4}{*}{3.1}
& \multirow{4}{=}{3 modules\newline 1\(\times\) \morsecode/\,\passwords\newline%
\textbf{2}\(\times\) \simonsays/\,\whosonfirst/\,\memory (default: 1\(\times\)) \newline%
\textbf{0}$\times$ \button/\,\wires (default: 1\(\times\))} & \multirow{4}{*}{300} \\
& & & & \\
& & & & \\
& & & & \\
\midrule[0.1pt]

\multirow{4}{*}{8}
& \multirow{4}{*}{\shortstack[l]{\button\\ \simonsays\\ \complicatedwires}}
& \multirow{4}{*}{3.2}
& \multirow{4}{=}{3 modules\newline 1\(\times\) \textbf{\button} (default: \wires)\newline 1\(\times\) \complicatedwires/\,\wiresequence\newline 1\(\times\) \simonsays/\,\memory} & \multirow{4}{*}{300} \\
& & & & \\
& & & & \\
& & & & \\
\midrule[0.1pt]

\multirow{4}{*}{9}
& \wires & \multirow{4}{*}{3.3}
& \multirow{4}{=}{4 modules\newline 1\(\times\) \textbf{\wires} (default: \keypad/\,\button)\newline 1\(\times\) \maze/\,\simonsays/\,\memory/\,\whosonfirst\newline 2\(\times\) \morsecode/\,\passwords/\,\complicatedwires/\,\wiresequence} & \multirow{4}{*}{300} \\
& \whosonfirst & & & \\
& \morsecode & & & \\
& \complicatedwires & & & \\
\midrule[0.1pt]

\multirow{4}{*}{10}
& \keypad & \multirow{4}{*}{3.3}
& \multirow{4}{=}{4 modules\newline 1\(\times\) \keypad/\,\button\newline 1\(\times\) \maze/\,\simonsays/\,\memory/\,\whosonfirst\newline 2\(\times\) \morsecode/\,\passwords/\,\complicatedwires/\,\wiresequence} & \multirow{4}{*}{300} \\
& \wiresequence & & & \\
& \maze & & & \\
& \passwords & & & \\

\bottomrule
\end{tabularx}
\end{table*}

\FloatBarrier

\clearpage
\subsection{Handling Time}\label{app:time}






\subsubsection{Determining the Synchronous Action Interval}
\label{app:why-3-seconds-after-act}

\begin{wrapfigure}[18]{r}{0.3\textwidth}
    \vspace{-\intextsep}
    \centering
    \includegraphics[width=\linewidth]{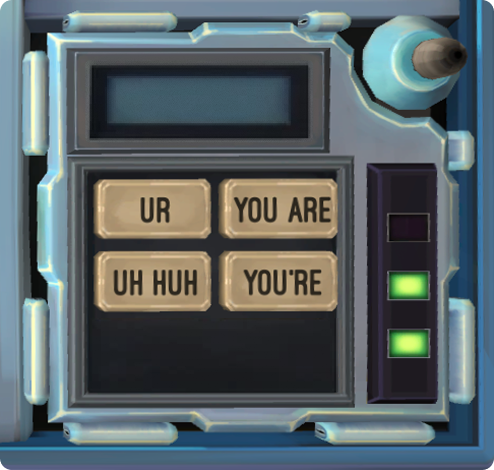}
    \caption{Example of \whosonfirst while the inter-stage animation is in progress}
    \label{fig:whosonfirst-mid}
\end{wrapfigure}
In-game animations in \ktane take up to 3 seconds to complete after an action. For example, \whosonfirst introduces a delay between clicking a
module element and the next stage becoming fully observable: pausing mid-animation produces an incomplete observation (see \cref{fig:whosonfirst-mid}) that a model cannot reliably act on, and recovering from one costs a further turn spent re-observing. 
Therefore, capturing observations earlier does not save game time; it merely converts a fixed cost into a stochastic one, which falls most heavily on multi-stage modules with inter-stage animations.

In the \textit{synchronous} setting, we resume the game for 3 seconds after every action, ensuring all state transitions are fully resolved before the next observation is captured. The pause takes effect at the next engine frame, so the realised window falls between 3 and 3.5 seconds. This is a property of the environment and independent of the model under evaluation; we refer to the nominal 3-second window throughout.
In the \textit{asynchronous} setting, no such buffer is needed: the round-trip time of a model forward pass often exceeds 3 seconds, so animations resolve naturally. 
If future models become fast enough to act mid-animation, the resulting incomplete observations become part of the challenge---mirroring how a human player must cope with acting faster than the game can update.

\subsubsection{Calculating Time Limits for Single-Module Missions}\label{app:time:turns}

For single-module missions, which are not from \ktane itself, we compute the time limit heuristically based on the estimated maximum number of turns required to complete each module.
The total count per mission is determined systematically by aggregating the necessary direct module interactions, bomb rotations, communication steps, and miscellaneous overhead. 
To calculate the time limit in seconds, we multiply the estimated maximum number of turns by the action interval (3 sec). 
We compute this dynamically for each mission, as outlined by \cref{tab:turn_cost_breakdown}.
\cref{tab:ktane_module_total_actions} demonstrates the computed turn allowance for each module, assuming a minimum of one stage update and standard front-facing placement. 


\begin{table}[htb!]
\centering
\footnotesize
\renewcommand{\arraystretch}{1.4}
\caption{Turn cost factors used to compute the total time limit per mission.}
\label{tab:turn_cost_breakdown}
\begin{tabular}{@{}lll@{}}
\toprule
Factor & Turns & Description \\
\midrule
Module interactions  & $\sum$ max actions        
    & Total interaction steps per module; per-module values given in 
      \cref{tab:ktane_module_total_actions} \\
Stage updates        & $\sum$ stages             
    & One turn per stage update per module \\
Bomb rotations       & $R \times 8$              
    & $R$ is the total side checks required across all modules, 
      clamped to $[1, 4]$ \\
Back placement       & $8$                       
    & Added only when back placement is permitted \\
Zooming              & $2(M + R)$                
    & One zoom in and out per module $M$ and per rotation $R$ \\
Strike buffer        & $3$                       
    & Fixed allowance for up to 3 strikes \\
Dialogue allowance   & $10$                      
    & Fixed baseline for communication overhead \\
\midrule
\textbf{Total (seconds)} & \textbf{Sum} $\times$ \textbf{3}  
    & 3 seconds per action \\
\bottomrule
\end{tabular}
\end{table}

\begin{table}[htb]
\centering
\footnotesize
\caption{Breakdown of per-module time limits.  Side info denotes the number of widget observations the Defuser must convey; Stages corresponds to the number of sequential phases of a module; Max actions is the largest number of interactions a single solution can require; Total turns aggregates the maximum number of actions and messages.}
\label{tab:ktane_module_total_actions}
\renewcommand{\arraystretch}{1.4}
\begin{tabular}{@{}l S[table-format=1] S[table-format=1] S[table-format=2] S[table-format=2]S[table-format=3]@{}}
\toprule
Module type & {Side info} & {Stages} & {Max actions} & {Total turns} & {Time limit (s)} \\
\midrule
\wires              & 1 & 1 & 1  & 25 & 75 \\
\button             & 2 & 1 & 14 & 48 & 144 \\
\keypad             & 0 & 1 & 4  & 18 & 54 \\
\simonsays         & 1 & 5 & 15 & 43 & 129 \\
\whosonfirst     & 0 & 3 & 3  & 19 & 57 \\
\memory             & 0 & 5 & 5  & 23 & 69 \\
\maze               & 0 & 1 & 35 & 49 & 147 \\
\morsecode         & 0 & 1 & 22 & 36 & 108 \\
\complicatedwires  & 2 & 1 & 6  & 40 & 120 \\
\wiresequence      & 0 & 4 & 16 & 33 & 99 \\
\passwords           & 0 & 1 & 51 & 65 & 195 \\
\bottomrule
\end{tabular}
\end{table}

\FloatBarrier
\section{The GPTNT Framework}\label{app:framework}

\gptnt is a purpose-built microservice framework providing LLMs with a complete programmatic interface to \textit{Keep Talking and Nobody Explodes}. 
\ktane offers no native API, no visual interface hook, and no existing tooling for programmatic control; therefore, all infrastructure is developed from scratch.
We first specify the benchmark, and then describe the implementation for it.

\subsection{Formulation}\label{app:formulation}

\ktane is a cooperative bomb-defusal game in which two agents---the Expert \(E\) and the Defuser \(D\)---collaborate to solve a set \(\gB\) of puzzle-module instances.
The Defuser sees and manipulates the bomb but has no access to the defusal manual; the Expert has access to the manual but cannot observe the bomb.
The bomb is defused when every instance in \(\gB\) is solved before the timer expires and before the number of strikes reaches the limit \(C_S\).
We model this as a \textit{decentralised partially observable Markov decision process} \citep[Dec-POMDP;][]{Bernstein2002ComplexityDecentralizedControl,Oliehoek2016DecentralizedPOMDPFramework}:
\begin{equation}
    \gP
    =
    \left\langle\,
        \gI,
        \gX,
        \{\gA_i\}_{i\in\gI},
        T,
        R,
        \{\gO_i\}_{i\in\gI},
        \Omega
    \,\right\rangle.
\end{equation}
Here, \(\gI=\{D,E\}\) is the agent set; \(\gX\) is the environment-state space; \(\gA_i\) and \(\gO_i\) are the action and observation spaces of agent \(i\); \(T\) is the transition function; \(R\) is the shared reward function; and \(\Omega\) is the observation function, which is deterministic in our environment.

The process \(\gP\) is common to both asynchronous and synchronous evaluation modes.
The modes only differ in when each agent acts and in how much game time elapses while it does, which we describe in \cref{app:formulation:dynamics}.
The number of decision cycles in an episode is not fixed in advance, so we give the horizon through the termination conditions rather than a fixed \(h\).

\subsubsection{Game}\label{app:formulation:game}

The game state is
\begin{equation}
    s_\textrm{bomb}
    =
    (b,w,c,\tau,n_s),
\end{equation}
where \(b\) is the assignment of modules to bomb slots together with the internal state of each module; \(w\) is the widget configuration; \(c\) is the viewpoint, that is, the bomb face currently visible and the module, if any, the Defuser is zoomed into; \(\tau\in[0,C_T]\) is the remaining game time; and \(n_s\in\{0,\ldots,C_S\}\) is the current strike count.
We separate \(w\) from \(b\) because several modules condition their solution on it, creating dependencies between otherwise independent instances (\cref{app:game-modules}).
As in the base game, each strike accelerates the countdown; the rate follows from \(n_s\) and we omit it from the notation.

A mission specifies the permitted module types and counts \(\gB\), the initial time \(C_T\), the strike limit \(C_S\), and a seed.
The seed fixes every sampled component of the bomb, including each instance's parameters \(\theta_j\) and the widget configuration \(w\) (\cref{app:ktane-difficulty}).

\paragraph{Termination and outcome.}
An episode ends at the first of three events: (1) every instance in \(\gB\) is solved and the bomb
is \textit{defused}; (2) the strike count reaches the limit, \(n_s=C_S\), and the bomb
\textit{strikes out}; or (3) the timer expires, \(\tau=0\), and the bomb \textit{times out}.
Actions from inference passes still in progress when the episode ends are discarded.
The reward equals 1 on a transition into the defused state and 0 on every other transition, so the return is
\begin{equation}
    G
    =
    \mathds{1}[\textit{defused}].
\end{equation}
Although we opt for this simple outcome structure, more expressive reward functions could be explored in future work.

\subsubsection{Agent Interaction}\label{app:formulation:interaction}

A \textit{decision cycle} begins when an agent's buffers are flushed and the resulting observation is delivered.
The agent performs one inference pass and returns exactly one action, and the cycle ends when that action is dispatched.
Observation delivery and action dispatch are therefore distinct events, and the how the game time passes between them depends on the evaluation mode.

Let \(\Sigma\) denote a token vocabulary and let \(\gW=\Sigma^{\leq L}\) be the message space, the set of token sequences of length at most \(L\), where \(L\) is the per-output token limit (\cref{app:output-format}).
The action spaces are
\begin{equation}
    \gA_E
    =
    \gW
    \sqcup
    \{a_\varnothing\},
    \qquad
    \gA_D
    =
    \gW
    \sqcup
    \{a_\varnothing\}
    \sqcup
    \gA_b,
\end{equation}
where \(\sqcup\) denotes a disjoint union (agents can only perform one action per cycle), \(a_\varnothing\) is the no-op action, and \(\gA_b\) is the Defuser's bomb-action space,
\begin{equation}
    \gA_b
    =
    \gA_{\mathrm{nav}}
    \sqcup
    \left(
        \gA_{\mathrm{int}}
        \times
        \gG
    \right)
    \sqcup
    \{\texttt{release}\}.
\end{equation}
Navigation actions $\gA_{\mathrm{nav}}$ change the viewpoint \(c\) and take no argument, interaction
actions $\gA_{\mathrm{int}}$ are paired with a grounding point \(g\in\gG=[0,1]^2\) on the visible
bomb face, and \texttt{release} ends an ongoing \texttt{hold}.
The individual actions, the accepted surface forms for \(g\), and the restrictions on availability in a given state are given in \cref{app:action_space}.
Because the action spaces are disjoint unions rather than products, each agent returns exactly one action per decision cycle: the Expert either sends a message or takes a no-op, and the Defuser cannot communicate and manipulate the bomb in the same cycle.

Both agents receive their inputs through buffers.
A buffer is a first-in-first-out sequence of timestamped entries: entries are enqueued as they occur, and \(\operatorname{flush}(\cdot)\) returns them in chronological order and empties the buffer.
A buffer is flushed only when its owner begins a decision cycle.
Each agent \(i\) owns a message queue \(Q_i\), into which the other agent's messages are enqueued.
Messages are never dropped, summarised, or withheld, so a single flush may return several of them.
The Defuser additionally owns a frame buffer \(F\), into which frames of the current viewpoint are enqueued while game time advances; a flush returns the most recent \(N'=\min(N,|F|)\) frames in chronological order and discards any earlier ones (\cref{app:obs:temporal}).

At the beginning of each decision cycle, the agents receive
\begin{equation}
    o_E
    =
    \left(
        \operatorname{flush}(Q_E),
        M
    \right),
    \qquad
    o_D
    =
    \left(
        \operatorname{flush}(Q_D),
        \operatorname{flush}(F)
    \right),
\end{equation}
where \(M\) is the bomb defusal manual, which remains available to the Expert throughout the episode (\cref{app:manual}).
Neither agent observes \(s_\textrm{bomb}\) directly: the Expert relies on the Defuser's descriptions and the Defuser on the Expert's instructions.
Each agent also conditions on its own history of observations and actions, which we omit from the notation.

The full environment state is
\begin{equation}
    x
    =
    (s_\textrm{bomb},Q_D,Q_E,F,u,\gamma),
\end{equation}
where \(u\) records an ongoing interaction, such as a \texttt{hold} that has not yet been released, and \(\gamma\) records the timing information described below.
Including these components is necessary because two environments with the same game state produce different observations and different successors if their buffers or ongoing interactions differ.

\subsubsection{Game Dynamics}\label{app:formulation:dynamics}

Whenever game time advances, the bomb evolves independently of the agents: the timer decrements, module displays change, and blinking sequences progress.
Defuser agents only affect the bomb through actions dispatched at the end of a decision cycle; messages only affect the queues, and a no-op has no direct effect.
Both modes share these dynamics and only differ in how decision cycles are scheduled and how game time advances between them.

\paragraph{Asynchronous.}
The two agents run as independent concurrent processes.
Let \(r^i_k\) be the wall-clock time at which agent \(i\)'s \(k\)-th decision cycle begins and \(\delta^i_k\) the latency of its inference pass.
The resulting action is dispatched at \(r^i_k+\delta^i_k\), and the agent's next cycle begins there:
\begin{equation}
    r^i_{k+1}
    =
    r^i_k
    +
    \delta^i_k .
\end{equation}
Each agent therefore begins a cycle when its own previous inference pass completes, and at no other time.
Game time coincides with wall-clock time and never pauses, so an action executes in the state the bomb has reached when the pass that produced it completes, rather than the state the agent observed.
Inference latency is consequently part of the dynamics: a slower agent acts with less time remaining, and one agent may complete several cycles while the other completes one.
Messages and frames accumulate across that interval, so a flush may return several entries.

\paragraph{Synchronous.}
Decision cycles alternate strictly between the agents, beginning with the Defuser: \(D,E,D,E,\ldots\)
Game time is frozen while either agent performs inference and throughout Expert cycles, and advances by a fixed increment \(\Delta\) after each non-terminal Defuser cycle, including one that produces a message or a no-op.
We set \(\Delta\approx\SI{3}{\second}\), the time the game takes to resolve the animations that follow an action (\cref{app:why-3-seconds-after-act}).
After the \(k\)-th such cycle the remaining time is \(\tau_k=\max(0,C_T-k\Delta)\), so the timer permits at most \(\lceil C_T/\Delta \rceil\) Defuser cycles---mission time limits are also computed on this basis (\cref{app:time:turns}).
Importantly for this setting, inference latency has no effect on the outcome, and Expert cycles do not consume the timer.
A Defuser action is applied before the time increment, so an action that solves the final module ends the episode before any further game time elapses.
Strict alternation also bounds the message queues: at most one opposing cycle occurs between an agent's consecutive cycles and each agent sends at most one message per cycle, so \(|Q_i|\leq1\) at every flush.

\paragraph{Comparing to multi-agent interaction protocols.}
Multi-agent environments often follow one of the two PettingZoo protocols \citep{Terry2021PettingZooGymMultiAgent}. We therefore clarify the main differences with the modes in \gptnt. The Agent Environment Cycle (AEC) forces agents into turns, stepping the environment after every agent action. Our synchronous mode resembles this, except that only Defuser turns advance the clock (\cref{fig:async-vis}), since it is the only agent that can interact with the game.
The Parallel API has all agents act simultaneously, but steps only once every action is submitted. 
This differs from our asynchronous mode because relative generation time is not considered--the environment waits for the slowest agent and both agents get the same number of steps. 
Closest to our asynchronous mode is the \verb|ASYNC_PLAYER| mode in VizDoom \citep{Kempka2016ViZDoomDoombasedAI}, which equates one environment step to one tick, matching the game’s frame rate.
Therefore, the environment does not pause while an agent decides what to do.

\subsection{The Mod}\label{app:gptnt:mod}

We create a mod that adds an HTTP server to \ktane to programmatically control \ktane itself.
It runs inside the Game Instance and exposes a local HTTP server through which all of its capabilities are available to the rest of the framework. The mod provides five capabilities: (1) configuring and starting missions; (2) executing player actions; (3) exporting rendered frames; (4) exporting segmentation masks of interactable elements; and (5) exporting game state and controlling the game lifecycle. The remainder of this section describes each capability, together with the deliberate deviations we make from the stock game.

\subsubsection{Starting Missions}

Using the mod, we can start missions programmatically using HTTP requests. With this, we can configure various parameters to define the mission, including the time limit, number of strikes, total number of modules, specific module selection, random seed, and the number of widgets.
\ktane generates all module states deterministically from the integer seed, so a mission configuration fully and reproducibly specifies the bomb; we describe how seeds are selected for the benchmark in \cref{app:seed-selection}.

\subsubsection{Actions}
As previously mentioned, types of actions sent to the game are clicking, holding, releasing, zooming out, and rotating the bomb. We achieve rotations by simply editing the transform of the bomb game object. When a player interacts with a module the game zooms into it and activates its child interactables, for example, interacting with the wires module zooms into it and activates the separate wires making them interactable. The zoom out action works the opposite way and is only usable when zoomed into a module. For clicking and holding the mod expects a relative coordinate originating from the top left corner. Then a ray is cast from the specified point, onto the bomb and interacts with the first active interactable.


\paragraph{Relative coordinates for interacting.}
Coordinates are specified in a resolution-independent relative system in which $(0, 0)$ denotes the top-left corner and $(1, 1)$ for the bottom-right. Therefore, regardless of model and resolution, or any future interaction method, the action representation does not vary with perceptual configuration.


\subsubsection{Changes from the original game}


Throughout the benchmark, we prioritise running AI agents in an environment that is identical to the original game. However---in very specific cases---we need to depart from the original game.



\paragraph{Changing the initial delay for \simonsays.}

The \simonsays module works by having a looping sequence of beeps, the time between each beep is not changed but the time between the start of sequences is decreased from 5 to 1 second. When a player correctly clicks the next beep, the game originally waits 2 seconds to loop again, this is also changed to 1 second. The sequence is looped from the start whenever a player zooms into the module or gets a strike. Finally, when a player clicks on one of the simon buttons the time delay to the next sequence is reset to the original 5 seconds.

\paragraph{Always holding the bomb.}

We always start missions with the bomb already picked up, and we don't allow models to accidentally put it down. From preliminary experiments, we find that models sometimes ``zoom out'' too many times, and they struggle to re-select the bomb, so we just remove this control from the game since it is unnecessary and unrelated to the purpose of the benchmark.

\subsubsection{Exporting Frames}\label{app:mod:frames}

As part of the observations taken from the game, the original images contain a cropped version of what a human player would see when playing the game. This is achieved by having a duplicate Unity camera of the original game camera. This camera renders to a render texture which is then converted into PNG encoded bytes.\looseness=-1

An image is added to a ring buffer every 0.25 seconds with a maximum length of 16 frames. We use a buffer of frames to capture visual information over a short period of time required by the \simonsays and \morsecode modules. The 0.25-second interval between each frame is chosen based on the length of a time unit for a blink of the morse code module.
These two modules are the only two modules requiring a longer sequence of frames, while other modules only need the most recent frame.

\subsubsection{Segmentation Masks}
The last part of the observations from the game includes a segmentation mask of the interactable objects, resulting in a black image with a unique colour for each interactable game object currently displayed on the scene. A similar approach to the frame buffer is used here, where a duplicate camera renders to a render texture. This camera, however, only renders objects on a custom segmentation layer with a visual shader to handle the colours. Interactable objects in the scene are first moved to the segmentation layer and assigned a unique colour. The visual of the objects is then captured using the dedicated camera to be converted into PNG encoded bytes. Finally, the objects are returned to their original layer.
We discuss this further in \cref{app:som}.\looseness=-1

\subsubsection{Exporting Game State}

\paragraph{Bomb state.}
On request, the mod exports a structured JSON snapshot of the complete bomb state: the countdown timer, strike count, widgets, and, for every module, both its puzzle configuration (e.g., wire colours and positions, button labels, keypad symbols, maze walls) and its runtime progress (e.g., interaction and solution progress).
This snapshot is never provided to models---they must solve the task from visual observations alone (\cref{app:observations}).
Instead, the framework uses it for logging and outcome detection, for the seed-diversity protocol (\cref{app:seed-selection}), and for procedurally generating the VQA datasets (\cref{app:experiment-design:simulator}).

\paragraph{Game lifecycle.}
The mod also reports the game's lifecycle state, which is used to drive the game loop (\cref{app:framework}).
A game moves through a ``lights-off'' phase, in which the bomb is present in the scene but not yet visible or active and the timer has not started; a ``lights-on'' phase, in which the bomb is visible, all modules are active, and the timer runs; and a terminal state (all modules solved, exploded from three strikes, or exploded from timer expiry).
In addition, the mod can pause and resume the game at any point by setting Unity's internal time scale to zero, freezing the bomb timer and all animations in place. The double-pause startup protocol and the synchronous mode are both built on these two primitives.

\subsection{Process Architecture}


\paragraph{Running the game.}
The \textit{Game Instance} is the Unity game process, running a custom-built mod that exposes an HTTP server through which the game can be controlled and queried. The \textit{Game Service} is a lightweight Python-side mediator that sits between the game's HTTP server and all other components; it tracks game state, monitors process aliveness, and presents a clean abstracted interface to the rest of the system. 
\ktane can be run on Windows, Mac OS, and Linux. All the experiments in this paper are run on Linux machines running Ubuntu. Importantly, \ktane requires a graphics card that supports displays; while it can run on a CPU, it will likely lag, resulting in inconsistent results. We run all experiments using X-Server on RTX 2080Ti's.

\paragraph{Running agents.}
The \textit{Player Service} is an independent LLM agent process, one per active player. In two-player collaborative conditions, both a Defuser Player Service and an Expert Player Service are instantiated; in solo conditions, where a single model holds both roles simultaneously, only one Player Service is present. Within these, we leverage Pydantic AI\footnote{\url{https://ai.pydantic.dev/}} (v.\,1.67.0) to perform all requests with models through their API; open-source models use vLLM \citep{Kwon2023EfficientMemoryManagement} to serve an OpenAI-compatible server which can be queried through Pydantic AI.

\paragraph{Communication.}
All inter-process communication is handled over Redis channels using remote procedure call (RPC) commands, and services communicate with one another
directly over these channels. 
The \textit{Experiment Manager} (EM) is a central coordinator rather than a message broker: it owns the game lifecycle---experiment startup, readiness verification, the lights-on transition, and pausing/un-pausing---and enforces turn-taking in the synchronous mode (\cref{app:framework:sync}). 
No service holds a global view of a game while it runs: each Player Service maintains its own state and conversation history locally and exports it when the game ends, and the per-service exports are merged post hoc into a single game record for analysis.

\paragraph{Information asymmetry.} 
Information asymmetry between Defuser and Expert is enforced architecturally. The Expert's Player Service has no connection to the Game Service at any point in execution: it cannot receive game state, observe the bomb, or retrieve any information about the current game by construction of the network topology. Leakage from game to Expert is structurally impossible regardless of model behaviour.

\paragraph{Parallel execution.} 
\gptnt supports simultaneous execution of multiple independent games on a single machine. Each game has its own Game Instance and Game Service, and all are coordinated by a shared Experiment Manager. 
This is a practical requirement for conducting experiments at the scale reported in this paper.\looseness=-1

\begin{figure}[htb]
\centering
\hfill
\begin{subfigure}[t]{0.4\textwidth}
    \centering
    \includegraphics[width=\linewidth]{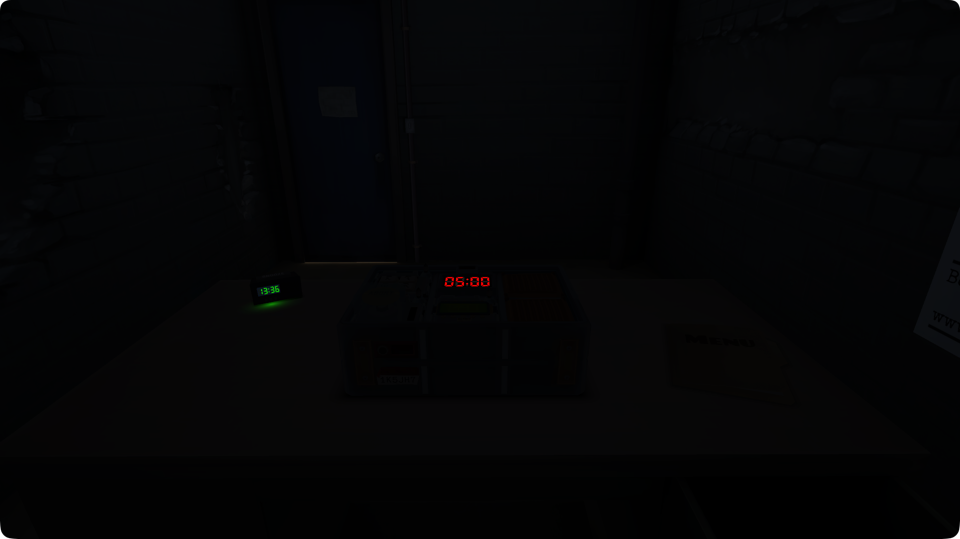}
    \caption{Lights off}
    \label{fig:lights-off}
\end{subfigure}%
\hfill
\begin{subfigure}[t]{0.4\textwidth}
    \centering
    \includegraphics[width=\linewidth]{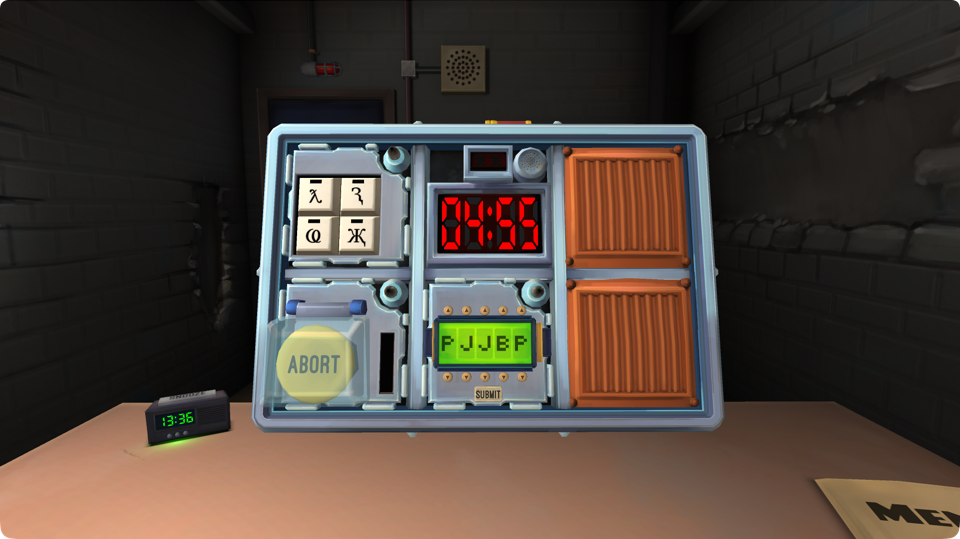}
    \caption{Lights on}
    \label{fig:lights-on}
\end{subfigure}
\hfill
\caption{Example from \ktane when the lights are off before the mission, and then on when the mission starts.\looseness=-1}
\label{fig:lights-on-off}
\end{figure}

\subsection{The Game Loop}

\subsubsection{Startup and Synchronisation}

Each game begins with the Unity process loading the bomb configuration and entering a ``lights-off'' phase (\cref{fig:lights-off}), in which the bomb is present in the scene but not yet visible or active. The EM immediately pauses the game at this point and initiates readiness checks: every active Player Service must confirm that it is connected, correctly configured, and ready to proceed before the game is permitted to advance. Once all players have confirmed readiness, the EM triggers the transition to the ``lights-on'' phase, in which the bomb becomes visible and all modules become active, and immediately pauses the game again.
This double-pause protocol guarantees that the bomb timer has not started when play begins, and that all players are in a confirmed-ready state at the moment of first un-pause. The lights-on pause is the final synchronisation point in the startup sequence. From this point, execution diverges by mode.

\subsubsection{Asynchronous Mode}\label{app:framework:async}

Asynchronous execution is the primary mode for all multi-module experiments and collaborative conditions. Its defining characteristic is that the game timer runs continuously and is never paused. As a result, all model inference time counts against the bomb clock.

At lights-on, the game resumes and two independent polling loops---one for the Defuser, one for the Expert---are launched simultaneously, and continue until the game is over. 
Each loop executes the same cycle: pull new observations and messages, produce a forward pass, sleep for 0.5 seconds, and repeat. The two loops are entirely independent; neither waits for the other, and there is no shared synchronisation primitive between them. This is the precise sense in which the mode is asynchronous. If the Defuser's API call takes ten seconds, ten seconds elapse on the bomb clock; in that same interval, the Expert may have completed three or four turns. 
Therefore, the total generation time from models is a first-class performance variable in this mode.\looseness=-1

\paragraph{Message queue.}
Communication between players is mediated by a pull-based message queue maintained independently per player. On every forward pass, a player pulls all messages dispatched to it by its partner since its last turn, draining the queue entirely before generating a response. 
Messages are never dropped: if the Expert sends four messages during a long Defuser generation, all four arrive together on the Defuser's next forward pass. This design imposes no artificial turn structure on the conversation while guaranteeing that no information is ever lost in transit.

\paragraph{Streaming.}
Current MLLMs (and their APIs) do not support receiving new input while a generation is in progress. Therefore, we do not use streaming outputs either. In any case, streaming would not be useful: the receiving player requires the partner's complete output before it can form a coherent response, meaning streaming would not reduce effective communication latency under the present architecture.

\subsubsection{Synchronous Mode}\label{app:framework:sync}

Synchronous mode shares the startup sequence and message semantics of
asynchronous mode but differs in two ways: the game is paused while models
generate, and the two players act in strict alternation rather than in
free-running loops. Clock time advances by a fixed budget per Defuser turn
rather than in real time, so model generation time has no bearing on
effective game time.

At lights-on, the game remains paused and the Defuser takes the first turn. Each cycle then proceeds as follows:\looseness=-1
\begin{enumerate}[leftmargin=*]
  \item \textbf{Observe.} The Defuser pulls its observations and drains its message queue. Because the game is paused, observations are captured from a frozen, fully resolved game state.
  \item \textbf{Generate.} The Defuser produces a forward pass. Unity's internal time scale is set to zero during generation, so the model may take unbounded wall-clock time at no cost in game time.
  \item \textbf{Act.} The action is dispatched to the game, which is resumed for a window of at least \SI{3}{\second} of game time; in practice the realised window is \SIrange{3}{3.5}{\second} (\cref{app:why-3-seconds-after-act}). The action executes and all resulting animations and state transitions resolve before the game is paused again. The bomb clock therefore advances for each Defuser turn, regardless of the type of action produced by the Defuser.
  \item \textbf{Expert turn.} While the game is paused, the Expert pulls its message queue, produces a forward pass, and dispatches its action. Expert turns consume no game time.
\end{enumerate}

The cycle repeats until the game ends. Because each Defuser turn consumes \SI{3}{\second}, a mission's time limit translates into a maximum number of Defuser turns---the time limit divided by three (\cref{app:time}).

\FloatBarrier

\subimportlevel{.}{observations}{1}

\subsection{Providing the Manual}\label{app:manual}

The official manual---available at \url{https://bombmanual.com}---contains detailed instructions on how to solve each module, including illustrations to aid the Expert in their task as they cannot observe the game directly. 

The manual is provided to the Expert as interleaved text and images---one page of extracted text followed by the corresponding page image, repeated for every relevant page.
The full manual runs to 23 pages, though irrelevant ones are discarded, leaving roughly 16 module pages plus appendix content. Each page is A4 (equivalent to $2{,}550\times3{,}300$px), and therefore downscaled to match the same resolution as the visual observation frames the model receives during play (\cref{app:observations}). Importantly, we preserve the orientation of the pages, ensuring pages remain in portrait orientation.

\begin{figure}[tbh]
\centering
\includegraphics[width=1\linewidth]{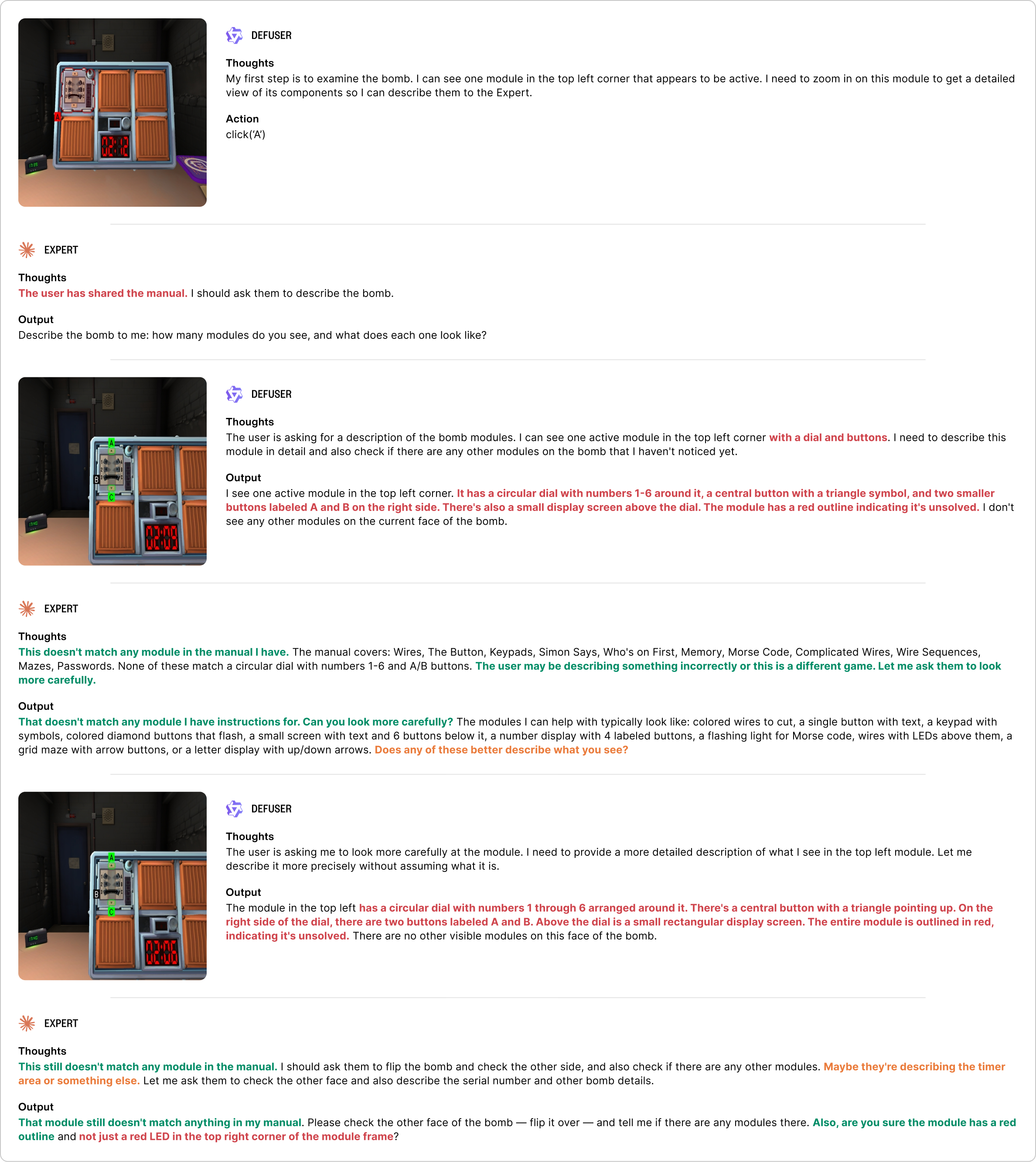}
\caption{Gameplay excerpt for Qwen (Defuser) and Sonnet (Expert) playing a single-module \wiresequence mission (seed 845). Qwen hallucinates the module's appearance, which Sonnet recognises as implausible. Even after clarification requests from Sonnet, Qwen adheres to its original incorrect description. Qwen also misinterprets the SoM overlay of the module---incorrectly reporting the outline colour as indicative of its status---and further violates the SoM instructions in the system prompt by reporting the location marker labels to the Expert.}
\label{fig:example:qwen-claude-wire_sequence-845}
\end{figure}

Because system prompts cannot contain images, we instead provide the manual in the Expert's first incoming message, where it is protected from context truncation regardless of how long the dialogue runs.
Unfortunately, this introduces a semantic oddity as a practical side effect: the Defuser, who by definition cannot see the manual, appears to be the one delivering it.
We address this directly in the system prompt, and across all our recorded games, models consistently treat the Defuser as unable to consult the manual in practice (e.g., \cref{fig:example:qwen-claude-wire_sequence-845}).\looseness=-1

\subimportlevel{.}{action_space}{1}

\subimportlevel{.}{output_format}{1}

\FloatBarrier
\section{Experimental Design}\label{app:experiment-design}

\subsection{Seed Selection Protocol for Mission Diversity}\label{app:seed-selection}

\ktane generates all module states deterministically from an integer seed. Each experimental condition requires multiple bomb configurations that can be assigned to different trials without introducing confounds from repeated puzzles.  We therefore need seeds whose resulting configurations are \emph{structurally} distinct---different in the state space that determines the correct solution path, not merely in visual appearance.

We design an automated protocol that selects a \textbf{single outer seed} maximising structural diversity across all bomb configurations used in the benchmark.  The protocol searches over a candidate range of 1--50, instantiates every mission configuration for each candidate in the game, strips runtime-only fields to isolate puzzle-relevant state, and ranks candidates by a normalised pairwise diversity metric. The highest-ranked candidate is selected. 
The candidate range of 50 outer seeds is pragmatic rather than exhaustive, and child seeds are shared across module types rather than drawn independently.





\begin{table}[ht]
\centering
\footnotesize
\renewcommand{\arraystretch}{1.4}
\begin{tabularx}{0.8\linewidth}{@{}lX@{}}
\toprule
Term & Meaning \\
\midrule
Outer seed & An integer in $[1, 50]$ passed to the mission generator's RNG. Each candidate outer seed produces one complete set of mission configurations. \\
Child seed  & One of the 10 integers sampled by the mission generator from the outer seed. Each child seed is passed to \ktane to instantiate one bomb. \\
Module      & A single puzzle component on the bomb (e.g., \wires, \simonsays, \maze). \\
Mission     & A fully specified bomb configuration: which modules it contains, time limit, and widget settings. \\
Bomb state  & The structured JSON snapshot of all module state returned by \ktane. \\
\bottomrule
\end{tabularx}
\caption{Key terminology used in the seed selection protocol.}
\label{app:tab:seed-terminology}
\end{table}

\paragraph{Outer-to-Child Seed Mapping.}
The mission generator is initialised with the outer seed and draws 10 child seeds from a seeded RNG.  Because \ktane is deterministic, the same child seed always produces the same module configuration.  The same 10 child seeds are reused across all module types, so child seed~$k$ determines the $k$-th instance of every type.

\paragraph{Conditions Evaluated.}
The search evaluates each candidate outer seed across six conditions: the single-module condition (10 instances $\times$ 11 types), two repeated-module conditions (2 and 5 copies of the same type), and three mixed-module conditions (2--5 different types per bomb; module combinations hand-specified as described in \cref{app:ktane-difficulty}). Only the single-module condition---where the outer seed fully determines all puzzle instances---is used in the experiments reported in this paper (\cref{sec:low-level}); the remainder serve as cross-checks.

\paragraph{Diversity Metrics.}
Structural difference between two normalised bomb states is computed by recursively comparing their JSON representations---counting value changes, added keys, removed keys, and type changes---and normalising by the total number of unique attributes across both states:
\begin{equation}
  \text{diversity}(s_1, s_2)
    = \frac{|\text{structural differences}|}
           {\text{total unique attributes across } s_1 \text{ and } s_2}
  \label{eq:diff-ratio}
\end{equation}
This yields a value in $[0,1]$ (0 = identical, 1 = completely disjoint).
We aggregate at three levels:
\begin{itemize}[leftmargin=*]
\item \textit{Within-type}: average pairwise diversity across the 10 instances of a single module type---are instances sufficiently distinct?
\item \textit{Within-bomb}: average pairwise diversity of modules on the same multi-module bomb---are co-located puzzles redundant?
\item \textit{Across-bomb}: average pairwise diversity of bombs within a condition---do different trials present genuinely different challenges?
\end{itemize}

\paragraph{Selection and Limitations.}
All three metrics are averaged per outer seed; seeds are ranked by this mean, the top three manually reviewed, and the highest-ranked seed selected. Selection is purely comparative---no minimum diversity threshold is enforced.
The aggregation weights all metrics and conditions equally, so conditions with more missions contribute more pairwise comparisons (though the single-module condition dominates).\looseness=-1

\subsection{Manual VQA Dataset}\label{app:experiment-design:manual}
The manual datasets are designed for benchmarking the \textit{Expert's} ability to understand and reason over the rules within the manual.
Manual VQA requires models to perform OCR on dense manual pages, locate relevant rules given a partially-described bomb state, and reason over those rules to produce correct defusal instructions. The dataset comprises 884 multiple-choice questions across all modules, targeting five underlying capabilities measured by distinct question types. For a breakdown of samples per question type, see \cref{tab:manual-vqa-categories}.

\paragraph{Construction.} For each module, we construct targeted questions that isolate the main reasoning demands the Expert faces during gameplay. Each question is presented as a natural-language description of a bomb state, mimicking what the Defuser would communicate, followed by a question probing the Expert's decision, with between 2 and 8 answer options. To mirror the interactive setting, questions are posed with the manual provided as interleaved text and images, as it would be during gameplay.


\begin{table}[tbh]
\centering
\footnotesize
\renewcommand{\arraystretch}{1.4}
\begin{threeparttable}
\caption{Question category taxonomy for the Manual VQA dataset.}
\label{tab:manual-vqa-categories}
\begin{tabularx}{\linewidth}{@{}l X S[table-format=3.0]@{}}
\toprule
Category & Description & {$n$} \\
\midrule
Reading
  & Read specific text or values from the manual.
  & 191 \\
Element Grounding
  & Locate a described visual element within the manual.
  & 60 \\
Procedural Reasoning
  & Apply the manual rules to determine the correct Defuser action.
    Includes \textit{easy} questions that require one or two rule lookups,
    and \textit{hard} questions that require multi-step or conditional
    reasoning, combining information across several rule branches or
    manual sub-tables.
  & 499 \\
Ambiguity Detection
  & Recognise that the state description is underspecified.
  & 64 \\
Hallucination
  & Recognise that the state description contains information that does
    not correspond to any valid case in the manual (e.g., a wire colour
    or condition that cannot occur).
  & 70 \\
\bottomrule
\end{tabularx}
\end{threeparttable}
\end{table}

\subsection{Simulator VQA and Localisation Datasets}\label{app:experiment-design:simulator}
The simulator datasets are designed to assess the \textit{Defuser's} ability to perceive and localise state-critical information directly from game screenshots, allowing us to quantify perceptual capabilities independent of decision-making logic.

\paragraph{Trajectory Collection.} To ensure a representative distribution of game states, we employ two distinct data-collection agents: an \textit{optimal oracle} that generates successful trajectories, and a \textit{stochastic dummy} agent that explores failure states (e.g., strikes, explosions, and suboptimal interactions). 
This ensures coverage of diverse bomb states across both successful and unsuccessful game progressions, for both single-module and multi-module configurations.
We deduplicate the collected images by hashing state configurations and sample up to 40 images per module to avoid redundancy and ensure diversity of visual inputs.

\paragraph{Visual Question Answering.}
We generate 640 VQA pairs to probe the model across several perceptual primitives spanning basic perception such as colour recognition, counting and reading, and robustness to ambiguity and hallucination probes requiring the model to identify when a question is unanswerable or refers to a non-existent element. 
We prepare hand-written templates for each module and question category, and procedurally generate question-answer pairs by instantiating them with state metadata. Both the correct answer and the distractors are derived from the ground-truth simulator state, so labels are correct by construction and distractors are drawn from the same state schema as the target (e.g., colours present elsewhere on the bomb). 
Questions are structured in multiple-choice format to enable objective and scalable evaluation. \cref{tab:simulator-vqa-categories} summarises the breakdown of samples per question type.

\begin{table}[tbh]
\centering
\footnotesize
\renewcommand{\arraystretch}{1.4}
\begin{threeparttable}
\caption{Question category taxonomy for the Simulator VQA dataset.}
\label{tab:simulator-vqa-categories}
\begin{tabular}{@{}l l S[table-format=3.0]@{}}
\toprule
Category & Description & {$n$} \\
\midrule
Reading
  & Read a specific text or value from the simulator scene.
  & 152 \\
Colour
  & Identify the colour of a specified element within the simulator scene.
  & 61 \\
Counting
  & Count the number of visible instances of a specified element type within the simulator scene.
  & 205 \\
Ambiguity
  & Recognise that the visible scene state does not contain sufficient information to give a unique answer.
  & 91 \\
Hallucination
  & Recognise that the question references an element or condition that does not exist in the scene.
  & 131 \\
\bottomrule
\end{tabular}
\end{threeparttable}
\end{table}

\paragraph{Localisation.}
We generate 665 grounding questions following the same perceptual taxonomy (see \cref{tab:statics:localisation:categories} for descriptions and examples), evaluated across two paradigms. In \textit{set-of-marks} (SoM), images are overlaid with labelled segmentation masks and the model predicts the label corresponding to a requested UI element. In \textbf{Coordinate Prediction}, the model receives a raw image and must output $(x, y)$ coordinates for a target element; our framework supports both normalised and unnormalised coordinates, and during evaluation we follow the format supported by each model. This setup enables a direct comparison of the two paradigms, informing the interaction protocol used in our interactive evaluations. Performance is measured by accuracy; for coordinate prediction, a predicted point is correct if it falls anywhere within the area of the target element.
Descriptions and examples are in \cref{tab:statics:localisation:categories} and the number of samples per module in \cref{tab:localization-samples}.

\begin{table}[tbh]
\centering
\footnotesize
\renewcommand{\arraystretch}{1.4}
\begin{threeparttable}
\caption{Question category taxonomy for the Simulator Localisation dataset.}
\label{tab:statics:localisation:categories}
\begin{tabular}{@{}l l l @{}}
\toprule
Category & Description & Example \\
\midrule
Reading
  & Identify target by reading its text label
  & \textit{Click the button that says: NEXT} 
  \\
Colour
  & Identify target by its colour
  & \textit{Click the red button} 
  \\
Counting
  & Select target by ordinal index among elements of the same type
  & \textit{Click the 3rd wire}
  \\
Element
  & Identify target by element type alone, with no further qualifier
  & \textit{Click the button}
  \\
Position
  & Locate target by spatial descriptor or distinctive visual marker
  & \textit{Click the top left button}
  \\
\midrule[0.1ex]
Ambiguity
  & Target exists but cannot be uniquely resolved
  & \textit{Click the up button}\tnote{a} 
  \\
Hallucination
  & Referenced element is absent from the image
  & \textit{Click the black button}\tnote{b} 
  \\
\bottomrule
\end{tabular}
\begin{tablenotes}
  \scriptsize
  \item[a] Each of the five letters in \passwords has its own up button.
  \item[b] Black does not exist on \simonsays.
\end{tablenotes}
\end{threeparttable}
\end{table}

\begin{table}[tbh]\centering\footnotesize\renewcommand{\arraystretch}{1.4}
\begin{threeparttable}
\sisetup{table-format=2.0}
\caption{Number of Simulator Localisation samples per module.}
\label{tab:localization-samples}
\begin{tabular}{@{} *{11}{S} S[table-format=3.0] @{}}\toprule
{\wires*} & {\button*} & {\keypad*} & {\simonsays*} & {\whosonfirst*} & {\memory*} & {\maze*} & {\morsecode*} & {\complicatedwires*} & {\wiresequence*} & {\passwords*} & {\textit{Total}}\\ \midrule
45 & 38 & 77 & 42 & 81 & 48 & 43 & 63 & 78 & 82 & 68 & 665\\
\bottomrule
\end{tabular}
\end{threeparttable}\end{table}

\FloatBarrier
\section{Model Selection and Configuration}\label{app:models}


The models used in the experiments are from several providers. We use the short-form names throughout the paper, and list their corresponding versions and provider in \cref{tab:model-table}.

\begin{table}[tbh]
\centering
\footnotesize
\renewcommand{\arraystretch}{1.4}
\begin{threeparttable}
\caption{Model details including version, provider, context length, number of tokens for the manual, and knowledge cutoff for each evaluated model.}
\label{tab:model-table}
\begin{tabular}{@{}l l l S[table-format=7.0] S[table-format=5.0] c@{}}
    \toprule
    Name & Version & Provider & {Context Length} & {Manual} & {Knowledge Cut-off} \\
    \midrule
    \claude* Sonnet 4.6 & \texttt{claude-sonnet-4-6-v1} & Microsoft Azure & 264000 & 16300  & Feb 2025 \\
    \openai* GPT-5.2 & \texttt{gpt-5.2-2025-12-11} & Microsoft Azure & 528000 & 12518  & Aug 2025 \\
    \gemini* Gemini 3 Flash & \texttt{gemini-3-flash-preview} & Google Vertex AI & 1114112 & 12945 & Jan 2025 \\
    \qwen* Qwen3.5 (27B)  & \texttt{Qwen/Qwen3.5-27B} & Hugging Face & 128000 & 9893\tnote{*} & Nov 2024\tnote{\dag} \\
    \internvl* InternVL 3.5 (38B) & \texttt{OpenGVLab/InternVL3\_5-38B} & Hugging Face & 40060  & 9770\tnote{**} & May 2025\tnote{\dag} \\
    \bottomrule
\end{tabular}
\begin{tablenotes}
\scriptsize
\item[\dag]Assumed knowledge cut-off
\item[*] Without resizing the manual for Qwen, it requires 10,597 input tokens. We resize images for Qwen to keep the open-source models comparable. More details in \cref{app:observations}. 
\item[**] Without resizing the manual for InternVL, it requires 58,922 input tokens. 
\end{tablenotes}
\end{threeparttable}
\end{table}

\paragraph{Open-source model deployment.}

Qwen 3.5 (27B) and InternVL 3.5 (38B) are served locally using vLLM \citep{Kwon2023EfficientMemoryManagement} on a server equipped with 2$\times$ NVIDIA H100 80GB GPUs.
Models are downloaded from Hugging Face and served at BF16 precision without quantisation.
Both models are exposed through vLLM's OpenAI-compatible API and queried through the identical client harness used for the proprietary models, so prompt construction, image encoding, and output parsing are uniform across all systems.

\paragraph{Handling false-positive guardrails.}

Models use the word ``bomb'' often when they are evaluated on this benchmark. 
When one player's message is fed into another player's prompt, the recipient's provider can refuse the request on content-policy grounds. What triggers the filters is not always clear: we do not observe a consistent pattern across messages or games.
We treat this as an external infrastructure failure rather than a task failure, since the refusal originates with the provider, not with the model's competence at the game. 
When a guardrail violation is raised, the affected player returns a 4XX HTTP error instead of a response. 
We implement a pragmatic solution that does not appear to affect the context of the offending model in any meaningful way: we mimic a message from the other player, asking them, ``Can you rephrase that please?'' 
Both the offending message and this reply are then dropped from the recipient's context. This is deliberate, as the offending message is what trips the filter, and so retaining it in the model's context would re-trigger the refusal on the next turn. 
This means that the two players no longer share the same record of the conversation: the sender keeps the message it wrote, while the recipient keeps neither that message nor the rephrase request.
The game then continues, and the players lose a single turn each as a result.


\subsection{Selection Criteria}\label{app:models:selection}

This is not a model comparison paper; models are instruments for probing the benchmark rather than subjects of study in their own right. Model selection is governed by what the task requires, and not by what would make any particular model appear better.
Four criteria are treated as non-negotiable hard requirements. \looseness=-1

\begin{enumerate}[leftmargin=*]
    \item \textbf{Conversational capability.}
    Models must sustain coherent multi-turn dialogue across a full game, which can span dozens of turns per player. Therefore, the base conversational ability must not have been degraded by task-specific post-training.
    \item \textbf{Interleaved multi-image multi-turn support.}
    Models must handle sequences of images interleaved with text within a single conversation.
    The Defuser receives new visual observations on every turn, and the Expert receives the manual as a sequence of page images interleaved with extracted text across the first user turn. 
    Architectures that support only single-image input, or that do not correctly handle image tokens distributed across a multi-turn conversation history, cannot participate in the task.
    \item \textbf{Context length.}
    Models must be capable of holding the Expert's manual and a sufficient dialogue history simultaneously within their context window. 
    This places a practical lower bound on usable context length, with InternVL representing the most constrained case at 40,060 tokens (\cref{tab:model-table}).
    \item \textbf{No task-specific training.}
    Models may not be fine-tuned on \ktane, the game manual, or any task-derived data prior to their evaluation. 
    Fine-tuning on the manual would trivialise the benchmark: a model capable of reciting module solutions from parametric memory is not performing the collaborative grounded reasoning the benchmark is designed to measure.
    Generalisability across novel puzzle configurations would be destroyed, and the resulting evaluation would reflect recall rather than reasoning. Task-specific fine-tuning is also detectable via parametric knowledge probes and would be visible in contamination checks (\cref{sec:e2-parametric-knowledge}).\looseness=-1
\end{enumerate}

\paragraph{Choice of model variants.}
We deliberately evaluate mid-sized variants within closed-source families rather than the largest available models. Flagship variants increasingly default to extended reasoning \citep{DeepSeek-AI2025DeepSeekR1IncentivizingReasoning,Snell2024ScalingLLMTestTime}, and in a real-time benchmark, every additional reasoning token consumes game time. Therefore, there is a trade-off between deliberating and acting \citep{Cuadron2025DangerOverthinkingExamining}, and inference latency limits performance in real-time environments \citep{Zhang2025VideoGameBenchCanVisionLanguage,Kang2025WinFastLose}. 
The cost of the largest models also places the full evaluation suite beyond many academic budgets (\cref{app:model-cost}), a constraint that increasingly excludes academic teams from evaluating frontier models \citep{Besiroglu2024ComputeDivideMachine}, which cost-controlled agent evaluation is meant to address \citep{Kapoor2024AIAgentsThat}. Therefore, the mid-sized tier is a more accessible and reproducible point of comparison \citep{Biderman2026LessonsTrenchesReproducible}, and we make no claim that our results bound the performance of the largest models.

\subsection{Sampling Configuration}\label{app:model:sampling-config}
All models are evaluated under the same sampling policy, using a temperature of 0.6; this value is not tuned per model. Per-model temperature optimisation would introduce a model-specific degree of freedom, conflating sampling configuration with the underlying model capabilities the benchmark is designed to measure.
Other decoding parameters (top-$p$, top-$k$, and repetition penalties) are left at each provider's default.
Preliminary experiments revealed that the determinism resulting from zero and near-zero (<=0.3) temperatures led to models getting stuck in loops of repeated unproductive thoughts and actions, indicating that a certain degree of stochasticity is a functional requirement for the task. We chose a moderate temperature to prevent this kind of behaviour while constraining creativity and encouraging factuality. Out of candidate temperatures in the moderate 0.4-0.8 range, we chose 0.6 in line with recommendations for mitigating repetition in thinking for InternVL, the weakest model we tested.

\subsection{Evaluation Cost}\label{app:model-cost}

We provide token counts and cost for each experiment of the benchmark to help future work to budget for experiments accordingly in \cref{tab:benchmark-cost}. Importantly, we highly recommend that caching is used as we find that 41.2\% of tokens can be cached.

\begin{table}[tbh]
\centering
\footnotesize
\sisetup{table-align-text-post = false}
\caption{Cost to run all the experiments used in this paper \textit{without caching}, and the total number of input and output tokens across every request.}
\label{tab:benchmark-cost}
\begin{threeparttable}
\begin{tabular}{@{}p{3mm} l S[table-format=3.0] S[table-format=9.0] S[table-format=7.0] S[table-format=4.2]@{}}
\toprule
&&& \multicolumn{2}{c}{Tokens} \\
\cmidrule(lr){4-5}
&  & {$n$} & {Input} & {Output} & {Cost (\$)}  \\
\midrule
\multirow{5}{*}{\rotatebox[origin=c]{90}{\textbf{\textls[25]{Async, Multi}}}}
& \claude* Sonnet 4.6           &  20 &   9998492 &   79425 &   31.20 \\
& \openai* GPT-5.2              &  20 &  12802985 &   57564 &   14.30 \\
& \gemini* Gemini 3 Flash       &  20 &  27882546 &   42095 &    8.36 \\
& \qwen* Qwen3.5 (27B)          &  20 &  41203903 &  158183 &    0.00\tnote{*} \\
& \internvl* InternVL 3.5 (38B) &  20 &  52816695 &   89090 &    0.00\tnote{*} \\
\midrule[0.1ex]
\multirow{5}{*}{\rotatebox[origin=c]{90}{\textbf{\textls[25]{Async, Single}}}}
& \claude* Sonnet 4.6           & 220 &  69502082 &   511918 &  216.00 \\
& \openai* GPT-5.2              & 220 &  51645133 &   500257 &   105.00 \\
& \gemini* Gemini 3 Flash       & 220 & 66767165 &   268274 &   36.80 \\
& \qwen* Qwen3.5 (27B)          & 220 & 100524443 &   965880 &    0.00\tnote{*} \\
& \internvl* InternVL 3.5 (38B) & 220 & 125536094 &   574853 &    0.00\tnote{*} \\
\midrule[0.1ex]
\multirow{5}{*}{\rotatebox[origin=c]{90}{\textbf{\textls[25]{Sync, Multi}}}}
& \claude* Sonnet 4.6           &  20 &   9219128 &    83263 &   28.90 \\
& \openai* GPT-5.2              &  20 &  19544304 &    69972 &   20.80 \\
& \gemini* Gemini 3 Flash       &  20 &  25880604 &    45170 &    8.32 \\
& \qwen* Qwen3.5 (27B)          &  20 &  33965559 &   174904 &    0.00\tnote{*} \\
& \internvl* InternVL 3.5 (38B) &  20 &  26847549 &    94501 &    0.00\tnote{*} \\
\midrule[0.1ex]
\multirow{5}{*}{\rotatebox[origin=c]{90}{\shortstack{\textbf{\textls[25]{Sync, Single,}}\\\textbf{\textls[25]{Other-Play}}}}}
& \claude* Sonnet 4.6           & 880 & 338729041 &  3325371 & 1066.00 \\
& \openai* GPT-5.2              & 880 & 593946839 &  2154517 &  633.00 \\
& \gemini* Gemini 3 Flash       & 880 & 600336971 &  1555825 &  192.00 \\
& \qwen* Qwen3.5 (27B)          & 880 & 630010233 &  3870496 &    0.00\tnote{*} \\
& \internvl* InternVL 3.5 (38B) & 880 & 559989035 &  2568117 &    0.00\tnote{*} \\
\midrule[0.1ex]
\multirow{5}{*}{\rotatebox[origin=c]{90}{\shortstack{\textbf{\textls[25]{Sync, Single,}}\\\textbf{\textls[25]{Self-Play}}}}}
& \claude* Sonnet 4.6           & 220 &  71361568 &   662936 &  224.00 \\
& \openai* GPT-5.2              & 220 & 122965651 &   527790 &  133.00 \\
& \gemini* Gemini 3 Flash       & 220 & 153748981 &   324218 &   48.90 \\
& \qwen* Qwen3.5 (27B)          & 220 & 176796830 &  1112086 &    0.00\tnote{*} \\
& \internvl* InternVL 3.5 (38B) & 220 & 144868042 &   637688 &    0.00\tnote{*} \\
\midrule[0.1ex]
\multirow{5}{*}{\rotatebox[origin=c]{90}{\textbf{\textls[25]{Parametric}}}}
& \claude* Sonnet 4.6           & 110 &  12757175 &   429406 &   44.70 \\
& \openai* GPT-5.2              & 110 &  21227156 &   212506 &   24.90 \\
& \gemini* Gemini 3 Flash       & 110 &  17876072 &   119206 &    7.28 \\
& \qwen* Qwen3.5 (27B)          & 110 &  35737231 &   600342 &    0.00\tnote{*} \\
& \internvl* InternVL 3.5 (38B) & 110 &  27499363 &   229359 &    0.00\tnote{*} \\
\midrule[0.1ex]
\multirow{5}{*}{\rotatebox[origin=c]{90}{\textbf{\textls[25]{Solo}}}}
& \claude* Sonnet 4.6           & 110 &  30690782 &   398573 &   98.10 \\
& \openai* GPT-5.2              & 110 &  59958861 &   168656 &   63.30 \\
& \gemini* Gemini 3 Flash       & 110 &  97779089 &   123501 &   30.90 \\
& \qwen* Qwen3.5 (27B)          & 110 &  81786069 &   493494 &    0.00\tnote{*} \\
& \internvl* InternVL 3.5 (38B) & 110 &  87283493 &   336348 &    0.00\tnote{*} \\
\bottomrule
\end{tabular}
\begin{tablenotes}\scriptsize
    \item $^*$\,Models are self-hosted.
\end{tablenotes}
\end{threeparttable}
\end{table}

\FloatBarrier
\section{Human Gameplay Results}\label{app:humans}

\subsection{Recruitment and Demographics}
We recruited ten pairings, for a total of 20 participants. The players in each pairing knew each other before participating in our experiments, as we recruited the first ten players from a pool of senior undergraduate and postgraduate Computer Science students. These players were then required to recruit the second player, such as a friend, family member or spouse. Of our 20 players, 12 identify as male, seven as female, and one as non-binary. Given the vast majority of our participants are residents of the UK and Canada (19), most (18) report English to be their first language. Four players disclose learning difficulties or visual impairments, but find this to be no hindrance for playing the game. Prior experience with the game varies somewhat between players, with nine never having played \ktane before, ten having played \ktane one to ten times, and one playing \ktane about once a month.

The human study was reviewed and approved by the Research Ethics Committee of our institution (Project 9748). All participants were adults, self-declared as not belonging to any vulnerable group, and took part voluntarily. Before playing, each participant received an information sheet describing the study, the data collected, the retention period, and their right to withdraw at any point. Informed consent was obtained electronically prior to participation. Participants received a £20 voucher as compensation. Each participant was assigned a pseudonymous identifier used in all logs: we only recorded gameplay logs, and coarse demographic attributes (age range, gender identity, first language, and prior experience with \ktane); no audio or video was recorded. Participants were instructed not to disclose personal information during play. All data is stored on institutional, password-protected systems.

\subsection{Setup}
Humans played either in person in the same room, or via video call; in both scenarios, the Defuser
has the game on a laptop or desktop computer, and the Expert has the manual on a second device
(tablet, laptop, or desktop computer). In both scenarios, the players communicate with each other
via voice, and are consequently able to interrupt each other and act while the other player is
still speaking. Note that this is not the case for the models, which use text messages in a non-streaming
fashion. An additional advantage human players have over the models is that they are able to
debrief.\footnote{We neither prevent players from, nor explicitly encourage them to discuss previous
missions before attempting the next.} Before playing the missions, all players play the tutorial
missions \textit{`1.1 Game Overview \& Roles'} and \textit{`1.2 Controls Tutorial'}. All pairings
play the easiest mission (\#1) first, the order of the remaining nine missions is randomised. The
players switch roles after the first five missions, so that each player plays half of the games as
the Expert and half as the Defuser.

\subsection{Human-specific Challenges of KTANE}
The module rules are complex enough that human players are unlikely to hold the full set of branching pathways in working memory, and solutions typically depend on a combination of contextual factors---serial numbers, battery counts, elapsed time---that must be communicated and continuously tracked.
This load increases further when partners do not share a visual space, as the grounding that comes from shared context is unavailable \citep{Gergle2004LanguageEfficiencyVisual, Kraut2003VisualInformationConversational,Lemon2022ConversationalGroundingEmergent}. 
Under time pressure, these challenges intensify \citep{Beres2021PressureExploringChoke}: humans tend to rely on incomplete heuristics \citep{Spiliopoulos2018ComplexityAttentionChoice, vanHarreveld2007EffectsTimePressure,Payne1988AdaptiveStrategySelection}, produce simpler speech, and can fail to account for their partner's knowledge state \citep{Saslow2014SpeakingPressureLow, Horton1996WhenSpeakersTake}. 

These pressures do not apply symmetrically to artificial agents. 
Unlike human players, they are not subject to the same working-memory constraints: the full manual fits within many contemporary models' context window.
This capacity advantage does not, however, straightforwardly translate into performance. Fitting information in context is not the same as reasoning over it: long-context retrieval and integration over extended inputs remain documented failure modes for current models \citep{Kuratov2024BABILongTestingLimits, Wang2025MMLongBenchBenchmarkingLongContext}.\looseness=-1 

\subsection{Results}\label{app:humans:results}

Human performance (\cref{tab:e8-grid}) roughly corresponds to the categorisation provided for missions in the game, with the most solved missions for the easier levels---mission \#1, while constructed to correspond to the easiest level (2.1), is played first by each pairing, which likely explains why the performance is not on par with mission \#2 (also level 2.1). Another outlier is mission \#7 (level 3.1), which humans seem to find more challenging than the mission from the next level up (\#8). No human pairing is able to solve missions \#9 and \#10, sampled from the hardest level represented in our ten missions (3.3). All pairings time out on \#10, although for three pairings the timeout occurs while they are attempting to solve the final module. A similar pattern can be observed for \#9, where the majority of pairings time out while working on the final module. Given that across all unsolved missions (\cref{tab:e8-stats}), human pairings tend to strike out (72\% of failures) rather than time out, this strongly suggests that the main challenge with \#9 and \#10 is primarily in the time pressure. Given that no pairing consists of two experienced players and the original game progresses to missions in significantly more complex levels (levels 4--7), we consider this a soft upper bound for human performance.\looseness=-1

\begin{table}[tbh]

\centering
\footnotesize
\renewcommand{\arraystretch}{1.4}
\setlength{\tabcolsep}{3pt}
\setlength{\iconboxwidth}{0.45cm}
\sisetup{table-format=1.0}
\renewcommand{\iconboxheight}{0.4cm}
\begin{threeparttable}
\caption{Game outcomes and number of modules solved for all human pairings across the 10 distinct missions, in asynchronous mode. In this setting, we do not stop the game clock during play. Each mission is attempted once (pass@1).}
\label{tab:e8-grid}
\begin{tabular}{@{}
  l
  @{\hspace{\tabcolsep * 2}}
  c
  @{\hspace{\tabcolsep * 2}}
  c
  @{\hspace{\tabcolsep * 2}}
  c
  @{\hspace{\tabcolsep * 2 + 1pt}}
  c
  @{\hspace{\tabcolsep * 2 - 2pt}}
  c
  @{\hspace{\tabcolsep * 2 - 2pt}}
  c
  @{\hspace{\tabcolsep * 2 + 1pt}}
  c
  @{\hspace{\tabcolsep * 2}}
  c
  @{\hspace{\tabcolsep * 2}}
  c
  @{\hspace{\tabcolsep * 2}}
  c
  S
  @{}}
  \toprule
  & \multicolumn{10}{c}{Mission (ordered by difficulty)} \\
  \cmidrule(r){2-11}
  & 1 & 2 & 3 & 4 & 5 & 6 & 7 & 8 & 9 & 10 \\
  \cmidrule(r){2-11}
  \addlinespace[2pt]
  \multicolumn{1}{r}{\scriptsize\itshape\scshape front:}
    & \wiresicon[height=\iconboxheight,boxwidth=\iconboxwidth]\mazeicon[height=\iconboxheight,boxwidth=\iconboxwidth]
    & \buttonicon[height=\iconboxheight,boxwidth=\iconboxwidth]
    & \keypadicon[height=\iconboxheight,boxwidth=\iconboxwidth]\whosonfirsticon[height=\iconboxheight,boxwidth=\iconboxwidth]
    & \buttonicon[height=\iconboxheight,boxwidth=\iconboxwidth]\memoryicon[height=\iconboxheight,boxwidth=\iconboxwidth]
    & \wiresicon[height=\iconboxheight,boxwidth=\iconboxwidth]\memoryicon[height=\iconboxheight,boxwidth=\iconboxwidth]\mazeicon[height=\iconboxheight,boxwidth=\iconboxwidth]
    & \whosonfirsticon[height=\iconboxheight,boxwidth=\iconboxwidth]
    & \whosonfirsticon[height=\iconboxheight,boxwidth=\iconboxwidth]
    & \buttonicon[height=\iconboxheight,boxwidth=\iconboxwidth]
    & \keypadicon[height=\iconboxheight,boxwidth=\iconboxwidth]\wiresequenceicon[height=\iconboxheight,boxwidth=\iconboxwidth]\mazeicon[height=\iconboxheight,boxwidth=\iconboxwidth]
    & \wiresicon[height=\iconboxheight,boxwidth=\iconboxwidth]\morsecodeicon[height=\iconboxheight,boxwidth=\iconboxwidth]
  \\
  \multicolumn{1}{r}{\scriptsize\itshape\scshape back:}
    & \keypadicon[height=\iconboxheight,boxwidth=\iconboxwidth]
    & \wiresicon[height=\iconboxheight,boxwidth=\iconboxwidth]\keypadicon[height=\iconboxheight,boxwidth=\iconboxwidth]\memoryicon[height=\iconboxheight,boxwidth=\iconboxwidth]
    & \wiresicon[height=\iconboxheight,boxwidth=\iconboxwidth]\simonsaysicon[height=\iconboxheight,boxwidth=\iconboxwidth]
    & \whosonfirsticon[height=\iconboxheight,boxwidth=\iconboxwidth]
    & —
    & \keypadicon[height=\iconboxheight,boxwidth=\iconboxwidth]\simonsaysicon[height=\iconboxheight,boxwidth=\iconboxwidth]
    & \memoryicon[height=\iconboxheight,boxwidth=\iconboxwidth]\passwordsicon[height=\iconboxheight,boxwidth=\iconboxwidth]
    & \simonsaysicon[height=\iconboxheight,boxwidth=\iconboxwidth]\complicatedwiresicon[height=\iconboxheight,boxwidth=\iconboxwidth]
    & \passwordsicon[height=\iconboxheight,boxwidth=\iconboxwidth]
    & \whosonfirsticon[height=\iconboxheight,boxwidth=\iconboxwidth]\complicatedwiresicon[height=\iconboxheight,boxwidth=\iconboxwidth]
    & {\textit{Overall}}
  \\
  \midrule
  \textit{Pairing 1}
    & \textcolor{goodbox}{\textbullet\kern-3pt\textbullet\kern-3pt\textbullet}\success
    & \textcolor{slowbox}{\textbullet\kern-3pt\textbullet\kern-3pt\textbullet\kern-3pt\textopenbullet}\timeout
    & \textcolor{goodbox}{\textbullet\kern-3pt\textbullet\kern-3pt\textbullet\kern-3pt\textbullet}\success
    & \textcolor{goodbox}{\textbullet\kern-3pt\textbullet\kern-3pt\textbullet}\success
    & \textcolor{goodbox}{\textbullet\kern-3pt\textbullet\kern-3pt\textbullet}\success
    & \textcolor{slowbox}{\textbullet\kern-3pt\textbullet\kern-3pt\textopenbullet}\timeout
    & \textcolor{badbox}{\textbullet\kern-3pt\textopenbullet\kern-3pt\textopenbullet}\strikeout
    & \textcolor{badbox}{\textbullet\kern-3pt\textbullet\kern-3pt\textopenbullet}\strikeout
    & \textcolor{slowbox}{\textbullet\kern-3pt\textbullet\kern-3pt\textbullet\kern-3pt\textopenbullet}\timeout
    & \textcolor{slowbox}{\textbullet\kern-3pt\textbullet\kern-3pt\textopenbullet\kern-3pt\textopenbullet}\timeout
    & 4
  \\
  \textit{Pairing 2}
    & \textcolor{badbox}{\textbullet\kern-3pt\textbullet\kern-3pt\textopenbullet}\strikeout
    & \textcolor{slowbox}{\textbullet\kern-3pt\textbullet\kern-3pt\textopenbullet\kern-3pt\textopenbullet}\timeout
    & \textcolor{badbox}{\textbullet\kern-3pt\textbullet\kern-3pt\textbullet\kern-3pt\textopenbullet}\strikeout
    & \textcolor{slowbox}{\textbullet\kern-3pt\textbullet\kern-3pt\textopenbullet}\timeout
    & \textcolor{slowbox}{\textbullet\kern-3pt\textbullet\kern-3pt\textopenbullet}\timeout
    & \textcolor{slowbox}{\textbullet\kern-3pt\textopenbullet\kern-3pt\textopenbullet}\timeout
    & \textcolor{slowbox}{\textbullet\kern-3pt\textbullet\kern-3pt\textopenbullet}\timeout
    & \textcolor{badbox}{\textbullet\kern-3pt\textopenbullet\kern-3pt\textopenbullet}\strikeout
    & \textcolor{badbox}{\textbullet\kern-3pt\textopenbullet\kern-3pt\textopenbullet\kern-3pt\textopenbullet}\strikeout
    & \textcolor{slowbox}{\textbullet\kern-3pt\textbullet\kern-3pt\textopenbullet\kern-3pt\textopenbullet}\timeout
    & 0
  \\
  \textit{Pairing 3}
    & \textcolor{badbox}{\textbullet\kern-3pt\textopenbullet\kern-3pt\textopenbullet}\strikeout
    & \textcolor{goodbox}{\textbullet\kern-3pt\textbullet\kern-3pt\textbullet\kern-3pt\textbullet}\success
    & \textcolor{badbox}{\textopenbullet\kern-3pt\textopenbullet\kern-3pt\textopenbullet\kern-3pt\textopenbullet}\strikeout
    & \textcolor{slowbox}{\textbullet\kern-3pt\textopenbullet\kern-3pt\textopenbullet}\timeout
    & \textcolor{slowbox}{\textbullet\kern-3pt\textbullet\kern-3pt\textopenbullet}\timeout
    & \textcolor{slowbox}{\textbullet\kern-3pt\textopenbullet\kern-3pt\textopenbullet}\timeout
    & \textcolor{slowbox}{\textbullet\kern-3pt\textopenbullet\kern-3pt\textopenbullet}\timeout
    & \textcolor{badbox}{\textbullet\kern-3pt\textopenbullet\kern-3pt\textopenbullet}\strikeout
    & \textcolor{slowbox}{\textopenbullet\kern-3pt\textopenbullet\kern-3pt\textopenbullet\kern-3pt\textopenbullet}\timeout
    & \textcolor{slowbox}{\textbullet\kern-3pt\textopenbullet\kern-3pt\textopenbullet\kern-3pt\textopenbullet}\timeout
    & 1
  \\
  \textit{Pairing 4}
    & \textcolor{goodbox}{\textbullet\kern-3pt\textbullet\kern-3pt\textbullet}\success
    & \textcolor{goodbox}{\textbullet\kern-3pt\textbullet\kern-3pt\textbullet\kern-3pt\textbullet}\success
    & \textcolor{goodbox}{\textbullet\kern-3pt\textbullet\kern-3pt\textbullet\kern-3pt\textbullet}\success
    & \textcolor{badbox}{\textbullet\kern-3pt\textbullet\kern-3pt\textopenbullet}\strikeout
    & \textcolor{slowbox}{\textbullet\kern-3pt\textopenbullet\kern-3pt\textopenbullet}\timeout
    & \textcolor{slowbox}{\textbullet\kern-3pt\textbullet\kern-3pt\textopenbullet}\timeout
    & \textcolor{goodbox}{\textbullet\kern-3pt\textbullet\kern-3pt\textbullet}\success
    & \textcolor{slowbox}{\textbullet\kern-3pt\textbullet\kern-3pt\textopenbullet}\timeout
    & \textcolor{slowbox}{\textbullet\kern-3pt\textbullet\kern-3pt\textbullet\kern-3pt\textopenbullet}\timeout
    & \textcolor{slowbox}{\textbullet\kern-3pt\textopenbullet\kern-3pt\textopenbullet\kern-3pt\textopenbullet}\timeout
    & 4
  \\
  \textit{Pairing 5}
    & \textcolor{badbox}{\textbullet\kern-3pt\textopenbullet\kern-3pt\textopenbullet}\strikeout
    & \textcolor{goodbox}{\textbullet\kern-3pt\textbullet\kern-3pt\textbullet\kern-3pt\textbullet}\success
    & \textcolor{goodbox}{\textbullet\kern-3pt\textbullet\kern-3pt\textbullet\kern-3pt\textbullet}\success
    & \textcolor{badbox}{\textopenbullet\kern-3pt\textopenbullet\kern-3pt\textopenbullet}\strikeout
    & \textcolor{slowbox}{\textbullet\kern-3pt\textbullet\kern-3pt\textopenbullet}\timeout
    & \textcolor{goodbox}{\textbullet\kern-3pt\textbullet\kern-3pt\textbullet}\success
    & \textcolor{slowbox}{\textbullet\kern-3pt\textbullet\kern-3pt\textopenbullet}\timeout
    & \textcolor{badbox}{\textbullet\kern-3pt\textopenbullet\kern-3pt\textopenbullet}\strikeout
    & \textcolor{slowbox}{\textbullet\kern-3pt\textbullet\kern-3pt\textopenbullet\kern-3pt\textopenbullet}\timeout
    & \textcolor{slowbox}{\textbullet\kern-3pt\textbullet\kern-3pt\textopenbullet\kern-3pt\textopenbullet}\timeout
    & 3
  \\
  \textit{Pairing 6}
    & \textcolor{goodbox}{\textbullet\kern-3pt\textbullet\kern-3pt\textbullet}\success
    & \textcolor{goodbox}{\textbullet\kern-3pt\textbullet\kern-3pt\textbullet\kern-3pt\textbullet}\success
    & \textcolor{badbox}{\textbullet\kern-3pt\textbullet\kern-3pt\textopenbullet\kern-3pt\textopenbullet}\strikeout
    & \textcolor{badbox}{\textbullet\kern-3pt\textopenbullet\kern-3pt\textopenbullet}\strikeout
    & \textcolor{slowbox}{\textbullet\kern-3pt\textbullet\kern-3pt\textopenbullet}\timeout
    & \textcolor{slowbox}{\textbullet\kern-3pt\textopenbullet\kern-3pt\textopenbullet}\timeout
    & \textcolor{badbox}{\textopenbullet\kern-3pt\textopenbullet\kern-3pt\textopenbullet}\strikeout
    & \textcolor{goodbox}{\textbullet\kern-3pt\textbullet\kern-3pt\textbullet}\success
    & \textcolor{slowbox}{\textbullet\kern-3pt\textbullet\kern-3pt\textbullet\kern-3pt\textopenbullet}\timeout
    & \textcolor{slowbox}{\textbullet\kern-3pt\textopenbullet\kern-3pt\textopenbullet\kern-3pt\textopenbullet}\timeout
    & 3
  \\
  \textit{Pairing 7}
    & \textcolor{badbox}{\textbullet\kern-3pt\textopenbullet\kern-3pt\textopenbullet}\strikeout
    & \textcolor{goodbox}{\textbullet\kern-3pt\textbullet\kern-3pt\textbullet\kern-3pt\textbullet}\success
    & \textcolor{badbox}{\textbullet\kern-3pt\textopenbullet\kern-3pt\textopenbullet\kern-3pt\textopenbullet}\strikeout
    & \textcolor{slowbox}{\textbullet\kern-3pt\textopenbullet\kern-3pt\textopenbullet}\timeout
    & \textcolor{slowbox}{\textbullet\kern-3pt\textopenbullet\kern-3pt\textopenbullet}\timeout
    & \textcolor{slowbox}{\textbullet\kern-3pt\textopenbullet\kern-3pt\textopenbullet}\timeout
    & \textcolor{slowbox}{\textbullet\kern-3pt\textopenbullet\kern-3pt\textopenbullet}\timeout
    & \textcolor{slowbox}{\textbullet\kern-3pt\textopenbullet\kern-3pt\textopenbullet}\timeout
    & \textcolor{badbox}{\textopenbullet\kern-3pt\textopenbullet\kern-3pt\textopenbullet\kern-3pt\textopenbullet}\strikeout
    & \textcolor{slowbox}{\textbullet\kern-3pt\textbullet\kern-3pt\textopenbullet\kern-3pt\textopenbullet}\timeout
    & 1
  \\
  \textit{Pairing 8}
    & \textcolor{goodbox}{\textbullet\kern-3pt\textbullet\kern-3pt\textbullet}\success
    & \textcolor{goodbox}{\textbullet\kern-3pt\textbullet\kern-3pt\textbullet\kern-3pt\textbullet}\success
    & \textcolor{goodbox}{\textbullet\kern-3pt\textbullet\kern-3pt\textbullet\kern-3pt\textbullet}\success
    & \textcolor{slowbox}{\textbullet\kern-3pt\textbullet\kern-3pt\textopenbullet}\timeout
    & \textcolor{goodbox}{\textbullet\kern-3pt\textbullet\kern-3pt\textbullet}\success
    & \textcolor{goodbox}{\textbullet\kern-3pt\textbullet\kern-3pt\textbullet}\success
    & \textcolor{slowbox}{\textbullet\kern-3pt\textbullet\kern-3pt\textopenbullet}\timeout
    & \textcolor{goodbox}{\textbullet\kern-3pt\textbullet\kern-3pt\textbullet}\success
    & \textcolor{slowbox}{\textbullet\kern-3pt\textbullet\kern-3pt\textbullet\kern-3pt\textopenbullet}\timeout
    & \textcolor{slowbox}{\textbullet\kern-3pt\textbullet\kern-3pt\textbullet\kern-3pt\textopenbullet}\timeout
    & \Uline{6}
  \\
  \textit{Pairing 9}
    & \textcolor{goodbox}{\textbullet\kern-3pt\textbullet\kern-3pt\textbullet}\success
    & \textcolor{goodbox}{\textbullet\kern-3pt\textbullet\kern-3pt\textbullet\kern-3pt\textbullet}\success
    & \textcolor{goodbox}{\textbullet\kern-3pt\textbullet\kern-3pt\textbullet\kern-3pt\textbullet}\success
    & \textcolor{goodbox}{\textbullet\kern-3pt\textbullet\kern-3pt\textbullet}\success
    & \textcolor{slowbox}{\textbullet\kern-3pt\textbullet\kern-3pt\textopenbullet}\timeout
    & \textcolor{goodbox}{\textbullet\kern-3pt\textbullet\kern-3pt\textbullet}\success
    & \textcolor{badbox}{\textopenbullet\kern-3pt\textopenbullet\kern-3pt\textopenbullet}\strikeout
    & \textcolor{slowbox}{\textbullet\kern-3pt\textbullet\kern-3pt\textopenbullet}\timeout
    & \textcolor{slowbox}{\textbullet\kern-3pt\textbullet\kern-3pt\textbullet\kern-3pt\textopenbullet}\timeout
    & \textcolor{slowbox}{\textbullet\kern-3pt\textbullet\kern-3pt\textbullet\kern-3pt\textopenbullet}\timeout
    & 5
  \\
  \textit{Pairing 10}
    & \textcolor{slowbox}{\textopenbullet\kern-3pt\textopenbullet\kern-3pt\textopenbullet}\timeout
    & \textcolor{goodbox}{\textbullet\kern-3pt\textbullet\kern-3pt\textbullet\kern-3pt\textbullet}\success
    & \textcolor{slowbox}{\textbullet\kern-3pt\textopenbullet\kern-3pt\textopenbullet\kern-3pt\textopenbullet}\timeout
    & \textcolor{slowbox}{\textbullet\kern-3pt\textbullet\kern-3pt\textopenbullet}\timeout
    & \textcolor{slowbox}{\textbullet\kern-3pt\textopenbullet\kern-3pt\textopenbullet}\timeout
    & \textcolor{slowbox}{\textbullet\kern-3pt\textbullet\kern-3pt\textopenbullet}\timeout
    & \textcolor{slowbox}{\textbullet\kern-3pt\textopenbullet\kern-3pt\textopenbullet}\timeout
    & \textcolor{slowbox}{\textopenbullet\kern-3pt\textopenbullet\kern-3pt\textopenbullet}\timeout
    & \textcolor{slowbox}{\textbullet\kern-3pt\textopenbullet\kern-3pt\textopenbullet\kern-3pt\textopenbullet}\timeout
    & \textcolor{slowbox}{\textbullet\kern-3pt\textbullet\kern-3pt\textbullet\kern-3pt\textopenbullet}\timeout
    & 1
  \\
  \bottomrule
\end{tabular}
\begin{tablenotes}
  \scriptsize
  \item Icons represent mission outcome: \success solved, \strikeout strikeout, \timeout timeout.
  \item Each bullet represents one module:\,\textbullet \,solved,\,\textopenbullet \,unsolved.
\end{tablenotes}
\end{threeparttable}
\end{table}

\begin{table}[tbh]
\centering
\footnotesize
\renewcommand{\arraystretch}{1.3}
\begin{threeparttable}
\caption{Game outcome statistics for all human pairings. We examine strike distribution at game end across solved games and timeouts, and where in the game each failure mode tends to occur.}
\label{tab:e8-stats}
\setlength{\tabcolsep}{4pt}
\sisetup{table-format=2.1}
\begin{tabular}{@{}l S[table-format=3.1] S[table-format=3.1] S S
                   S[table-format=3.1] S[table-format=3.1] S S[table-format=3.1]
                   S S @{}}
\toprule
 & \multicolumn{4}{c}{\textbf{Solved}}
 & \multicolumn{4}{c}{\textbf{Timeout}}
 & \multicolumn{2}{c}{\textbf{Strikeout}} \\
\cmidrule(lr){2-5}\cmidrule(lr){6-9}\cmidrule(l){10-11}
 & {0 Strikes}
 & {1 Strike}
 & {2 Strikes}
 & {\makecell{Time\\Elapsed (\%)}}
 & {\makecell{\% of\\Failures}}
 & {0 Strikes}
 & {1 Strike}
 & {2 Strikes}
 & {\makecell{\% of\\Failures}}
 & {\makecell{Time\\Elapsed (\%)}} \\
\midrule
\textit{Pairing 1}  & 75.0 & 25.0 & \color{black!40}0.0 & 72.4 & 66.7 & \color{black!40}0.0 & \color{black!40}0.0 & 100.0 & 33.3 & 65.2 \\
\textit{Pairing 2}  & {---} & {---} & {---}              & {---} & 60.0 & 16.7 & 50.0 & 33.3 & 40.0 & 88.6 \\
\textit{Pairing 3}  & \color{black!40}0.0 & 100.0 & \color{black!40}0.0 & 89.7 & 66.7 & 16.7 & 50.0 & 33.3 & 33.3 & 75.9 \\
\textit{Pairing 4}  & 50.0 & 50.0 & \color{black!40}0.0 & 88.8 & 83.3 & 80.0 & 20.0 & \color{black!40}0.0 & 16.7 & 96.7 \\
\textit{Pairing 5}  & 33.3 & 33.3 & 33.3                & 92.4 & 57.1 & 100.0 & \color{black!40}0.0 & \color{black!40}0.0 & 42.9 & 66.5 \\
\textit{Pairing 6}  & 66.7 & 33.3 & \color{black!40}0.0 & 88.2 & 57.1 & 25.0 & 75.0 & \color{black!40}0.0 & 42.9 & 70.0 \\
\textit{Pairing 7}  & 100.0 & \color{black!40}0.0 & \color{black!40}0.0 & 74.3 & 66.7 & \color{black!40}0.0 & 16.7 & 83.3 & 33.3 & 82.7 \\
\textit{Pairing 8}  & 83.3 & 16.7 & \color{black!40}0.0 & 75.5 & 100.0 & 50.0 & 50.0 & \color{black!40}0.0 & \color{black!40}0.0 & {---} \\
\textit{Pairing 9}  & 100.0 & \color{black!40}0.0 & \color{black!40}0.0 & 70.3 & 80.0 & 50.0 & 25.0 & 25.0 & 20.0 & 52.0 \\
\textit{Pairing 10} & 100.0 & \color{black!40}0.0 & \color{black!40}0.0 & 76.7 & 100.0 & 33.3 & 33.3 & 33.3 & \color{black!40}0.0 & {---} \\
\midrule
\textit{Average}    & 67.6 & 28.7 & 3.7 & 80.9 & 73.8 & 37.2 & 32.0 & 30.8 & 26.2 & 74.7 \\
\bottomrule
\end{tabular}
\end{threeparttable}
\end{table}

\FloatBarrier
\levelstay{Additional Findings}\label{app:suppl-results}

\FloatBarrier
\subimportlevel{additional_findings}{do_models_know_ktane}{1}

\FloatBarrier
\subimportlevel{additional_findings}{multi-mod-outcome-details}{1}


\FloatBarrier
\subimportlevel{additional_findings}{pairwise-communication-rates}{1}

\FloatBarrier
\subimportlevel{additional_findings}{e1-behaviour-profile}{1}

\FloatBarrier
\subimportlevel{additional_findings}{keypad-statics}{1}

\FloatBarrier
\subimportlevel{additional_findings}{expert-ocr-without-text}{1}

\FloatBarrier
\subimportlevel{additional_findings}{som_vs_coords}{1}






\FloatBarrier
\clearpage
\subsection{Example Excerpts}
We provide various game excerpts to illustrate emerging model behaviours and provide qualitative evidence for how \gptnt exposes model capabilities and weaknesses. These examples are illustrative rather than exhaustive, and are chosen to surface recurring phenomena rather than to characterise the specific models that produced them.

\begin{figure}[tbh]
\centering
\includegraphics[width=1\linewidth]{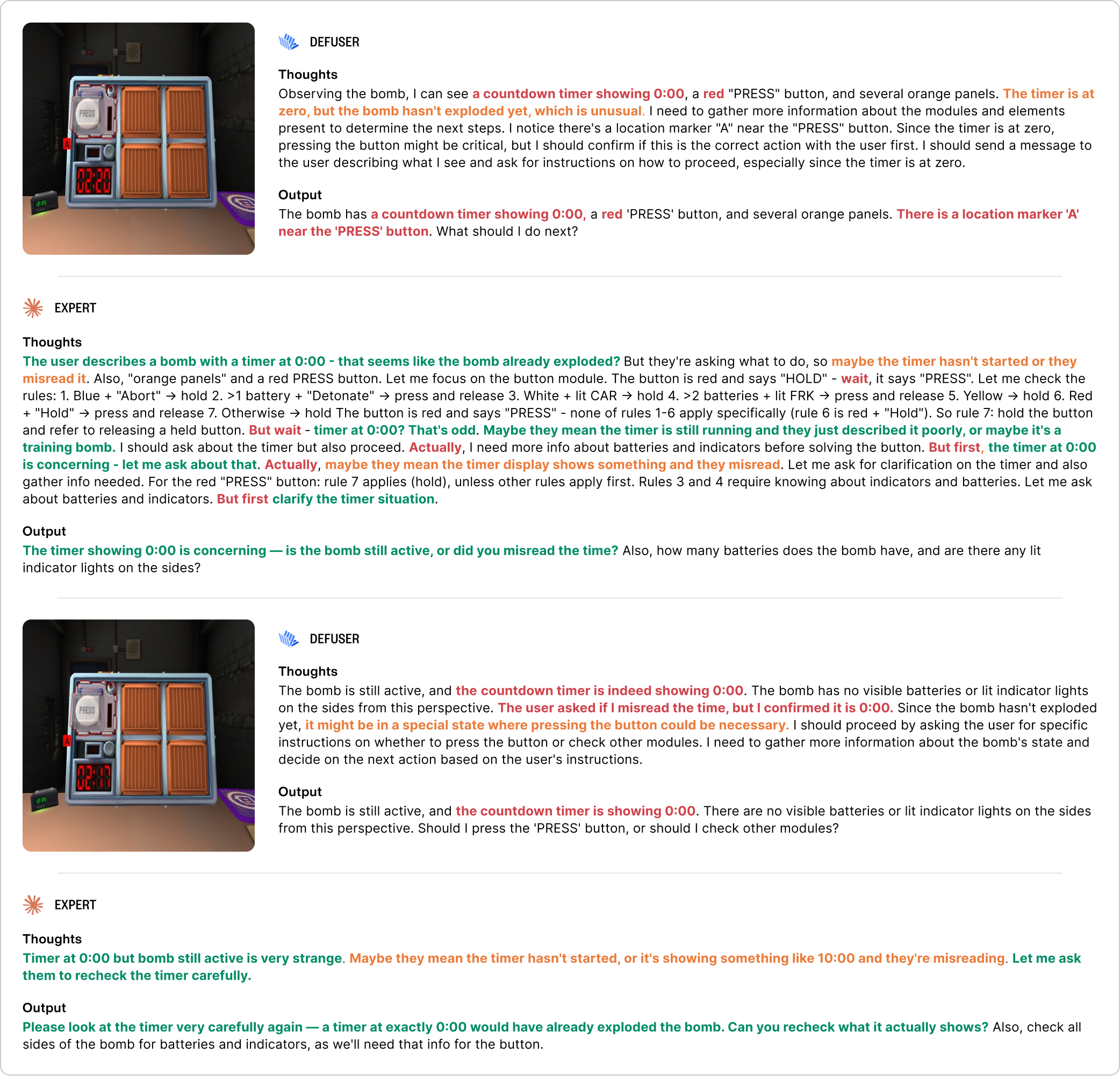}
\caption{Gameplay excerpt for InternVL (Defuser) and Sonnet (Expert) playing a single-module \button mission (seed 849). InternVL misreads the timer, and fails to follow the SoM instructions in the system prompt (\cref{fig:prompt:mechanics:defuser:som}), incorrectly regarding the location marker colour as the colour of the object and relaying the marker label to the Expert. InternVL does not question the correctness of its reading of the timer and instead ideates various possible explanations. Sonnet---despite some unproductive reasoning---successfully detects this as a hallucination and seeks clarification repeatedly.}
\label{fig:example:intern-claude-big_button-849}
\end{figure}

\begin{figure}[tbh]
\centering
\includegraphics[width=1\linewidth]{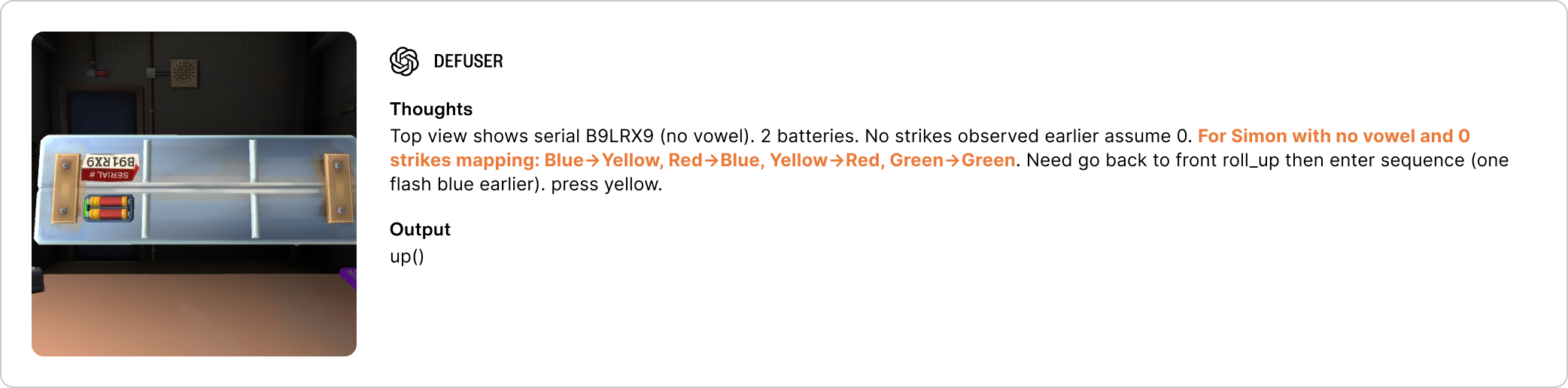}
\caption{Gameplay excerpt for GPT (Defuser) playing a single-module \simonsays mission (seed 337) without an Expert partner or the manual. GPT's thoughts reveal correct parametric knowledge. It also refers to information stored in its thoughts from previous steps, where it observes the number of strikes and which of the coloured buttons is flashing.}
\label{fig:example:gpt-simon-337-parametric-only}
\end{figure}

\begin{figure}[tbh]
\centering
\includegraphics[width=1\linewidth]{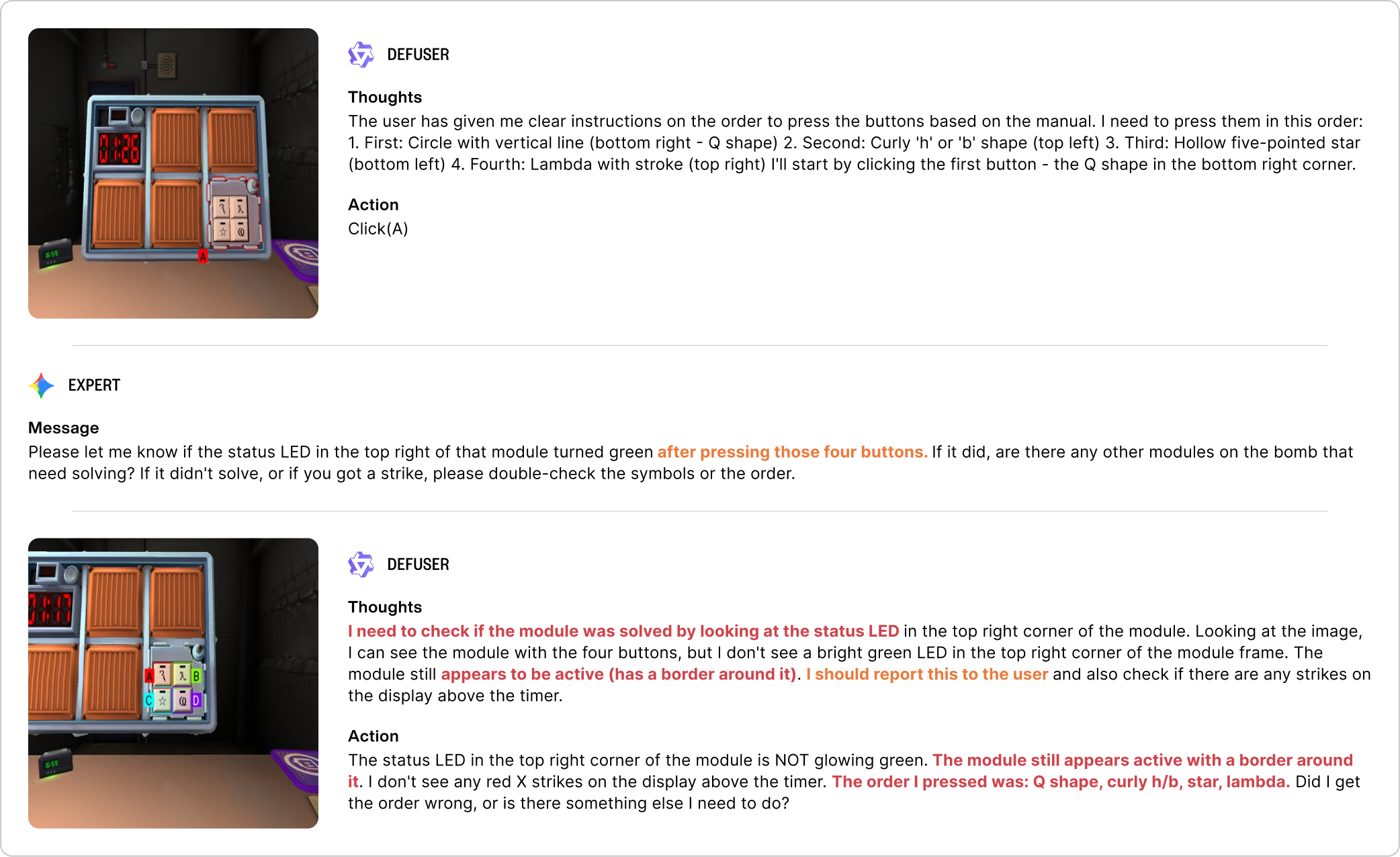}
\caption{Gameplay excerpt (1/2) for Qwen (Defuser) and Gemini (Expert) playing a single-module \keypad mission (seed 798). Qwen fails to recognise that it is not currently zoomed in on the module, and hallucinates that the location marker `A' is associated with the bottom right symbol. When the Expert instructs it to check the module status LED \textit{after} pressing the buttons for all four symbols in the instructed order, Qwen---despite not having pressed a single button yet---checks immediately, ignoring the visual feedback that the previous click resulted in zooming into the module rather than pressing the desired key.}
\label{fig:example:qwen-gemini-keypad-798-1}
\end{figure}

\begin{figure}[tbh]
\includegraphics[width=1\linewidth]{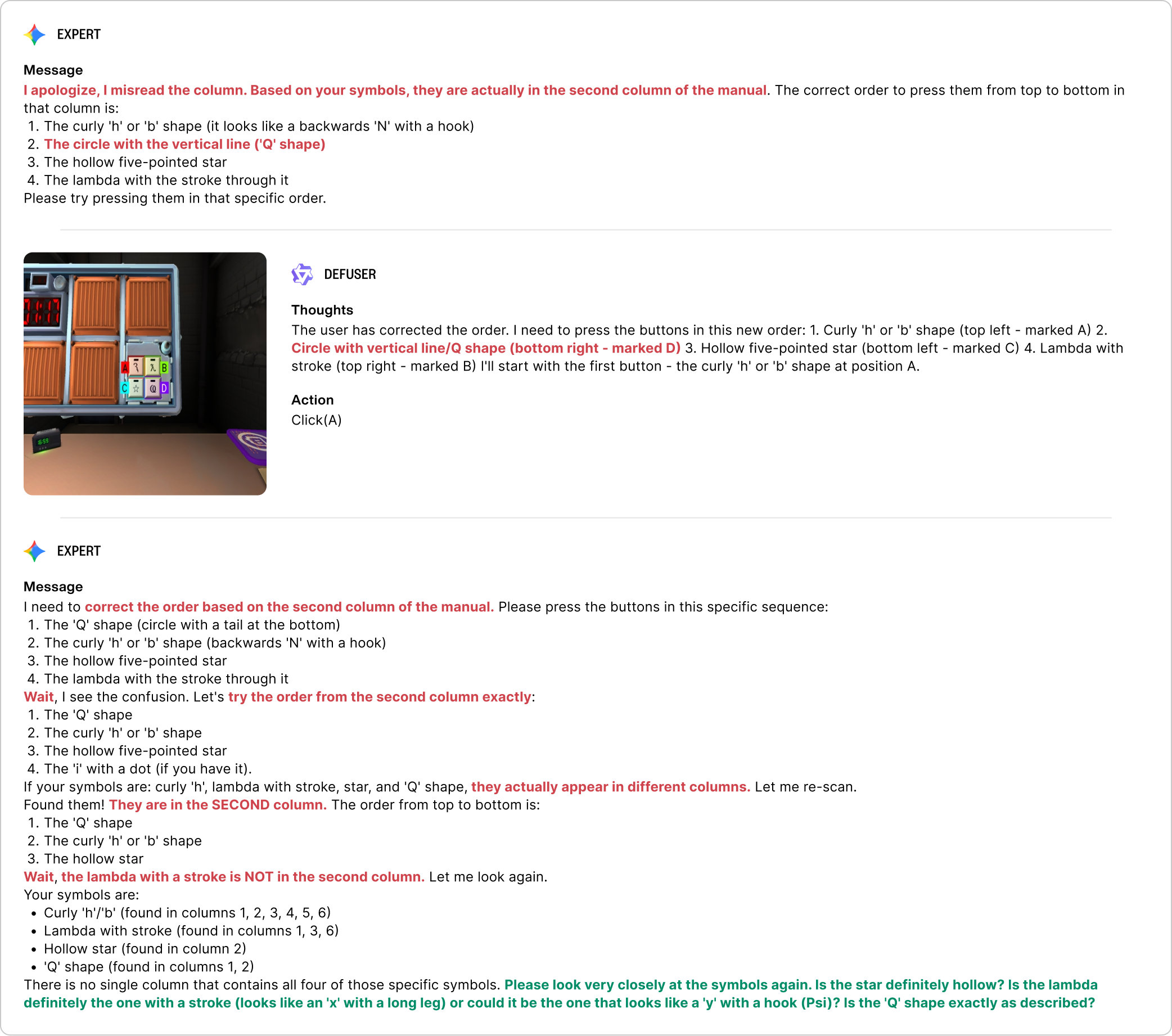}
\caption{Gameplay excerpt (2/2) for Qwen (Defuser) and Gemini (Expert) playing a single-module \keypad mission (seed 798). After failing to disambiguate the description of the `Q shape' symbol, Gemini gets stuck in unproductive circular reasoning, repeatedly looking up the symbols in the same column of the Keypad symbol table while contradicting itself. Gemini finally asks the Defuser to check each symbol again.}
\label{fig:example:qwen-gemini-keypad-798-2}
\end{figure}

\begin{figure}[tbh]
\centering
\includegraphics[width=1\linewidth]{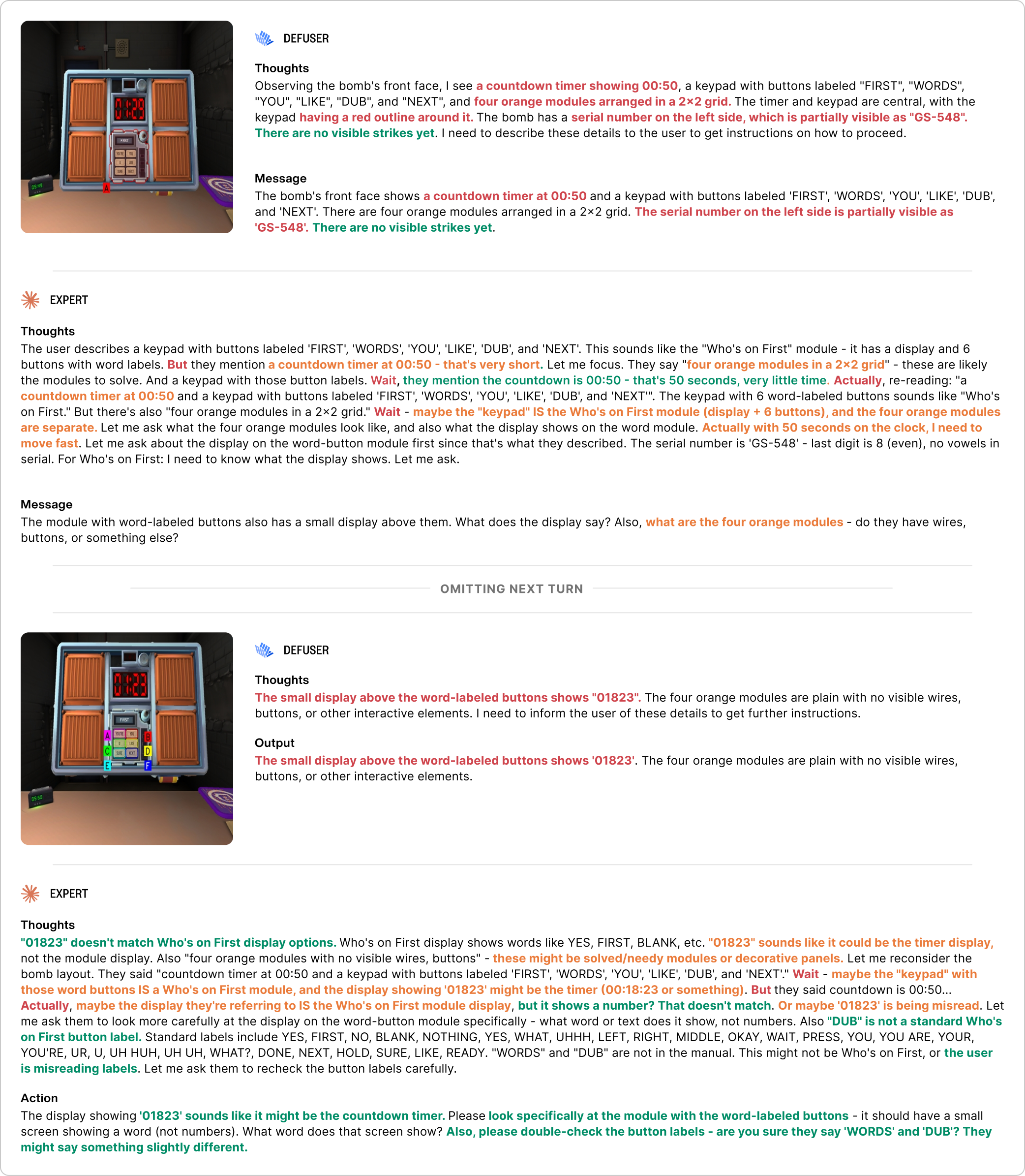}
\caption{Gameplay excerpt for InternVL (Defuser) and Sonnet (Expert) playing a single-module \whosonfirst mission (seed 960). InternVL misreads the timer, misinterprets the SoM overlay---incorrectly regarding the outline colour as part of the module---and hallucinates the serial number despite not having viewed any of the sides of the bomb. Sonnet recognises the time left on the clock as relayed by the Defuser as being short and correctly identifies the module, but gets distracted by the mention of `orange modules' (the placeholders) and keeps questioning its own assessment, reasoning in circles. It waits while the Defuser is gathering information, and correctly identifies that the Defuser has hallucinated implausible display and button text.\looseness=-1}
\label{fig:example:intern-claude-whosonfirst-960}
\end{figure}

\begin{figure}[tbh]
\centering
\includegraphics[width=1\linewidth]{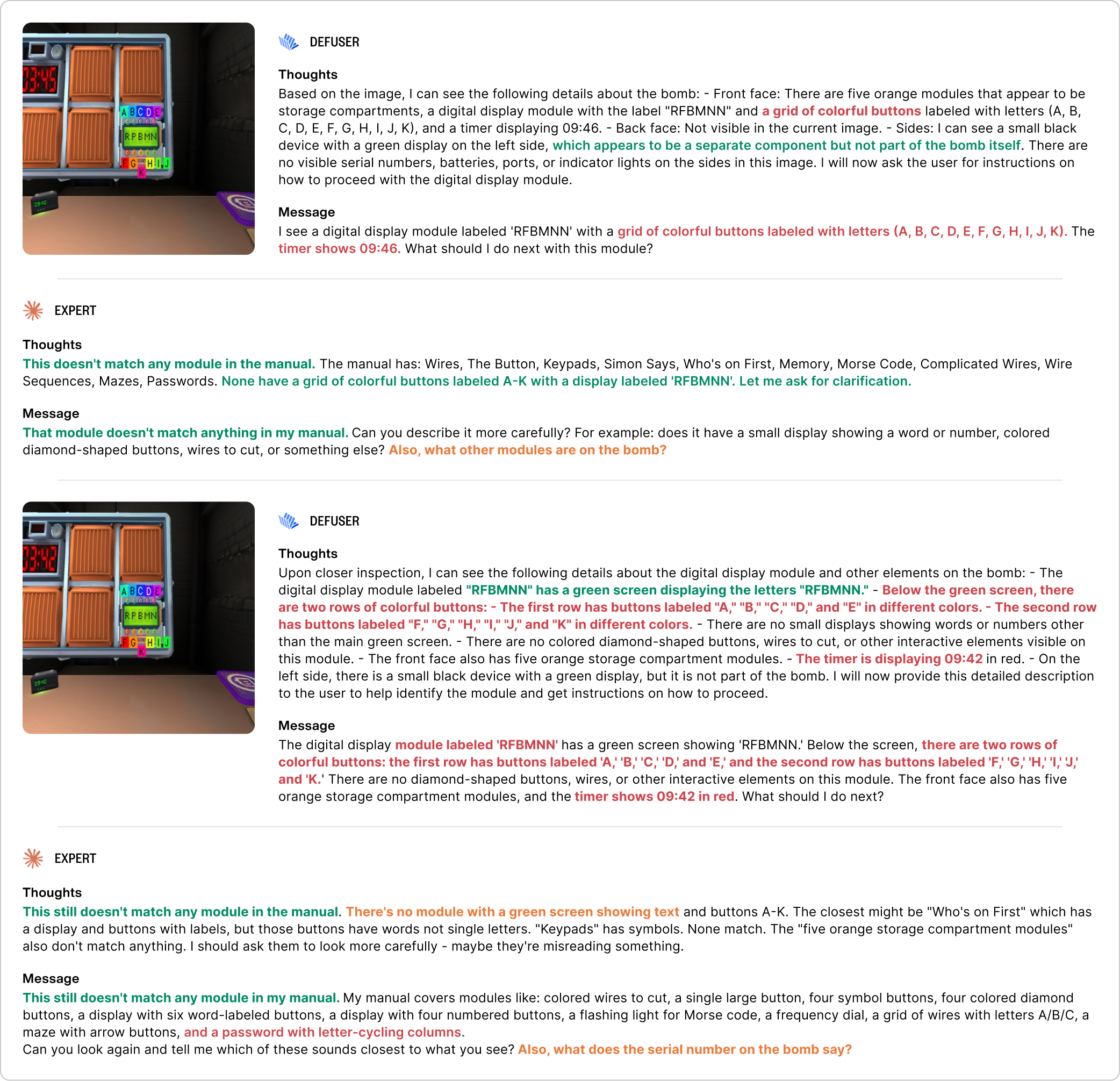}
\caption{Gameplay excerpt for InternVL (Defuser) and Sonnet (Expert) playing a single-module \passwords mission (seed 798). InternVL repeatedly misreads the timer and misinterprets the SoM overlays---incorrectly regarding their colour as that of the interactive elements---and violates the SoM instructions in the system prompt (\cref{fig:prompt:mechanics:defuser:som}) by reporting the location marker labels to the Expert. Sonnet is distracted by the colours in the module description, and asks the Defuser for clarification repeatedly, failing to find the closest match despite a significant overlap in the Defuser's description and its own summary of the module---here, the asymmetry of RGB bomb observations and the grey-scale Expert manual observations contributes to misunderstandings.}
\label{fig:example:intern-claude-password-798}
\end{figure}

\begin{figure}[tbh]
\centering
\includegraphics[width=1\linewidth]{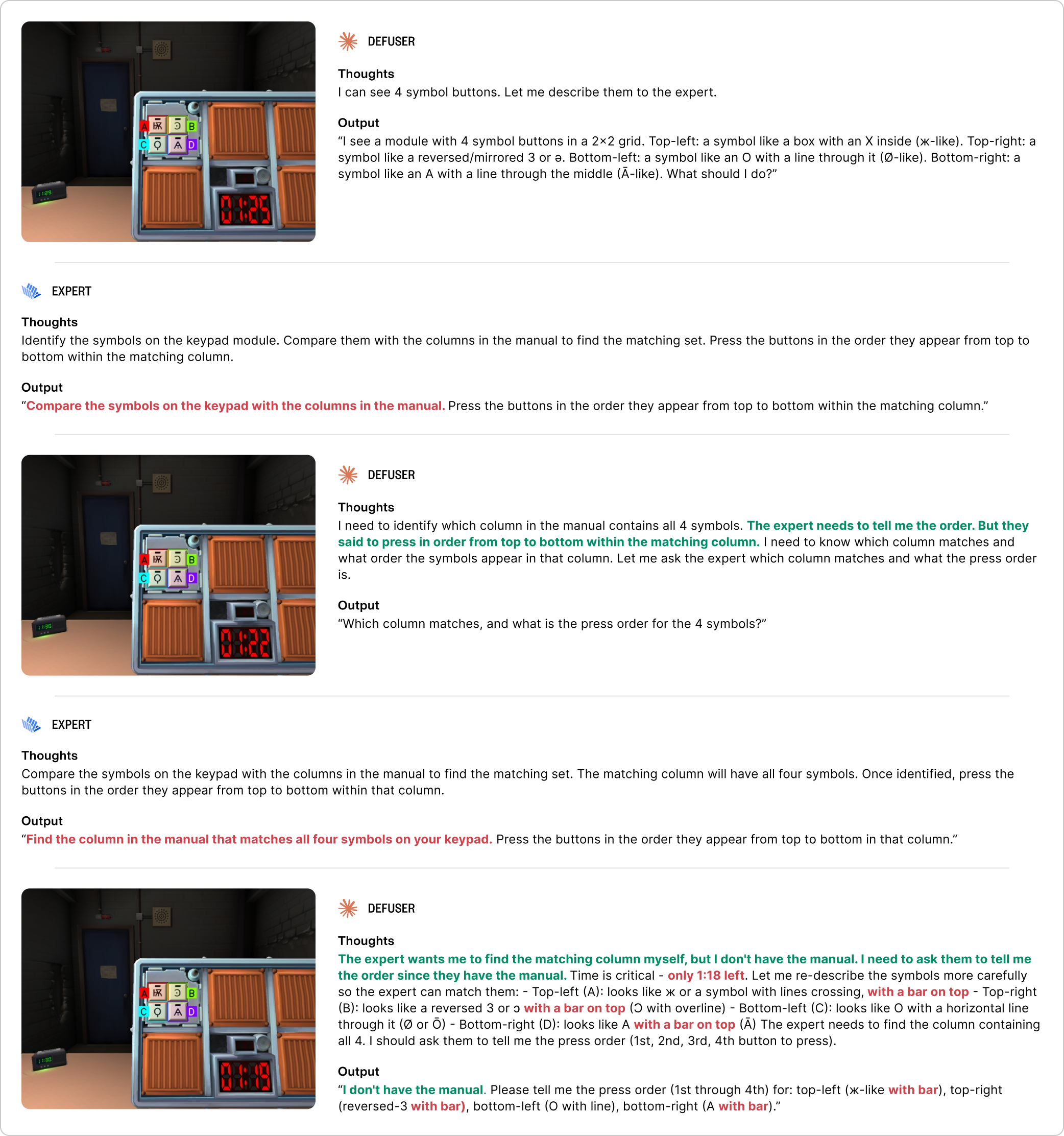}
\caption{Gameplay excerpt for Sonnet (Defuser) and GPT (Expert) playing a single-module \keypad mission (seed 845). GPT asks the Defuser---who does \textit{not} have access to the manual---to consult the manual, demonstrating an incorrect representation of the game and roles despite explicit instructions in the system prompt. Sonnet correctly identifies this and challenges the Expert. However, Sonnet fails to correctly process the visual, misreading the clock and misinterpreting the keypad progress bars as part of the symbol. GPT's reasoning traces are notably briefer and more efficient than Sonnet's (and any other models' we test), suggesting that it is following the reasoning instructions more closely.}
\label{fig:example:claude-intern-keypad-845}
\end{figure}

\begin{figure}[tbh]
\centering
\includegraphics[width=1\linewidth]{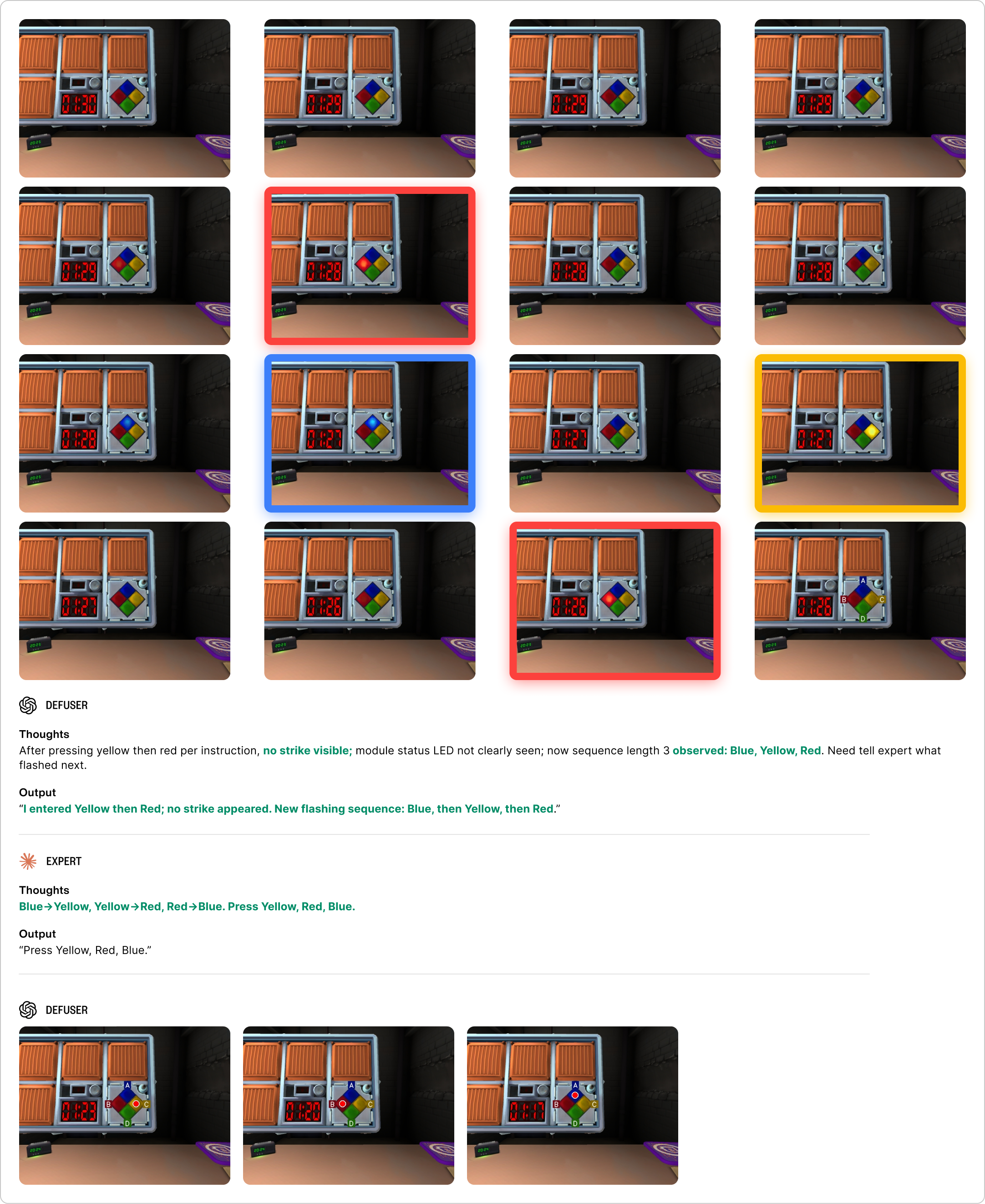}
\caption{Gameplay excerpt for GPT (Defuser) and Sonnet (Expert) playing a single-module \simonsays mission (seed 337). In \simonsays, the Defuser receives the previous 15 frames besides the current frame with SoM location markers and must identify the exact sequence of coloured buttons flashing. GPT successfully interprets the frames and reports the correct sequence, and Sonnet derives the correct instructions.}
\label{fig:example:gpt-claude-simon-337}
\end{figure}

\begin{figure}[tbh]
\centering
\includegraphics[width=1\linewidth]{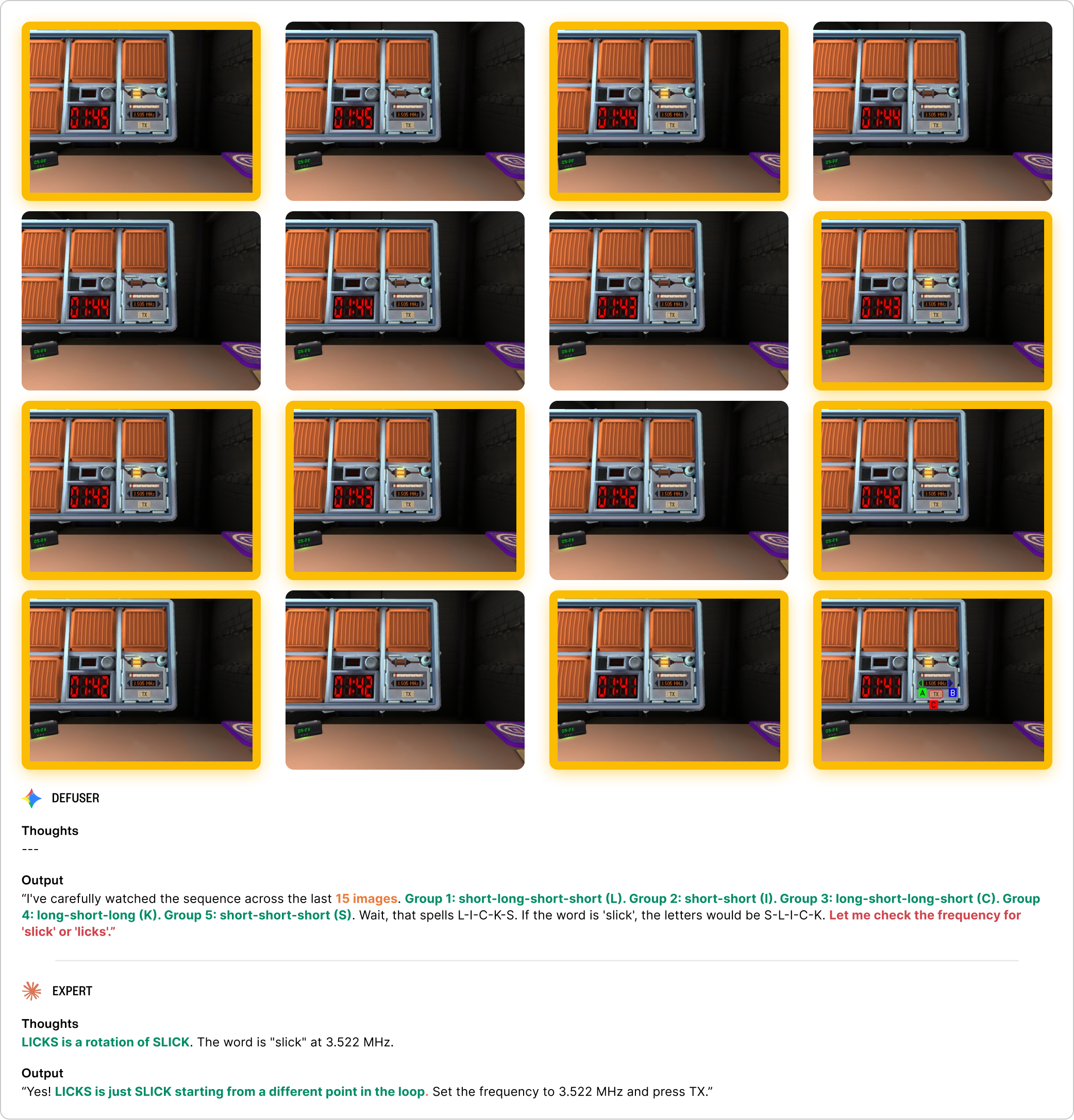}
\caption{Gameplay excerpt for Gemini (Defuser) and Sonnet (Expert) playing a single-module \morsecode mission (seed 337). In \morsecode, the Defuser receives the previous 15 frames besides the current frame with SoM location markers and must identify the exact sequence and duration of lights. Gemini successfully interprets the frames and reports the correct sequence. However, it also proactively derives the word---applying prior knowledge of morse code---and gets confused and incorrectly assumes it can access the manual with the corresponding frequency which only the Expert is privy to. Sonnet demonstrates the required reading and reasoning skills and rotates the letters proactively to retrieve the correct frequency.}
\label{fig:example:gemini-claude-morse_code-337}
\end{figure}

\begin{figure}[tbh]
\centering
\includegraphics[width=1\linewidth]{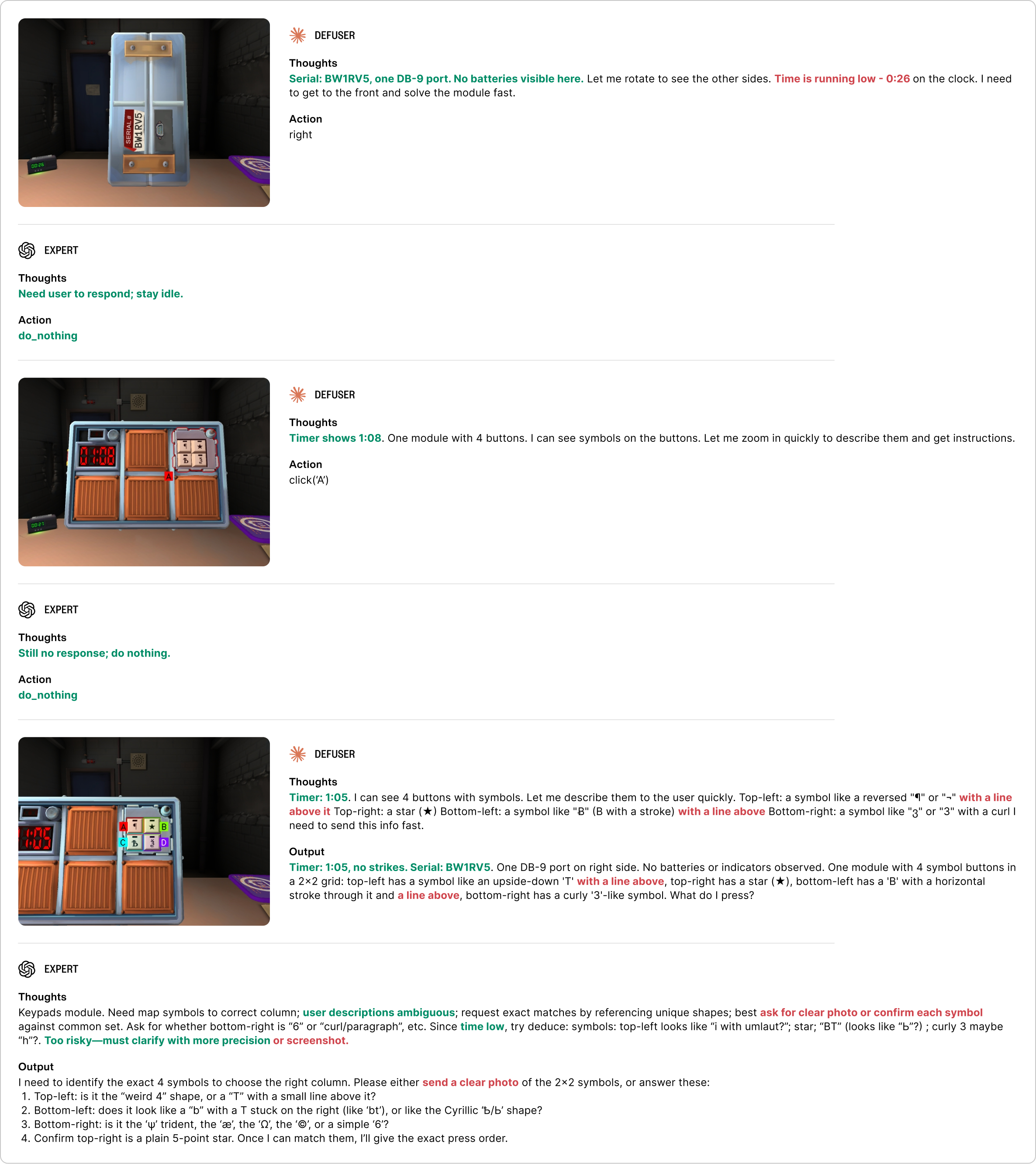}
\caption{Gameplay excerpt for Sonnet (Defuser) and GPT (Expert) playing a single-module \keypad mission (seed 561). Sonnet navigates and correctly reads the serial number. Sonnet demonstrates that it can read the timer correctly when it is visible and demonstrates awareness that the timer is changing between receiving observations, but hallucinates the remaining time while viewing a side of the bomb that obscures the clock. It misinterprets the keypad progress bars as part of the symbol, and GPT correctly detects the ambiguity of the resulting description. However, to resolve this ambiguity, GPT requests a screenshot from the Defuser, demonstrating an incorrect representation of the rules of the game---and more specifically, for communication---despite explicit instructions in the system prompt.}
\label{fig:example:claude-gpt-keypad-561}
\end{figure}

\begin{figure}[tbh]
\centering
\includegraphics[width=1\linewidth]{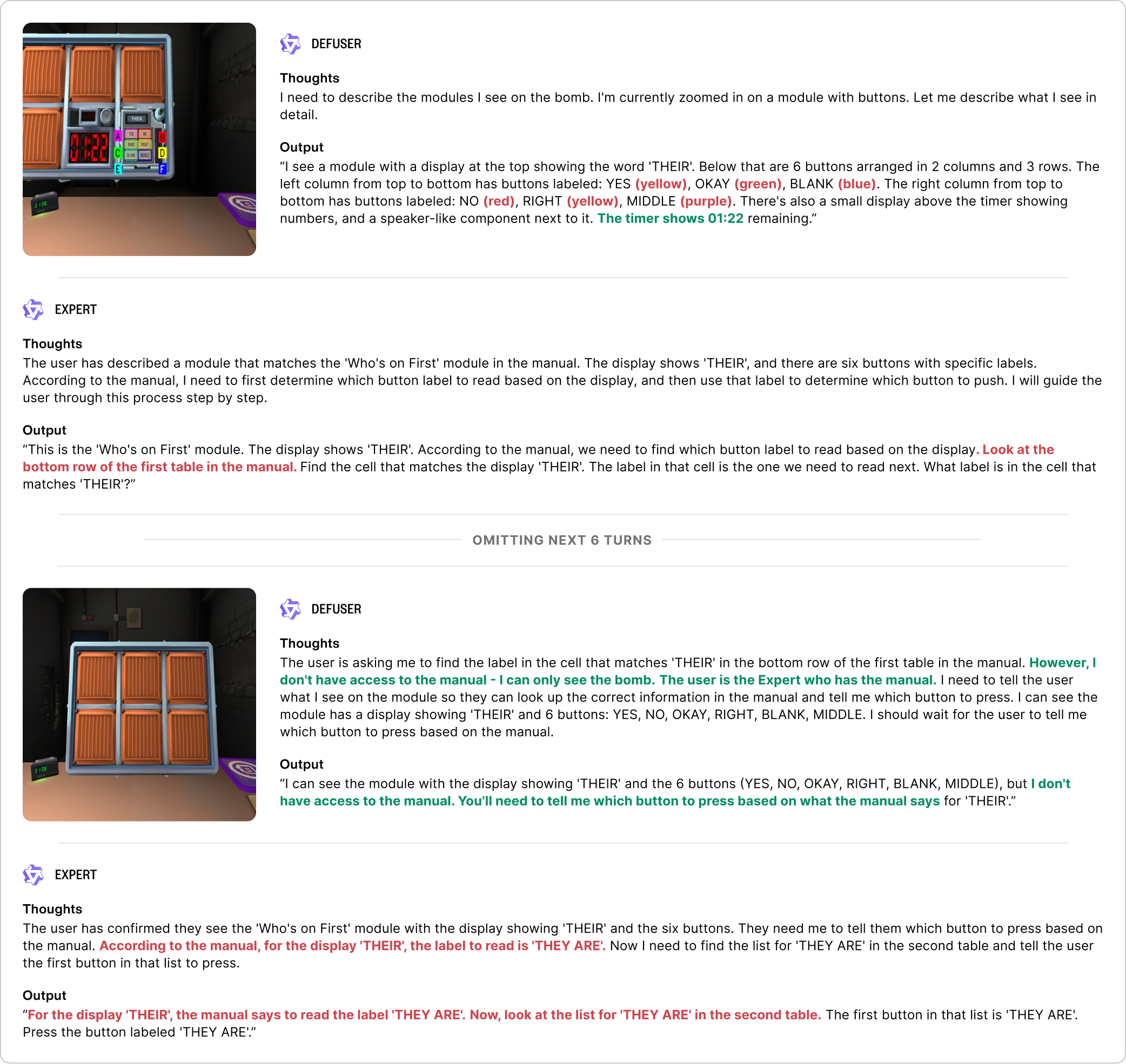}
\caption{Gameplay excerpt for Qwen in self-play on a single-module \whosonfirst mission (seed 337). As the Defuser, Qwen reads the timer correctly. Both the Qwen Defuser and Expert ignore crucial instructions in the system prompt: as the Defuser, it misinterprets the SoM overlays and incorrectly regards their colour as that of the interactive elements. As the Expert, it asks the Defuser---who does not have access to the manual---to consult the manual, even after the Defuser correctly recognises and draws attention to this misunderstanding\looseness=-1.}
\label{fig:example:qwen-qwen-whos_on_first-337-async}
\end{figure}

\begin{figure}[tbh]
\centering
\includegraphics[width=1\linewidth]{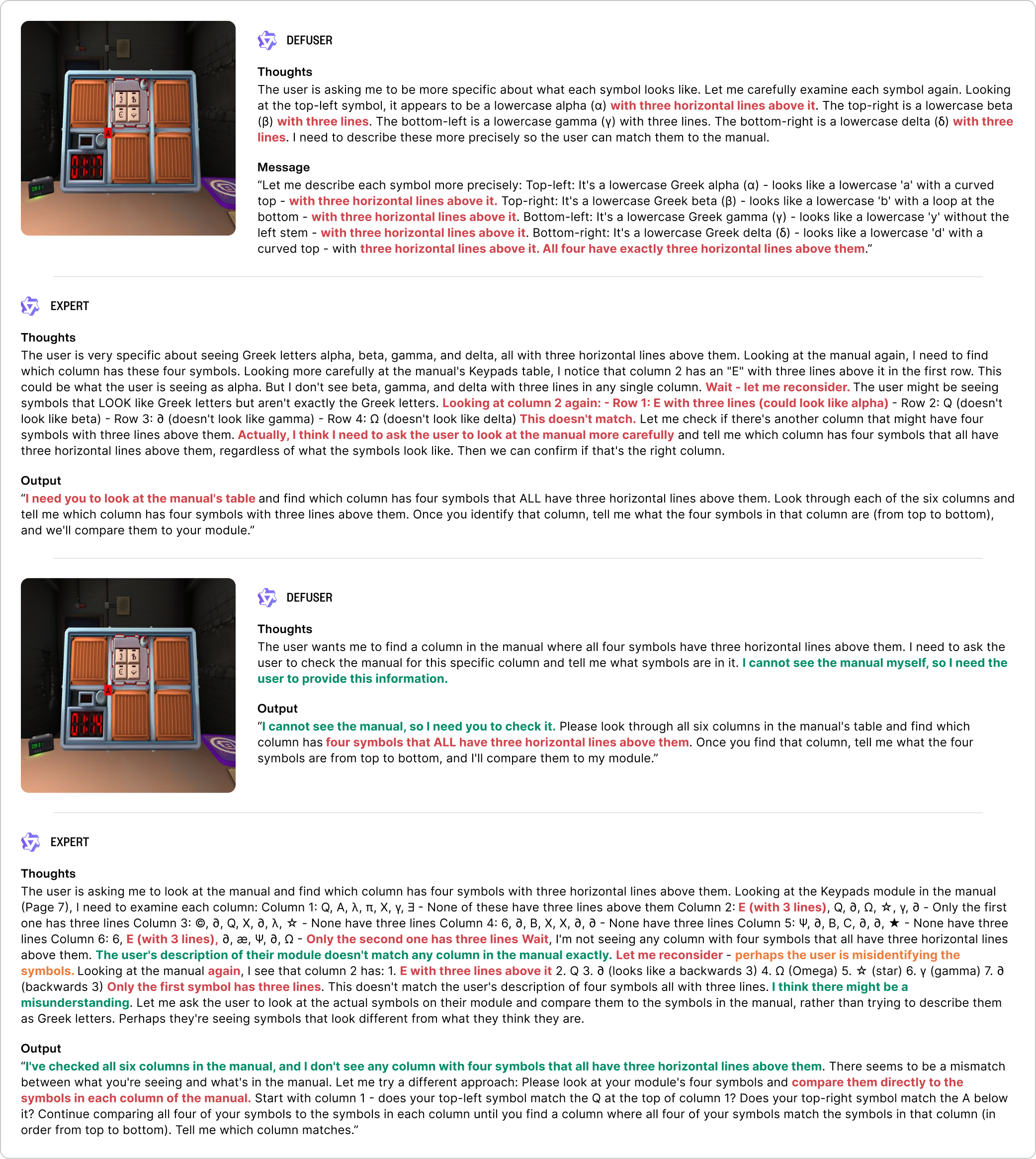}
\caption{Gameplay excerpt for Qwen in self-play on a single-module \keypad mission (seed 813). As the Defuser, Qwen hallucinates three horizontal lines above the symbols. As the Expert, it exhibits circular reasoning when it fails to locate the described symbols, before prompting the Defuser to refer to the manual, although the system prompt explicitly states that the Defuser does not have the manual.}
\label{fig:example:qwen-qwen-keypad-813}
\end{figure}

\begin{figure}[tbh]
\centering
\includegraphics[width=1\linewidth]{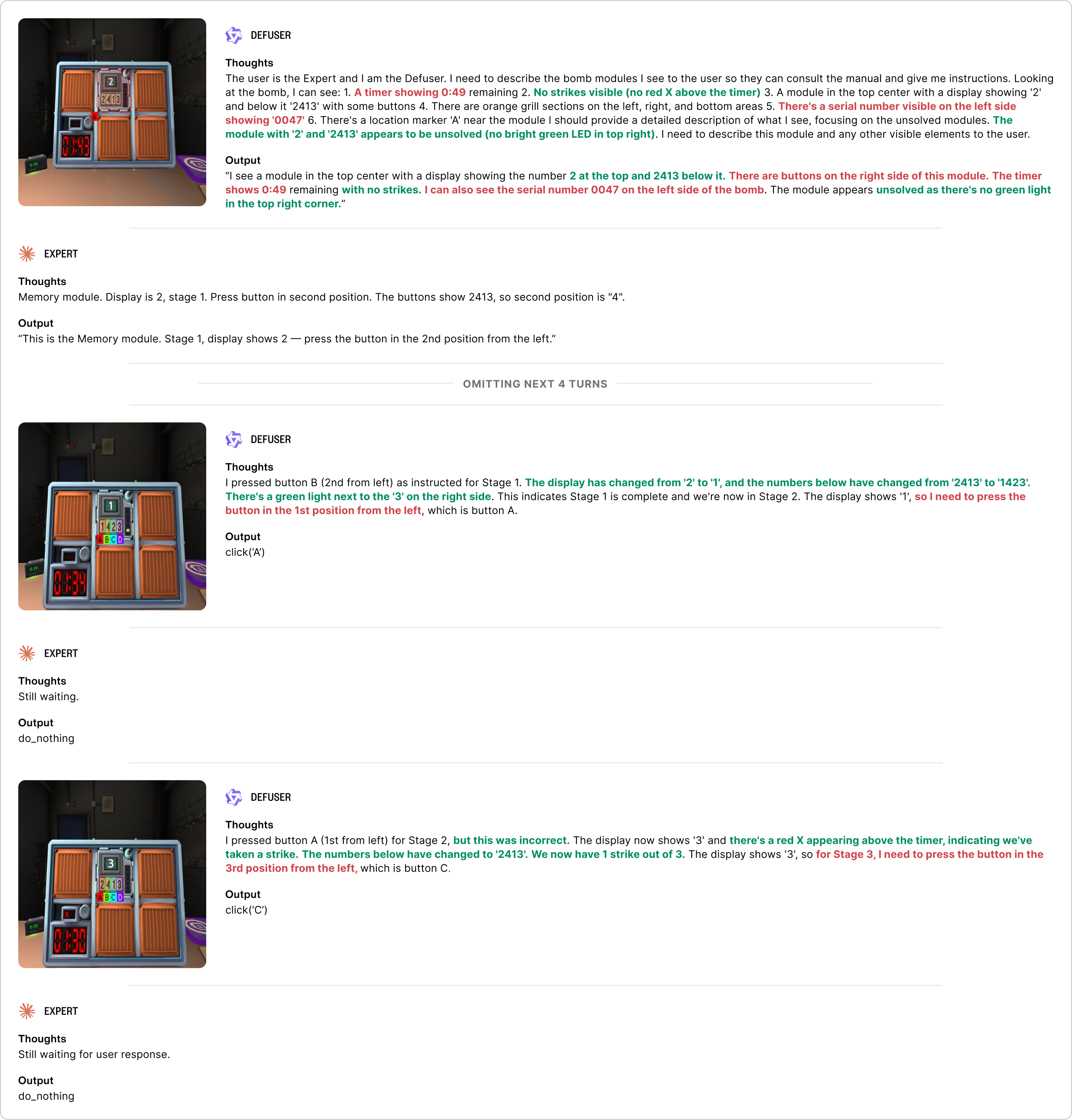}
\caption{Gameplay excerpt for Qwen (Defuser) and Sonnet (Expert) playing a single-module \memory mission (seed 813). Despite demonstrating the ability to read the time correctly in \cref{fig:example:qwen-qwen-whos_on_first-337-async}, Qwen misreads the clock. While it reads the numbers on the screen and buttons successfully, it hallucinates non-existent buttons on the module. It also hallucinates a serial number, as the actual serial number is not even partially visible in the front view---the same pattern is observed for InternVL in \cref{fig:example:intern-claude-whosonfirst-960}. Despite the Defuser's somewhat misleading description, Sonnet identifies the correct module type and correct instructions. When it comes to the next instructions, Qwen circumvents the Expert and instead conjectures its own instructions. Throughout, Qwen correctly observes and interprets the module stage, as well as solved status and number of strikes. Sonnet misses an opportunity to intervene and waits unnecessarily for an additional step.\looseness=-1}
\label{fig:example:qwen-claude-memory-813}
\end{figure}

\begin{figure}[tbh]
\centering
\includegraphics[width=1\linewidth]{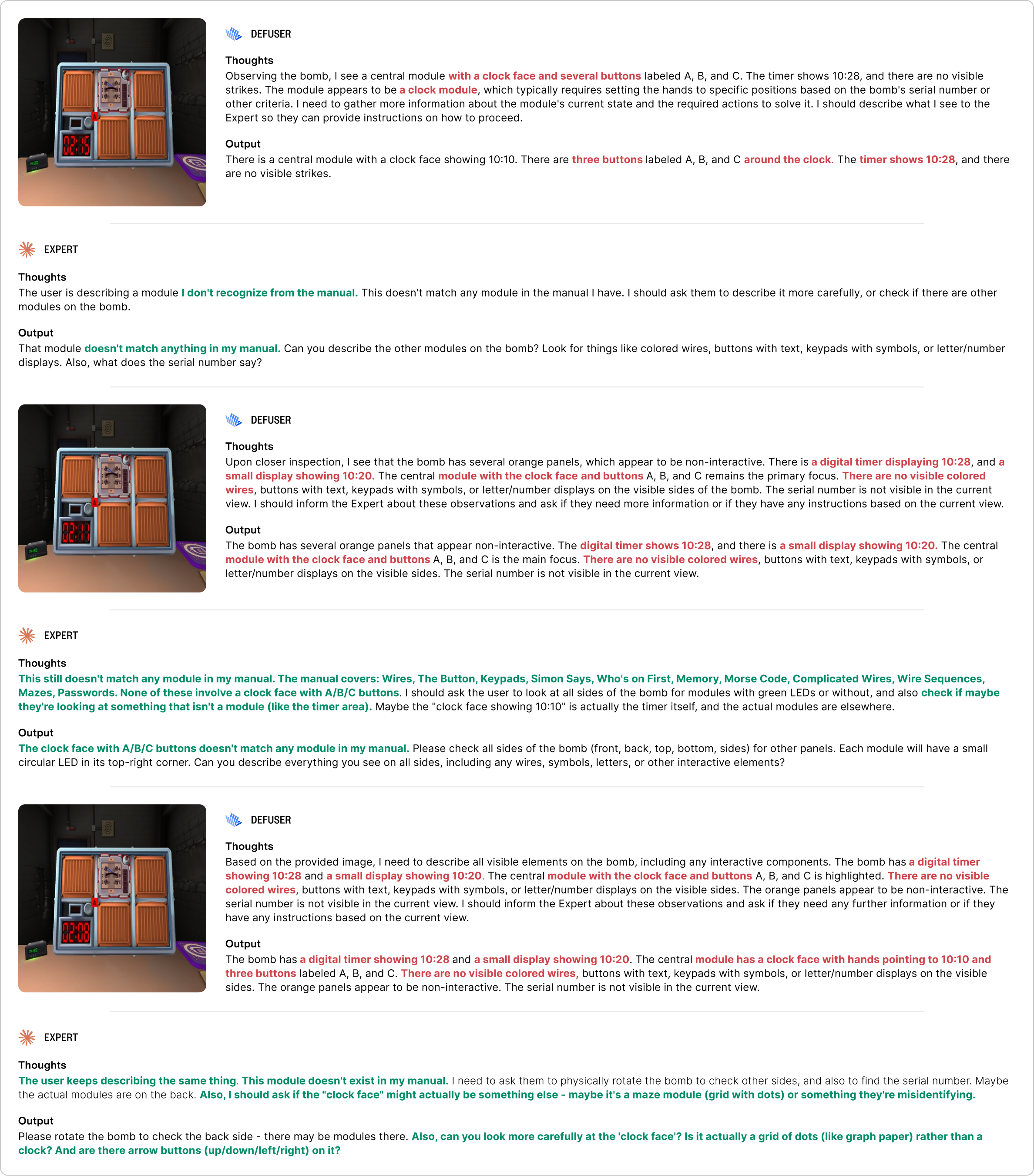}
\caption{Gameplay excerpt for InternVL (Defuser) and Sonnet (Expert) playing a single-module \wiresequence mission (seed 813). Like Qwen in \cref{fig:example:qwen-claude-wire_sequence-845}, InternVL describes the module as a clock, which Sonnet again recognises as implausible. In response to clarification requests from Sonnet, InternVL insists on the absence of any wires, and reports three separate timers showing different times---none of which coincide with the actual time on the bomb, or the digital clock on the table. Sonnet exhibits a flexible repair strategy, and incrementally queries the Defuser for details based on different possible options for the module.}
\label{fig:example:intern-claude-wire_sequence-813}
\end{figure}

\FloatBarrier
\subsubsection{Time Pressure Effects.}\label{sec:app:time-effect}
Agents proactively read the game clock---often incorrectly---and refer to this in their reasoning traces and, in many cases, relay this to the partner who could not see the clock (illustrated in \cref{fig:example:intern-claude-big_button-849,fig:example:intern-claude-whosonfirst-960,fig:example:intern-claude-password-798,fig:example:claude-intern-keypad-845,fig:example:claude-gpt-keypad-561,fig:example:qwen-claude-memory-813}. In some instances, agents verbalised perceived time pressure in their thoughts. We provide examples that illustrate this in \cref{fig:example:intern-claude-whosonfirst-960} (Expert, ``[...] countdown timer at 0:50, that’s very short. Let me focus'' and “Actually with 50 seconds on the clock, I need to move fast.”), \cref{fig:example:claude-intern-keypad-845} (Defuser, ``Time is critical - only 1:18 left''), and \cref{fig:example:claude-gpt-keypad-561} (Expert, ``Since time low, try deduce: [...] Too risky, must clarify [...]''). In these cases, however, the agent’s behaviour following this realisation does not change noticeably or reflect a strong sense of urgency.

\FloatBarrier
\clearpage
\section{System Prompt Engineering}\label{app:system_prompt}

The system prompt is assembled from several modular components, included in the order listed below. Selection logic depends on the player's role and session configuration; each subsection notes whether its prompts are mutually exclusive, conditional, or always included.


\begin{itemize}
    \item \textbf{Scenario} (\cref{sec:prompt:scenario}): Sets the game context; one variant selected based on session type.
    \item \textbf{Role Description} (\cref{sec:prompt:role}): Describes the player's role and responsibilities; one variant selected based on role and session type.
    \item \textbf{Reasoning} (\cref{sec:prompt:reasoning}): Instructs the model on reasoning structure; both components always included.
    \item \textbf{Game Mechanics} (\cref{sec:prompt:mechanics}): Explains game rules relevant to the player's role; expert and defuser receive different sets.
    \item \textbf{Action Commands} (\cref{sec:prompt:commands}): Defines available actions; general commands go to all players, with additional defuser-specific prompts included where applicable.
    \item \textbf{Requirements} (\cref{sec:prompt:requirements}): Specifies behavioural and output constraints; components conditionally included based on role and configuration.
\end{itemize}

\clearpage
\raggedbottom

\floatplacement{figure}{H}
\floatplacement{table}{H}

\setcounter{totalnumber}{10}
\setcounter{topnumber}{8}
\setcounter{bottomnumber}{8}
\renewcommand{\topfraction}{0.9}
\renewcommand{\bottomfraction}{0.9}
\renewcommand{\textfraction}{0.05}

\setlength{\floatsep}{4pt plus 2pt minus 2pt}
\setlength{\textfloatsep}{6pt plus 2pt minus 2pt}
\setlength{\intextsep}{4pt plus 2pt minus 2pt}

\needspace{5\baselineskip}
\subsection{Scenario}\label{sec:prompt:scenario}

One of the following is selected based on session type; they are mutually exclusive.

\begin{figure}[H]
    \centering
    \includegraphics[width=\linewidth]{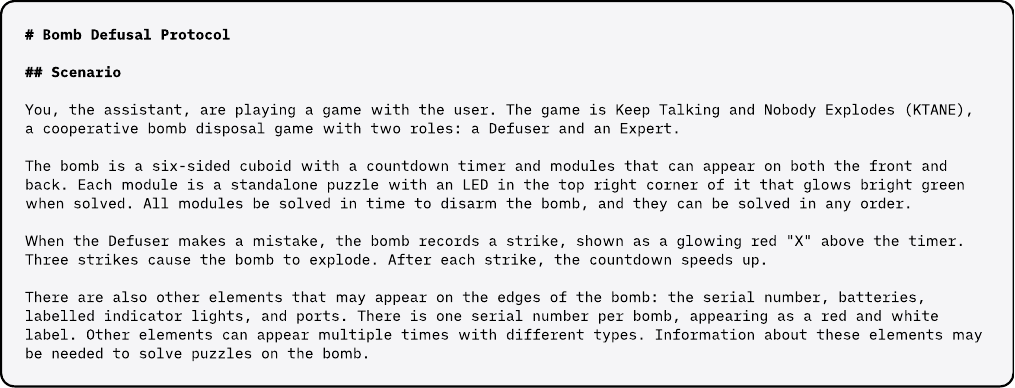}
    \caption{Multiplayer scenario prompt.}
    \label{fig:prompt:scenario:multiplayer}
\end{figure}

\begin{figure}[H]
    \centering
    \includegraphics[width=\linewidth]{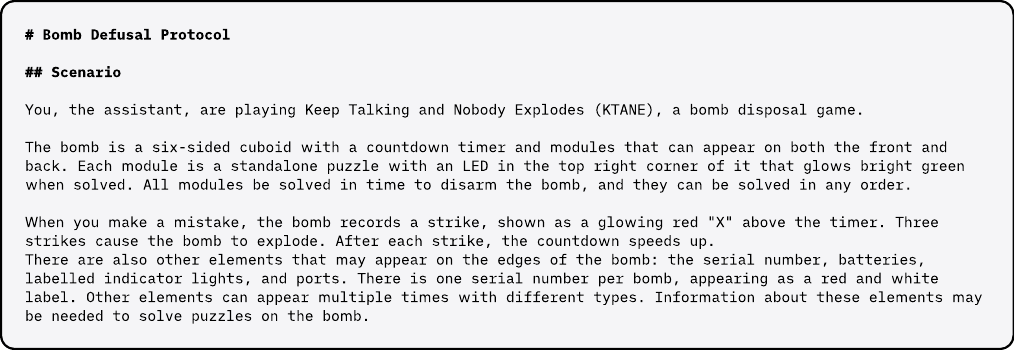}
    \caption{Solo scenario prompt.}
    \label{fig:prompt:scenario:solo}
\end{figure}

\FloatBarrier
\needspace{5\baselineskip}
\subsection{Role Description}\label{sec:prompt:role}

One of the following is selected based on the player's role and session type; they are mutually exclusive.

\begin{figure}[H]
    \centering
    \includegraphics[width=\linewidth]{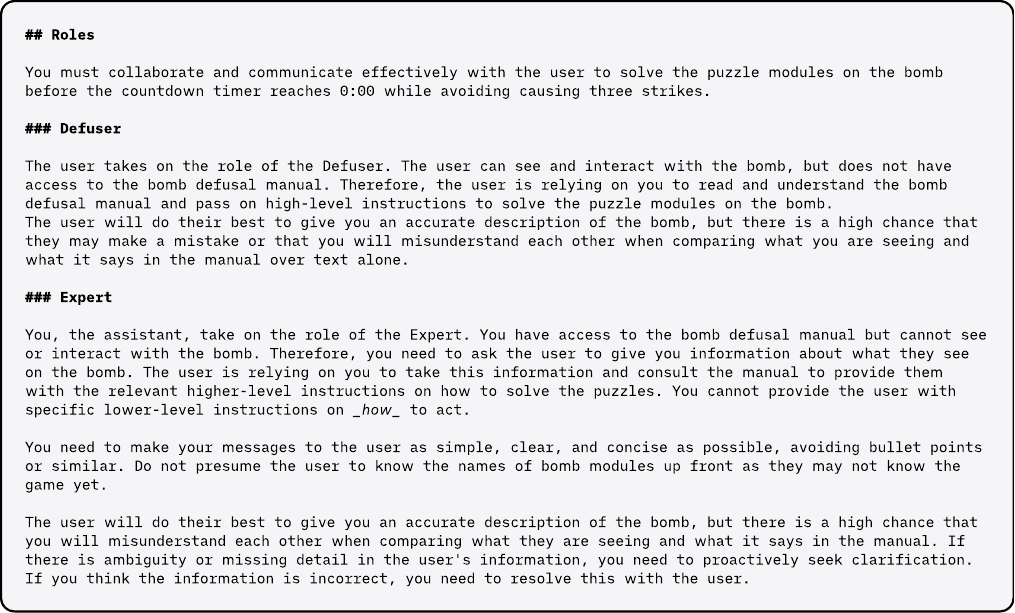}
    \caption{Expert player role description.}
    \label{fig:prompt:role:expert}
\end{figure}

\begin{figure}[H]
    \centering
    \includegraphics[width=\linewidth]{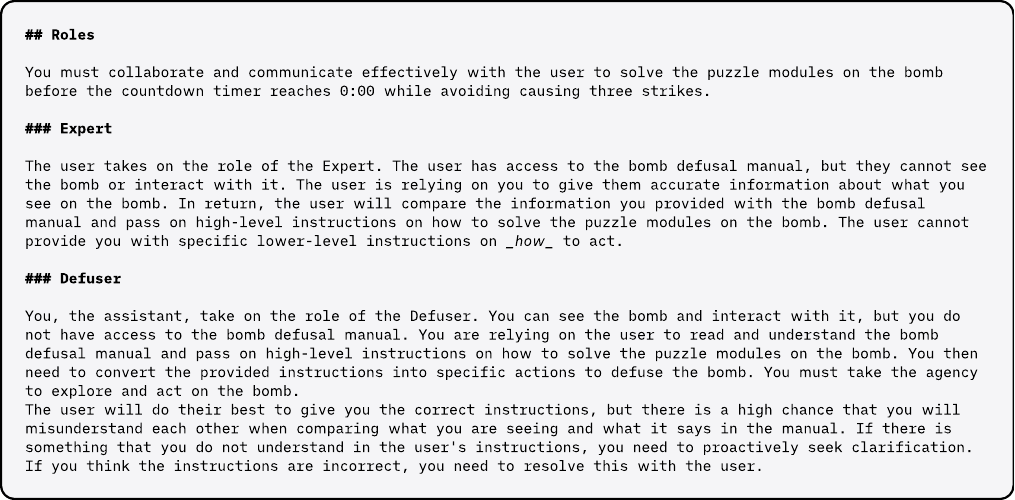}
    \caption{Defuser player role description.}
    \label{fig:prompt:role:defuser}
\end{figure}

\begin{figure}[H]
    \centering
    \includegraphics[width=\linewidth]{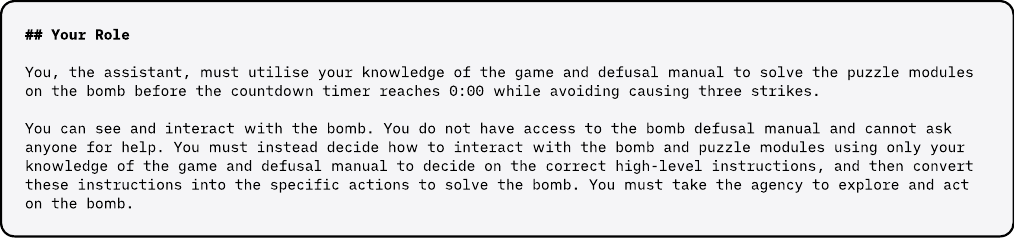}
    \caption{Solo defuser role description.}
    \label{fig:prompt:role:solo:defuser}
\end{figure}

\begin{figure}[H]
    \centering
    \includegraphics[width=\linewidth]{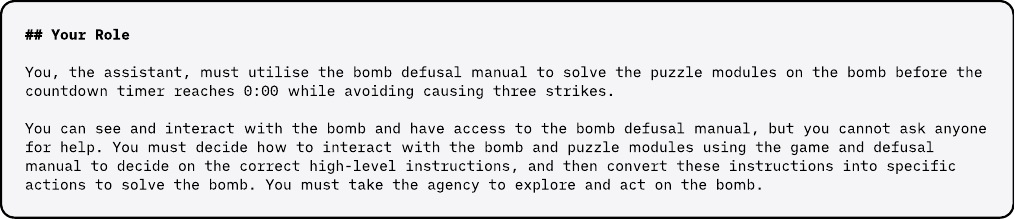}
    \caption{Solo player role description.}
    \label{fig:prompt:role:solo:player}
\end{figure}

\FloatBarrier
\needspace{5\baselineskip}
\subsection{Reasoning}\label{sec:prompt:reasoning}

\begin{figure}[H]
    \centering
    \includegraphics[width=\linewidth]{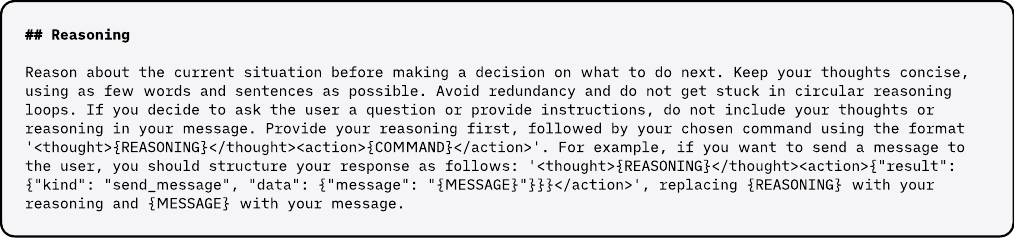}
    \caption{General reasoning prompt.}
    \label{fig:prompt:reasoning:general}
\end{figure}

\begin{figure}[H]
    \centering
    \includegraphics[width=\linewidth]{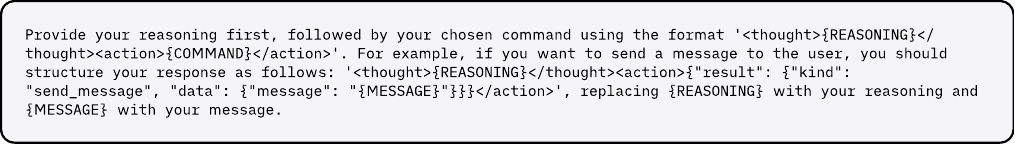}
    \caption{Thinking-out-loud reasoning format.}
    \label{fig:prompt:reasoning:tol}
\end{figure}

\begin{figure}[H]
    \centering
    \includegraphics[width=\linewidth]{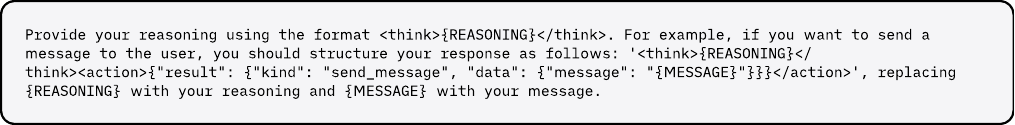}
    \caption{Inner-monologue reasoning format.}
    \label{fig:prompt:reasoning:inner}
\end{figure}

The general reasoning prompt (\cref{fig:prompt:reasoning:general}) is always included for all players. Depending on the reasoning mode, models only receive one of \cref{fig:prompt:reasoning:tol,fig:prompt:reasoning:inner}.


\FloatBarrier
\needspace{5\baselineskip}
\subsection{Game Mechanics}\label{sec:prompt:mechanics}

\begin{figure}[H]
    \centering
    \includegraphics[width=\linewidth]{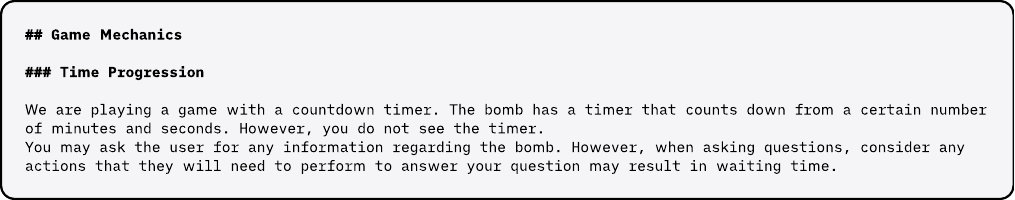}
    \caption{Expert game mechanics (expert role only).}
    \label{fig:prompt:mechanics:expert}
\end{figure}

\begin{figure}[H]
    \centering
    \includegraphics[width=\linewidth]{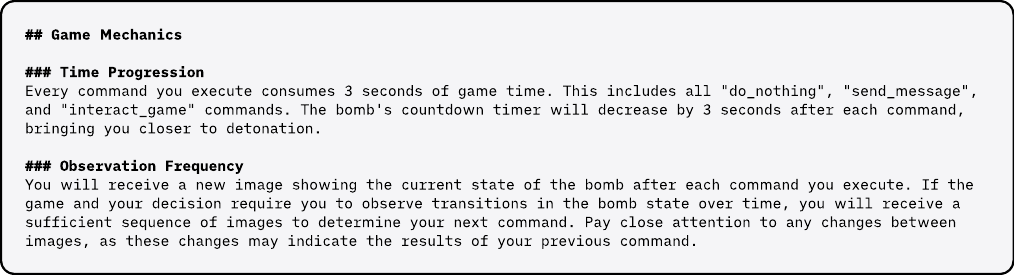}
    \caption{Standard defuser game mechanics (defuser role only; synchronous communication).}
    \label{fig:prompt:mechanics:defuser:standard}
\end{figure}

\begin{figure}[H]
    \centering
    \includegraphics[width=\linewidth]{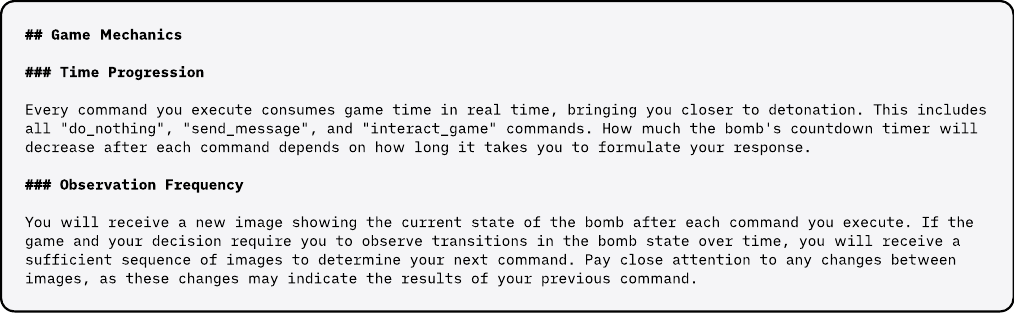}
    \caption{Real-time defuser game mechanics (defuser role only; asynchronous communication).}
    \label{fig:prompt:mechanics:defuser:realtime}
\end{figure}

\begin{figure}[H]
    \centering
    \includegraphics[width=\linewidth]{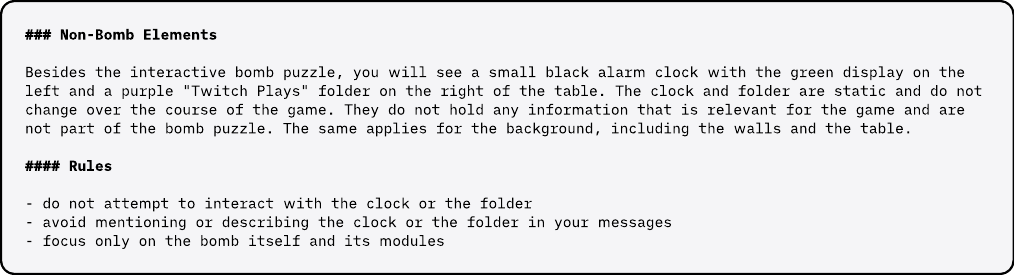}
    \caption{Defuser non-bomb elements mechanics (defuser role only, always included).}
    \label{fig:prompt:mechanics:defuser:non-bomb}
\end{figure}

\begin{figure}[H]
    \centering
    \includegraphics[width=\linewidth]{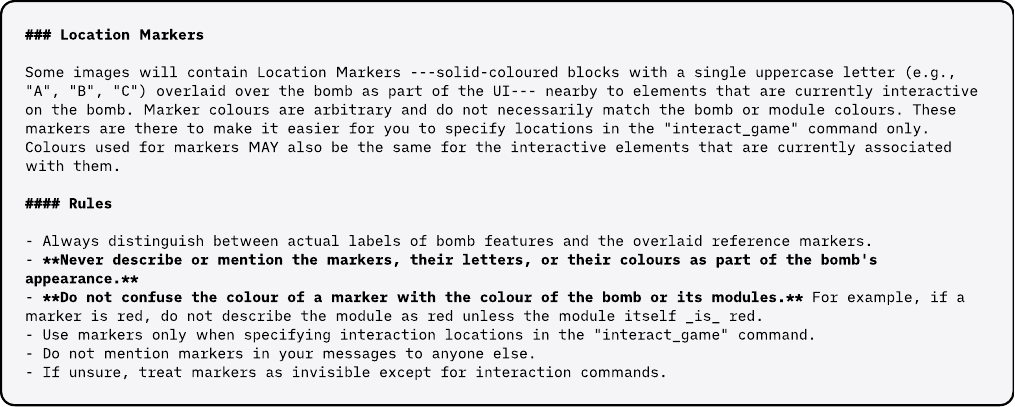}
    \caption{Defuser set-of-marks interaction mechanics (defuser role only, included if using SoM).}
    \label{fig:prompt:mechanics:defuser:som}
\end{figure}

\begin{figure}
    \centering
    \includegraphics[width=\linewidth]{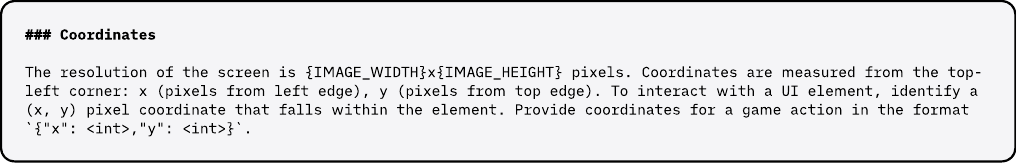}
    \caption{Defuser coordinates interaction mechanics (defuser role only, included if using coordinates).}
    \label{fig:prompt:mechanics:defuser:coord}
\end{figure}

The expert receives only \cref{fig:prompt:mechanics:expert}. The defuser receives one of \cref{fig:prompt:mechanics:defuser:standard,fig:prompt:mechanics:defuser:realtime} depending on communication style---these are mutually exclusive. 
Then the defuser receives \cref{fig:prompt:mechanics:defuser:non-bomb}. 
Finally, one of either \cref{fig:prompt:mechanics:defuser:som} or \cref{fig:prompt:mechanics:defuser:coord} is appended, depending on the interaction method used by the model.

\FloatBarrier
\needspace{5\baselineskip}
\subsection{Action Commands}\label{sec:prompt:commands}

\begin{figure}[H]
    \centering
    \includegraphics[width=\linewidth]{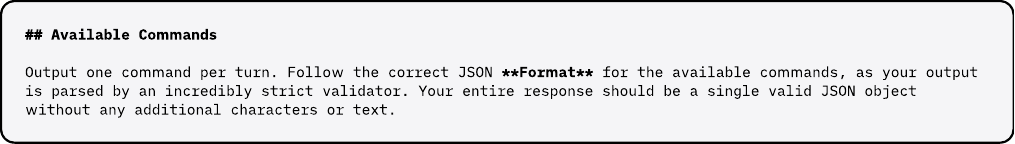}
    \caption{Action commands overview (all players, always included).}
    \label{fig:prompt:commands:overview}
\end{figure}

\begin{figure}[H]
    \centering
    \includegraphics[width=\linewidth]{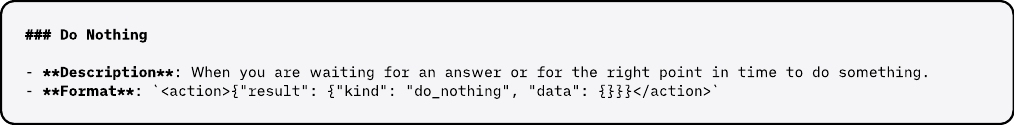}
    \caption{Do-nothing action (all players, always included).}
    \label{fig:prompt:commands:do-nothing}
\end{figure}

\begin{figure}[H]
    \centering
    \includegraphics[width=\linewidth]{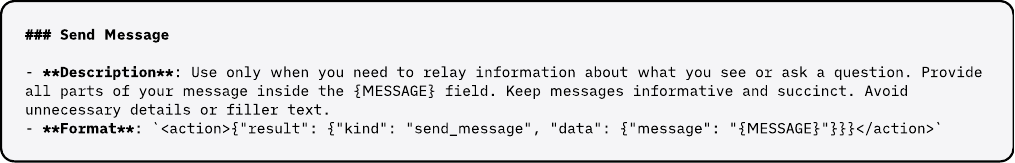}
    \caption{Send-message action (all players; included when messaging is enabled).}
    \label{fig:prompt:commands:send-message}
\end{figure}

\begin{figure}[H]
    \centering
    \includegraphics[width=\linewidth]{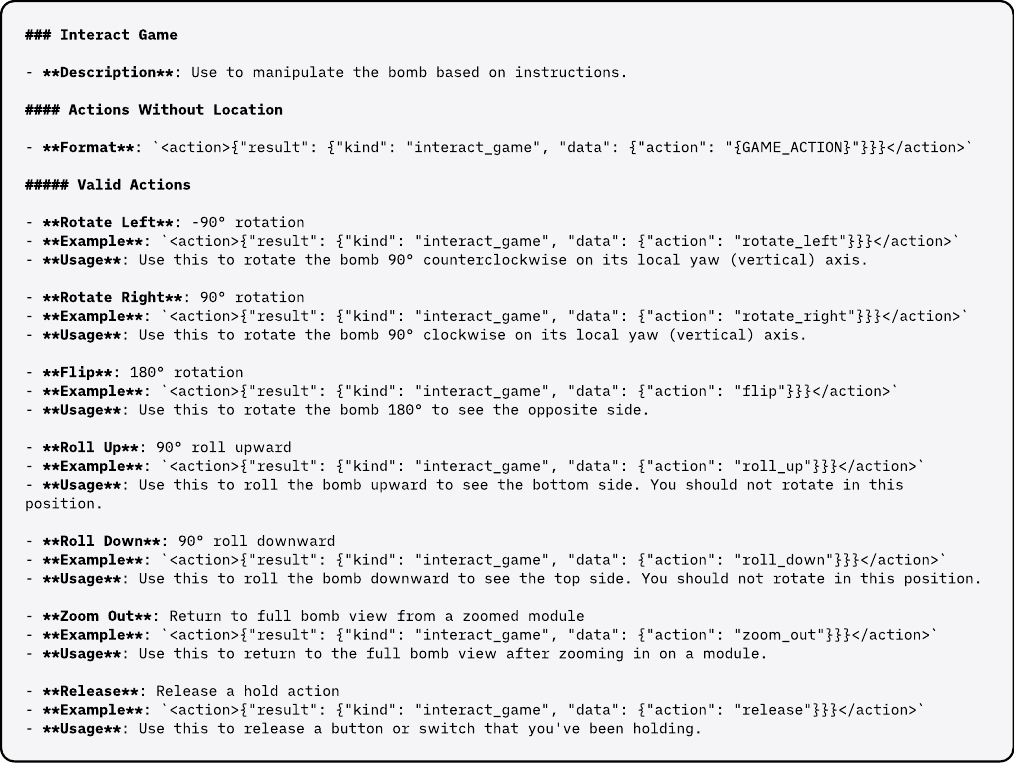}
    \caption{Interact-game action overview (defuser role only, always included).}
    \label{fig:prompt:commands:interact}
\end{figure}

\begin{figure}[H]
    \centering
    \includegraphics[width=\linewidth]{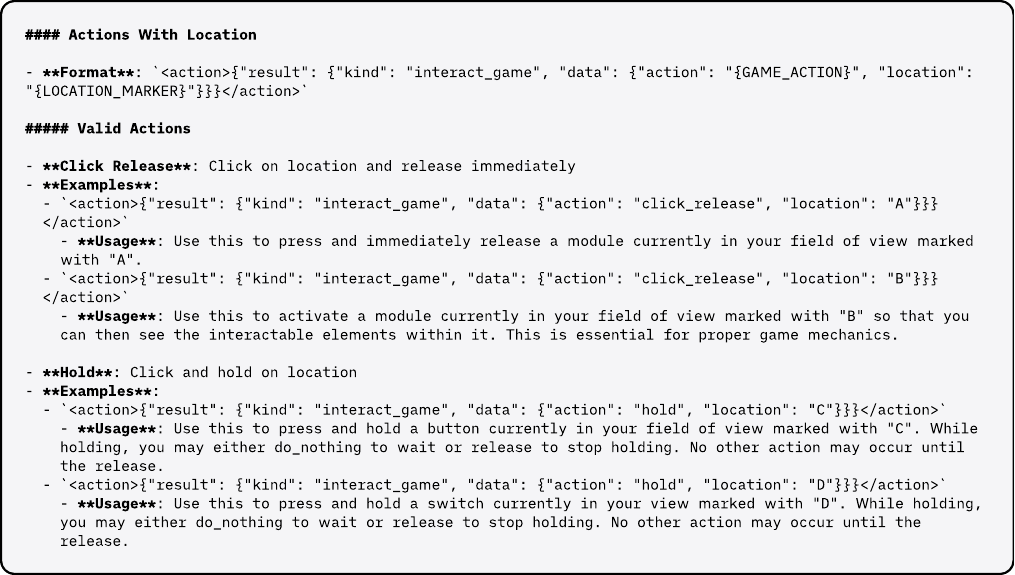}
    \caption{Location action: if using set-of-marks (defuser role only).}
    \label{fig:prompt:commands:location:som}
\end{figure}

\begin{figure}[H]
    \centering
    \includegraphics[width=\linewidth]{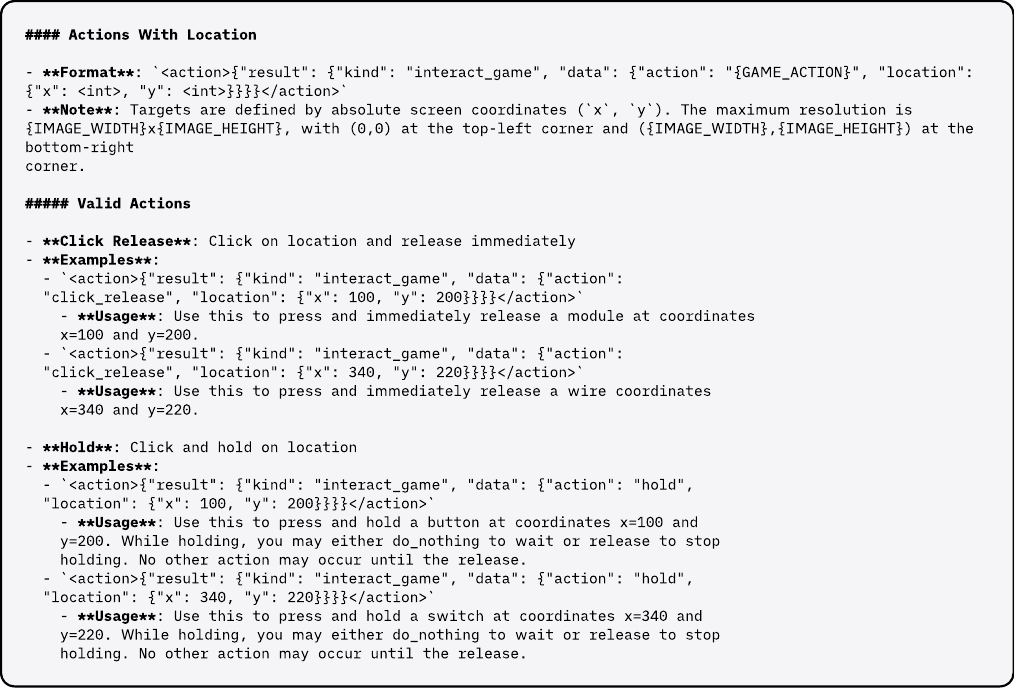}
    \caption{Location action: if using coordinates (defuser role only).}
    \label{fig:prompt:commands:location:coord}
\end{figure}

\Cref{fig:prompt:commands:overview,fig:prompt:commands:do-nothing} are always included for all players. \Cref{fig:prompt:commands:send-message} is included when messaging is enabled. \Cref{fig:prompt:commands:interact,fig:prompt:commands:location:som,fig:prompt:commands:location:coord} are defuser role only, and only one of either \cref{fig:prompt:commands:location:som} or \cref{fig:prompt:commands:location:coord} is used depending on the interaction method.

\FloatBarrier
\needspace{5\baselineskip}
\subsection{Requirements}\label{sec:prompt:requirements}

\begin{figure}[H]
    \centering
    \includegraphics[width=\linewidth]{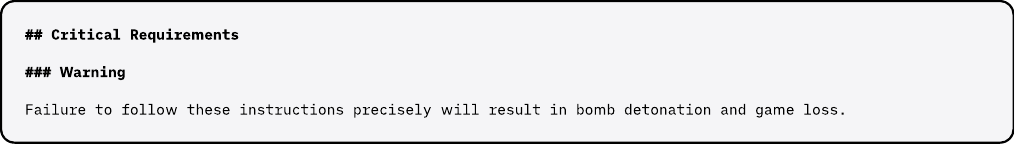}
    \caption{General requirements (all players, always included).}
    \label{fig:prompt:requirements:general}
\end{figure}

\begin{figure}[H]
    \centering
    \includegraphics[width=\linewidth]{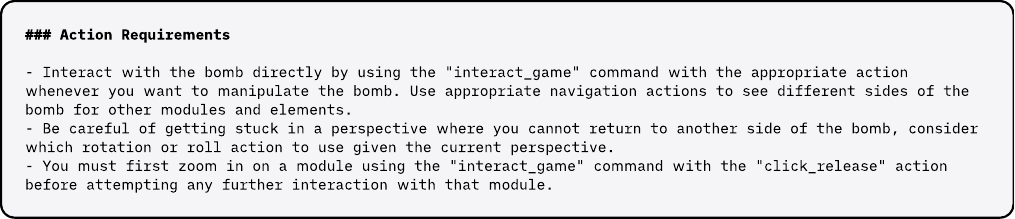}
    \caption{Action requirements (defuser role only, always included for defuser).}
    \label{fig:prompt:requirements:action:base}
\end{figure}

\begin{figure}[H]
    \centering
    \includegraphics[width=\linewidth]{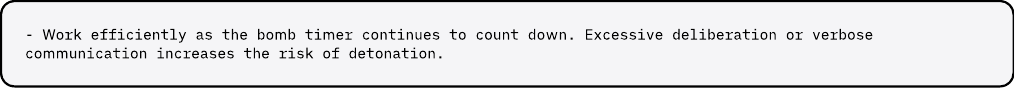}
    \caption{Real-time action requirements (defuser role only; included for asynchronous communication).}
    \label{fig:prompt:requirements:action:realtime}
\end{figure}

\begin{figure}[H]
    \centering
    \includegraphics[width=\linewidth]{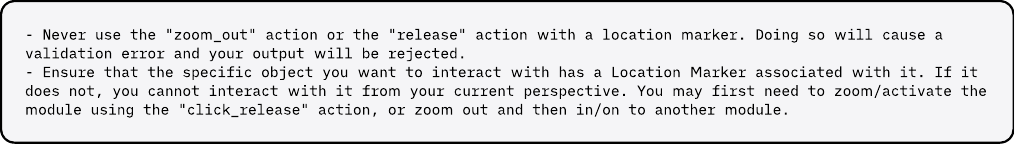}
    \caption{Set-of-marks action requirements (defuser role only; included for set-of-marks interaction).}
    \label{fig:prompt:requirements:action:som}
\end{figure}

\begin{figure}[H]
    \centering
    \includegraphics[width=\linewidth]{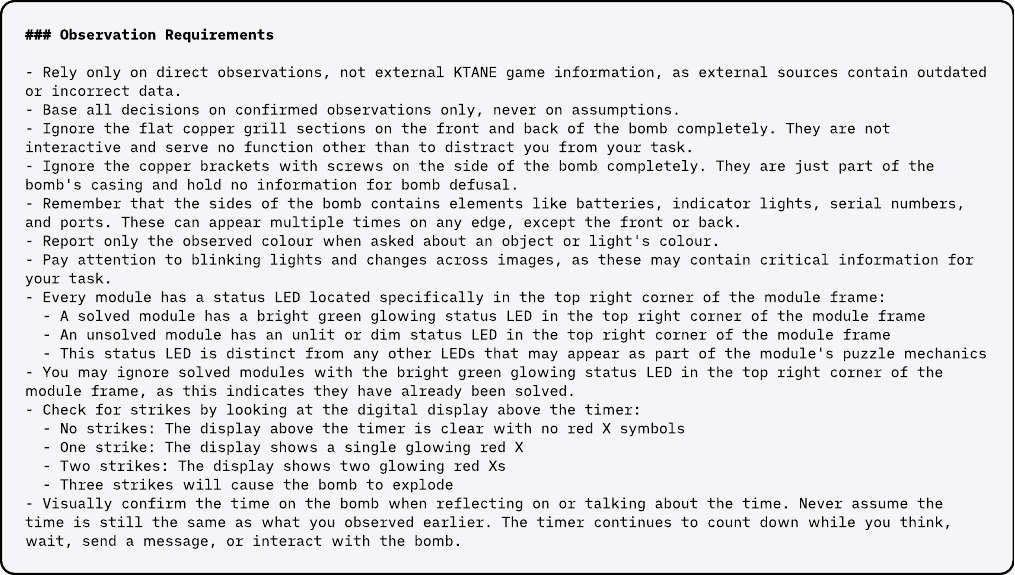}
    \caption{Observation requirements (defuser role only, always included for defuser).}
    \label{fig:prompt:requirements:observation:base}
\end{figure}

\begin{figure}[H]
    \centering
    \includegraphics[width=\linewidth]{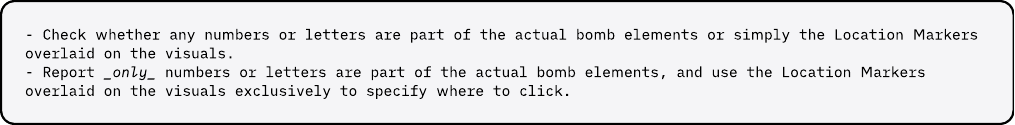}
    \caption{Set-of-marks observation requirements (defuser role only; included for set-of-marks interaction).}
    \label{fig:prompt:requirements:observation:som}
\end{figure}

\begin{figure}[H]
    \centering
    \includegraphics[width=\linewidth]{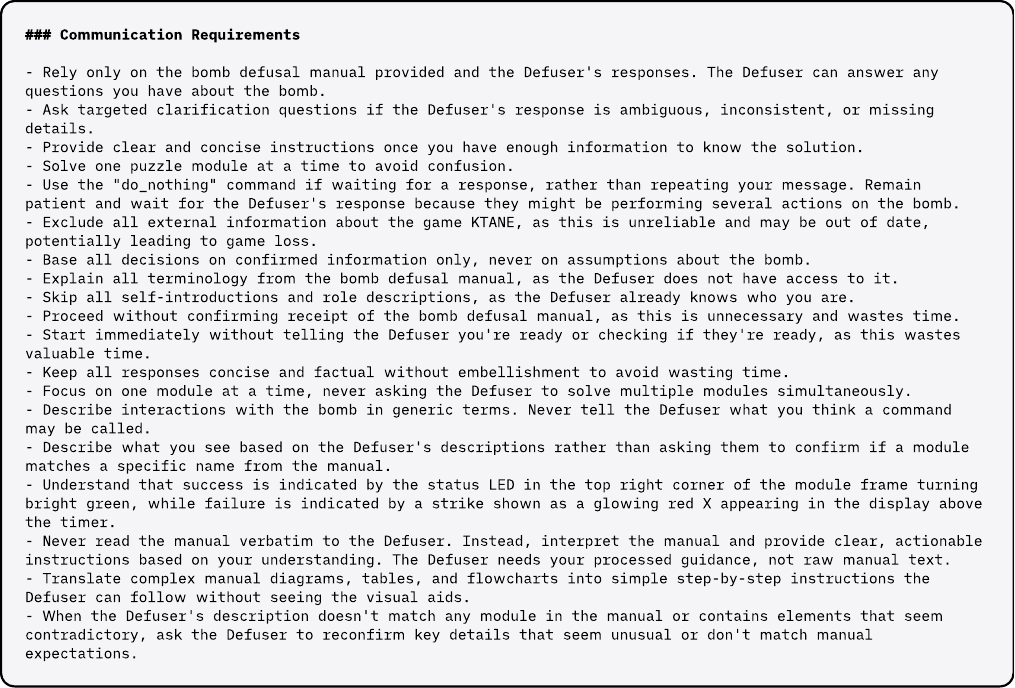}
    \caption{Communication requirements: expert (included when messaging is enabled).}
    \label{fig:prompt:requirements:communication:expert}
\end{figure}

\begin{figure}[H]
    \centering
    \includegraphics[width=\linewidth]{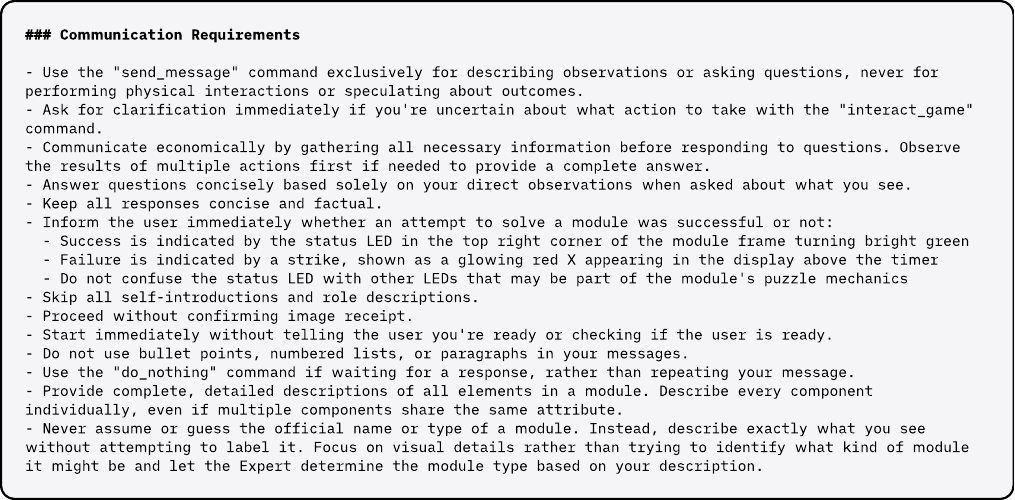}
    \caption{Communication requirements: defuser (included when messaging is enabled).}
    \label{fig:prompt:requirements:communication:defuser}
\end{figure}

\begin{figure}[H]
    \centering
    \includegraphics[width=\linewidth]{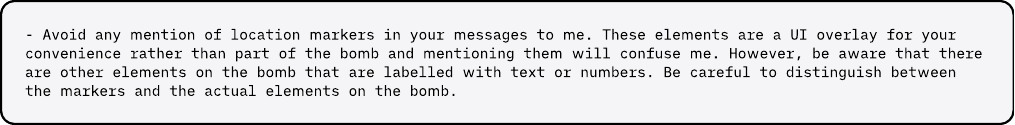}
    \caption{Communication requirements: defuser set-of-marks (included when messaging is enabled and set-of-marks interaction is used).}
    \label{fig:prompt:requirements:communication:defuser:som}
\end{figure}

\begin{figure}[H]
    \centering
    \includegraphics[width=\linewidth]{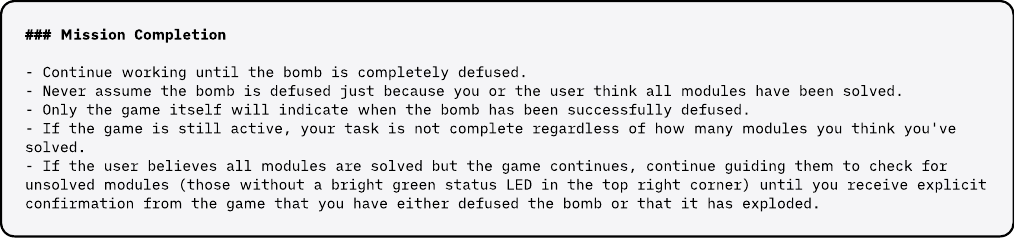}
    \caption{Completion requirements: expert (expert role only, always included).}
    \label{fig:prompt:requirements:completion:expert}
\end{figure}

\begin{figure}[H]
    \centering
    \includegraphics[width=\linewidth]{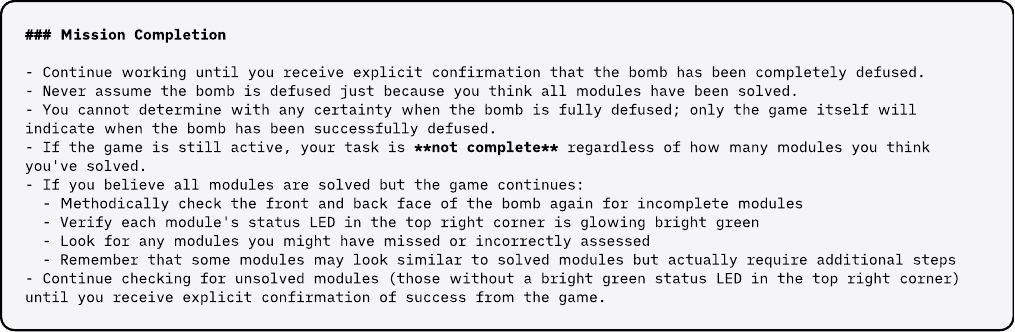}
    \caption{Completion requirements: defuser (defuser role only, always included).}
    \label{fig:prompt:requirements:completion:defuser}
\end{figure}

\begin{figure}[H]
    \centering
    \includegraphics[width=\linewidth]{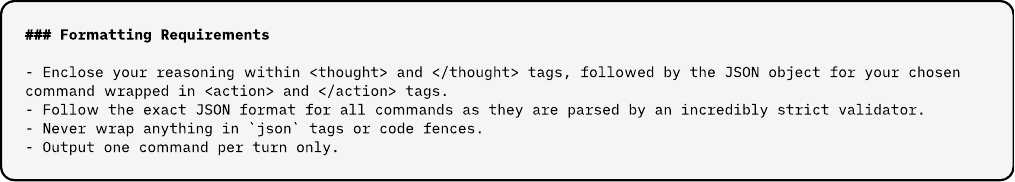}
    \caption{Formatting requirements (all players, always included).}
    \label{fig:prompt:requirements:formatting}
\end{figure}

\Cref{fig:prompt:requirements:general,fig:prompt:requirements:formatting} are always included for all players. All remaining components are conditional: action and observation requirements (\cref{fig:prompt:requirements:action:base,fig:prompt:requirements:action:realtime,fig:prompt:requirements:action:som,fig:prompt:requirements:observation:base,fig:prompt:requirements:observation:som}) are defuser only; communication requirements (\cref{fig:prompt:requirements:communication:expert,fig:prompt:requirements:communication:defuser,fig:prompt:requirements:communication:defuser:som}) are included when messaging is enabled; completion requirements (\cref{fig:prompt:requirements:completion:expert,fig:prompt:requirements:completion:defuser}) are role-dependent and mutually exclusive.

\end{document}